\documentclass[journal]{IEEEtran}
\usepackage{amsmath,amsfonts}
\usepackage{bm}     
\usepackage{amstext}
\usepackage{algorithmic}
\usepackage{algorithm}
\usepackage{array}
\usepackage[caption=false,font=normalsize,labelfont=sf,textfont=sf]{subfig}
\usepackage{textcomp}
\usepackage{stfloats}
\usepackage{url}
\usepackage{verbatim}
\usepackage{graphicx}
\usepackage{cite}

\usepackage{hyperref}
\usepackage[justification=centering]{caption}
\usepackage[utf8]{inputenc}

\usepackage{graphicx}
\usepackage{array}
\usepackage{amsmath}

\usepackage{tabularx}
\usepackage{booktabs}
\usepackage{ragged2e}  

\usepackage{hyperref}

\usepackage{graphicx}
\usepackage{amsmath}
\usepackage{amsfonts}
\usepackage{algorithm}
\usepackage{algorithmic}
\usepackage{makecell}
\usepackage{booktabs}

\usepackage{xcolor}
\usepackage{amsmath}
\usepackage{graphicx}
\usepackage{array}

\usepackage{booktabs}
\usepackage{multirow}

\usepackage{amssymb} 
\usepackage{pifont} 
\usepackage{multirow}
\usepackage{graphicx}

\begin{document}

\title{From Large AI Models to Agentic AI: A Tutorial on Future Intelligent Communications}

\author{Feibo Jiang, \textit{Senior Member, IEEE}, Cunhua Pan, \textit{Senior Member, IEEE}, Li Dong, Kezhi Wang, \textit{Senior Member, IEEE}, Octavia A. Dobre, \textit{Fellow, IEEE}, and Merouane Debbah, \textit{Fellow, IEEE}

	\thanks{
		Feibo Jiang (jiangfb@hunnu.edu.cn) is with Hunan Provincial Key Laboratory of Intelligent Computing and Language Information Processing, Hunan Normal University, Changsha, China.
		
		Cunhua Pan (cpan@seu.edu.cn) is with the National Mobile Communications Research Laboratory, Southeast University, Nanjing, China.
		
		Li Dong (Dlj2017@hunnu.edu.cn) is with Changsha Social Laboratory of Artificial Intelligence, Hunan University of Technology and Business, Changsha, China.
		
		Kezhi Wang (Kezhi.Wang@brunel.ac.uk) is with the Department of Computer Science, Brunel University London, UK.

        
        Octavia A. Dobre (odobre@mun.ca) is with the Faculty of Engineering and Applied Science, Memorial University, St. John’s, NL A1B 3X5, Canada.


        Merouane Debbah (merouane.debbah@ku.ac.ae) is with the 6G Research Center, Khalifa University of Science and Technology, Abu Dhabi 127788, UAE.

        GitHub link: \url{https://github.com/jiangfeibo/ComAgent}.

		

	}



}
\maketitle

\begin{abstract}
With the advent of 6G communications, intelligent communication systems face multiple challenges, including constrained perception and response capabilities, limited scalability, and low adaptability in dynamic environments. This tutorial provides a systematic introduction to the principles, design, and applications of Large Artificial Intelligence Models (LAMs) and Agentic AI technologies in intelligent communication systems, aiming to offer researchers a comprehensive overview of cutting-edge technologies and practical guidance. First, we outline the background of 6G communications, review the technological evolution from LAMs to Agentic AI, and clarify the tutorial's motivation and main contributions. Subsequently, we present a comprehensive review of the key components required for constructing LAMs, including Transformers, Vision Transformers (ViTs), Variational AutoEncoders (VAEs), diffusion models, Diffusion Transformers (DiTs), and Mixture of Experts (MoEs). We further categorize LAMs and analyze their applicability, covering Large Language Models (LLMs), Large Vision Models (LVMs), Large Multimodal Models (LMMs), Large Reasoning Models (LRMs), and lightweight LAMs. Next, we propose a LAM-centric design paradigm tailored for communications, encompassing dataset construction and both internal and external learning approaches. Building upon this, we develop an LAM-based Agentic AI system for intelligent communications, clarifying its core components such as planners, knowledge bases, tools, and memory modules, as well as its interaction mechanisms, including both single-agent and multi-agent interactions. We also introduce a multi-agent framework with data retrieval, collaborative planning, and reflective evaluation for 6G. Subsequently, we provide a detailed overview of the applications of LAMs and Agentic AI in communication scenarios. Finally, we summarize the research challenges and future directions in current studies, aiming to support the development of efficient, secure, and sustainable next-generation intelligent communication systems.

\end{abstract}


\section{Introduction}
With the continuous evolution of 6G communication, intelligence has become a core direction for the development of future wireless networks. Traditional communication systems, which rely on static rules and predefined algorithms, struggle to cope with rapidly changing network topologies and dynamic environments. In this context, Large Artificial Intelligence Models (LAMs) have achieved remarkable success in communications due to their advantages in cognitive decision-making and data generation. Meanwhile, Agentic AI, as a more advanced technology based on LAMs, can actively make decisions and self-optimize, offering novel solutions for intelligent resource management and optimization in 6G networks. Therefore, the technological transition from LAMs to Agentic AI holds significant importance for supporting the evolution of intelligent communication systems from model-driven to agent-driven paradigms.

\subsection{Background}
\begin{figure*}[htpb]
		\centering
		\includegraphics[width=0.7\textwidth,height=0.7\textwidth]{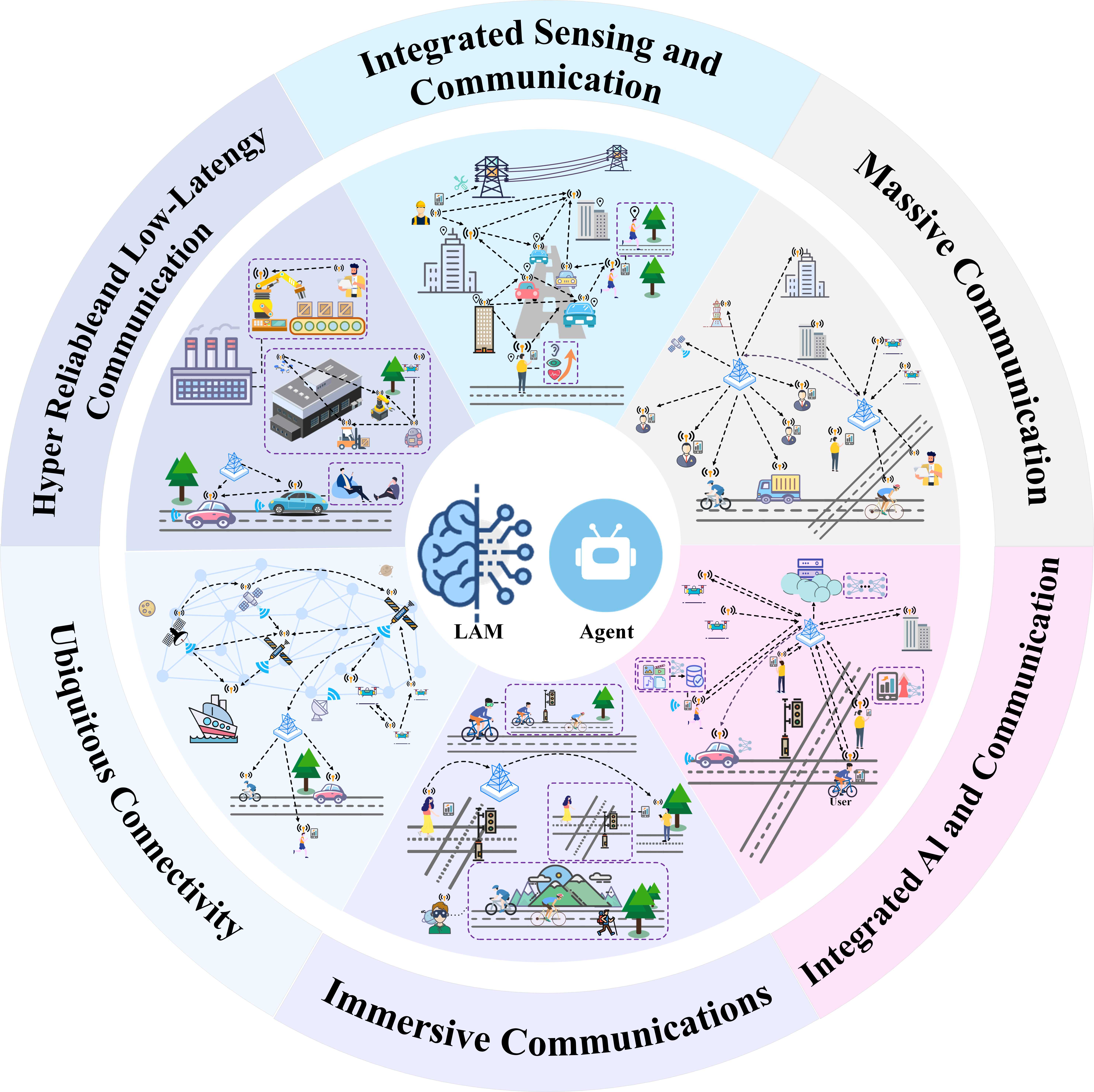}
		\caption{LAMs and Agentic AI empowered 6G.}
		\label{fig:fig0}
\end{figure*}

The goal of 6G is to build an intelligent world of ubiquitous connectivity, delivering unprecedented information experiences to human society. In the International Mobile Telecommunications for 2030 (IMT-2030) framework proposed by ITU-R, six key capability modules are defined to support the comprehensive development of the future wireless communication ecosystem. These include Integrated Sensing And Communication (ISAC), which deeply fuses environmental sensing with communication functions to endow networks with human-like perception capabilities, enabling applications such as intelligent transportation and smart grids; massive communication, which supports the concurrent access of densely distributed devices to meet the real-time communication demands of massive terminals in smart cities and industrial Internet of Things (IoT); integrated AI and communication, which embeds LAMs into communication systems to realize network self-adaptation, self-optimization, and intelligent decision-making, significantly enhancing resource allocation and Quality of Service (QoS); immersive communications, which offers low-latency, high-bandwidth experiences such as holography and virtual-real fusion, driving new forms of interaction including the metaverse and AR/VR; ubiquitous connectivity, which constructs an all-domain communication network integrating space, air, sea, and land to eliminate geographical and spatial limitations; and hyper reliable and low-latency communication, which meets the stringent requirements of ultra-low latency and high reliability for critical applications like remote healthcare and autonomous vehicles. As illustrated in Fig. \ref{fig:fig0}, 6G will rely on these six core capabilities to build an intelligent, ubiquitous, and highly efficient future wireless communication ecosystem, where communication, sensing, computing, AI, and security are deeply integrated to deliver advanced communication services to users \cite{zhang20196g}.

\subsection{Historical Development}
The development of AI has witnessed a progressive evolution from early simple discriminative models to generative models, and further to LAMs and highly intelligent Agentic AI. This process not only demonstrates continuous innovation in model architectures and learning algorithms but also reflects significant advancements in AI's capabilities across multiple domains, including comprehension, generation, reasoning, and decision-making. The rise of LAMs has laid a solid foundation for the development of Agentic AI, while the emergence of Agentic AI marks a substantial leap in autonomous decision-making and complex task handling. The progression from LAMs to Agentic AI can be categorized into the following stages:

\subsubsection{Emergence Stage} The development of LAMs began in 2018, marked by significant works such as Google’s Bidirectional Encoder Representations from Transformers (BERT) \cite{kenton2019bert}. BERT is a bidirectional transformer model that achieved remarkable results in various Natural Language Processing (NLP) tasks through pre-training and fine-tuning. Its bidirectional nature allows BERT to excel in tasks such as sentence understanding and text classification. At the same time, OpenAI introduced GPT-1 \cite{radford2018improving}, a unidirectional transformer model focused on generative tasks. GPT-1 innovatively introduced the pre-training and fine-tuning paradigm and demonstrated the potential of large-scale pre-trained language models in NLP tasks. Subsequently, in 2019, GPT-2 \cite{radford2019language} further expanded the language model's scale and capabilities, showcasing its strong potential in generating text with approximately 1.5 billion parameters.

\subsubsection{Initial Stage} In 2020, OpenAI released the colossal language model GPT-3 \cite{brown2020language}, with 175 billion parameters, marking the entry of Large Language Models (LLMs) into the initial stage. GPT-3, with its massive parameter scale and complex training methods, excelled in multiple NLP tasks such as text generation, translation, question answering, and code generation. GPT-3 demonstrated the potential of large pre-trained models in multi-task learning and zero-shot learning. Meanwhile, Google released the T5 model \cite{raffel2020exploring}. This model introduced a unified text-to-text transformation framework, allowing various NLP tasks to be converted into text-to-text formats, thereby simplifying the task processing workflow. Additionally, T5 could handle translation, summarization, question answering, and text classification tasks in a unified model architecture. 

\subsubsection{Mature Stage} In 2022, GPT-3.5 \cite{ouyang2022training} was released, based on further optimizations of GPT-3, enhancing the model's performance and efficiency while improving response speed and accuracy in practical applications. In the same year, Anthropic released Claude, an LLM focused on enhancing model safety and transparency, aimed at reducing bias and misleading information. Additionally, Facebook AI Research introduced the Segment Anything Model (SAM) \cite{kirillov2023segment}, a Large Vision Model (LVM) specializing in image segmentation tasks, which made significant progress in image processing through extensive data pre-training.

\subsubsection{Multimodal Stage} In 2023, OpenAI launched the Large Multimodal Model (LMM) GPT-4 \cite{achiam2023gpt}, capable of processing both text and image data, further expanding the application scope of LMMs. GPT-4 combined visual and language understanding to achieve richer and more complex interaction capabilities. Concurrently, Google DeepMind released Gemini \cite{geminiteam2024geminifamilyhighlycapable}, an LMM capable of recognizing text, images, video, and code simultaneously. Gemini demonstrated outstanding performance across various tasks and application scenarios by generating high-quality code in mainstream programming languages and providing comprehensive safety assessments.

\subsubsection{Reasonging stage} In 2024, OpenAI released Open AI-o1 \cite{hayawi2024cross}, a model with enhanced reasoning capabilities. This model combines powerful cognitive processing with complex environmental modeling and prediction, advancing AI applications in decision-making and problem-solving while improving the logical reasoning performance of LAMs. In 2025, DeepSeek launched DeepSeek R1 \cite{guo2025deepseek}, a model that introduces advanced logical reasoning algorithms, demonstrating exceptional performance in complex tasks and dynamic environments, marking the official transition of LAMs into the era of complex reasoning.

\subsubsection{Agentic Stage} With the continuous maturation of LAM technology, agent system frameworks based on LAMs began to emerge. Early-generation agent frameworks, represented by AutoGPT \cite{yang2023auto} and BabyAGI \cite{nakajima2023babyagi}, demonstrated the potential of language understanding in task planning and execution. Concurrently, frameworks such as Microsoft’s OpenAgents \cite{xie2023openagents} advanced multi-agent collaboration, role specialization, and environmental awareness, endowing agent systems with greater adaptability and generalization. By 2025, the emergence of LRMs like DeepSeek R1 substantially improved agent system performance, enabling more complex workflows for multi-task, multi-tool, and multi-agent collaboration—officially ushering in the era of Agentic AI\cite{guo2025deepseek}. Overall, the Agentic stage propelled LAMs from information understanding to task execution and behavioral control, laying a crucial foundation for embodied intelligence and higher-level general intelligence.

\subsection{Related Survey Work}
Table \ref{tab:1} presents a comparative analysis between this tutorial and existing related survey studies. While current research primarily focuses on the role of LAMs in communication systems and partially explores their application potential in specific communication tasks, it still shows deficiencies in the detailed classification of LAMs' learning mechanisms as well as in the construction and application methods of Agentic AI systems. Although these survey studies have made valuable contributions to exploring the applications of LAMs and Agentic AI in communications, the following areas remain open for improvement:

\subsubsection{Lack of Detailed Taxonomy for LAMs and Their Training Paradigms}
Although existing studies have partially explored the applications of LAMs in communications, they still lack a detailed taxonomy of model types and training paradigms. In terms of model types, most research focuses on LLMs, while the applications of LVMs, LMMs, LRMs, and lightweight LAM in communications remain understudied, with no comprehensive classification or application adaptation framework established. Regarding training methodologies, while some studies discuss internal learning mechanisms (e.g., pretraining, fine-tuning, and alignment), they rarely address extrinsic learning mechanisms, such as Retrieval-Augmented Generation (RAG) and structured knowledge learning (e.g., Knowledge Graphs (KG)). The applicability, differences, and synergistic relationships of these learning strategies in communications still lack systematic comparison and analysis.

\subsubsection{Lack of Systematic Review on Agentic AI in Communications}
Current research primarily focuses on the perception and generation capabilities of LAMs in communication scenarios, whereas systematic discussions on Agentic AI equipped with long-term planning, autonomous decision-making, and tool invocation remain insufficient. In communication systems, Agentic AI holds broad application potential, particularly in complex interactive scenarios such as semantic communications, federated learning, network management and optimization, and Unmanned Aerial Vehicle (UAV) communication. However, most existing studies fail to clearly define its system architecture, core modules (e.g., planners, tools, memory modules), or integration pathways with communication knowledge. Additionally, there is a lack of modeling for multi-agent collaboration mechanisms tailored to communication tasks, hindering a comprehensive understanding of Agentic AI’s potential and value in advancing intelligent communication evolution.

\begin{table*}[t]
\centering
\scriptsize
\caption{Comparison of previous works with our tutorial.}
\label{tab:1}
\setlength{\tabcolsep}{3pt}
\renewcommand{\arraystretch}{2}
\resizebox{\linewidth}{!}{%
\begin{tabular}{
|m{0.7cm}<{\centering}|m{1.3cm}<{\centering}|m{1.4cm}<{\centering}|
m{0.9cm}<{\centering}|m{1.2cm}<{\centering}|m{1.5cm}<{\centering}|
m{1.3cm}<{\centering}|m{1.3cm}<{\centering}|m{1.2cm}<{\centering}|
m{1.5cm}<{\centering}|>{\RaggedRight}m{7cm}|}
\hline
\multirow{2}{*}{\textbf{\makecell{Year\\Ref.}}} & 
\multicolumn{5}{c|}{\textbf{LAMs}} & 
\multicolumn{4}{c|}{\textbf{Agentic AI}} & 
\multirow{2}{*}{\textbf{\makecell{Remarks}}} \\
\cline{2-10}
& \makecell{Components\\(C1)} & 
  \makecell{Classification\\(C2)} & 
  \makecell{Training\\(C3)} & 
  \makecell{Application\\(C4)} & 
  \makecell{Challenges \\(C5)} & 
  \makecell{Components\\(C6)} & 
  \makecell{Interactions\\(C7)} & 
  \makecell{Application\\(C8)} & 
  \makecell{Challenges  \\(C9)} & \\ 
\hline
2025 \cite{ferrag2025llm} & No & No & No & Limit & No & Yes & Limit & Limit & Limit & 
-For C1 to C3 and C5, it is not mentioned. \newline
-For C4, it briefly mentions applications without detailed LAM use cases. \newline
-For C7 to C9, it touches on agent frameworks and future directions but lacks depth and interaction details.
\\ \hline
2025 \cite{acharya2025agentic} & No & No & No & Limit & No & Yes & Limit & Limit & Yes & 
-For C1 to C3, C5, it is not mentioned. \newline
-For C4, it mentions applications but lacks depth. \newline
-For C7 and C8, it gives high-level interaction ideas without implementation. \newline
-For C9, it focuses on ethics, not technical aspects.\\ \hline
2025 \cite{gridach2025agentic} & No & Limit & No & Limit & No & Yes & Limit & Limit & Yes & 
-For C1, C3, C5, it is not mentioned. \newline
-For C2 and C4, it is mentioned, but the coverage is not extensive enough.\newline
-For C7 and C8, it introduces collaboration but lacks detail on interactions and roles. \\ \hline
2025 \cite{jiang2025comprehensive} & Yes & Yes & Yes & Yes & Yes & No & No & Limit & No & 
-For C6, C7, and C9, it is not discussed.\newline
-For C8, it is only briefly mentioned without detailed scenarios. \\ \hline
2024 \cite{zhou2024large} & Limit & Limit & Limit & Limit & Limit & No & No & No & No & 
-For C1 to C5, it covers LLMs broadly but lacks a systematic description of other LAMs. \newline
-For C6 to C9, Agentic AI is not included.\\ \hline
2025 \cite{boateng2025survey} & Limit & Limit & No & Limit & Yes & No & No & Limit & No & 
-For C1, C2, C4, and C8, it gives partial coverage but lacks detailed context. \newline
-For C3, C6, C7 and C9, it is not discussed. 
 \\ \hline
2024 \cite{chen2024big} & Limit & No & Limit & Limit & Yes & No & No & No & No & 
-For C1, C3, and C4, it discusses LAMs for 6G and deployment ideas, but lacks detailed classification and training methods.\newline
-For C2, classification of LAMs is not covered. \newline
-For C6 to C9, Agentic AI is not included. \\ \hline
2024 \cite{huang2024large} & Limit & Limit & Limit & Limit & Limit & No & No & No & No & 
-For C1 to C5, it provides a high-level overview of LLM in networking but there is a lack of description of other LAMs.  \newline
-For C6 to C9, Agentic AI is not included. \\ \hline
\multicolumn{2}{|c|}{\textbf{Our Tutorial}} & \textbf{Yes} & \textbf{Yes} & \textbf{Yes} & \textbf{Yes} & \textbf{Yes} & \textbf{Yes} & \textbf{Yes} & \textbf{Yes} & 
-For C1 to C9, this work offers a comprehensive and up-to-date overview of LAMs and Agentic AI, presenting a systematic framework and clear guidance for their development and application. \\ \hline
\end{tabular}%
}
\end{table*}

\subsection{Motivations and Contributions}
As wireless communication systems advance toward 6G, they face unprecedented challenges in required intelligence levels, dynamic adaptability, and system efficiency \cite{chowdhury20206g}. Leveraging their massive parameter scales (typically reaching billions or even trillions of parameters), emergent capabilities, and powerful cognitive reasoning, LAMs are progressively transforming the application landscape of AI in communications. These models demonstrate effective support for complex communication tasks, including semantic communication, resource scheduling, and network self-optimization\cite{jiang2025comprehensive}. In 6G networks, LAMs can fulfill the following functional roles \cite {jiang2024large3}:
\begin{enumerate}
\item \textbf{Data Generator:} As generative AI models with strong generalization and representation capabilities, LAMs can efficiently generate various types of communication data based on domain knowledge. By incorporating advanced generative architectures (e.g., autoregressive decoders, diffusion models), LAMs can synthesize high-quality Channel State Information (CSI) data to support critical tasks such as positioning estimation, bandwidth allocation, and network architecture design. Such synthetic data exhibits realism while preserving user anonymity, providing cost-effective and reliable data support for 6G network modeling, optimization, and deployment without compromising privacy \cite{jiang2025comprehensive}.

\item \textbf{Knowledge Organizer:} With powerful cross-modal semantic reasoning and knowledge integration capabilities, LAMs can structurally process and deeply mine raw communication data to enable automated knowledge extraction and reorganization. In semantic communication systems, LAMs can serve as knowledge bases to assist semantic encoding processes. Leveraging their extensive world knowledge and communication expertise, they effectively reduce ambiguity, improve semantic alignment quality, and enhance information transmission accuracy and contextual adaptability, thereby supporting the development of intelligent semantic representation and understanding frameworks \cite{jiang2024large2}.

\item \textbf{Resource Manager:} Through real-time perception and modeling of network environmental states, user behavior patterns, and resource utilization efficiency, LAMs facilitate intelligent scheduling and optimal allocation of communication resources. Integrated with Reinforcement Learning (RL) or long-chain reasoning frameworks, LAMs can dynamically formulate management decisions such as spectrum allocation, power control, and access strategies in multi-user, multi-service scenarios, thereby enhancing overall system efficiency and fairness. Additionally, LAMs can predict future resource demand trends, enabling proactive network planning and QoS assurance\cite{boateng2025survey}. 

\end{enumerate}

The key distinction between LAMs and Agentic AI lies in their working methods and intelligent decision-making capabilities. LAMs typically respond within fixed input patterns and generate outputs through known knowledge, but lack autonomy and adaptability. In contrast, Agentic AI possesses the ability for proactive decision-making and self-optimization, enabling it to make independent decisions in complex and dynamic environments while actively learning and adjusting during task execution. This allows Agentic AI to handle more complex communication tasks, particularly in dynamic environments, where it can respond in real-time and continuously optimize its behavior. In 6G communications, Agentic AI can play the following roles:
\begin{enumerate}
\item \textbf{Task Scheduler:} Agentic AI possesses the capability to comprehend complex instructions, allocate subtasks, and coordinate multi-module execution, enabling it to serve as the core task scheduler in complex communication scenarios. It can dynamically deploy different algorithmic modules and collaboratively generate solutions that meet task objectives. For instance, in multi-UAV cooperative communication scenarios, Agentic AI can autonomously plan service areas and flight paths, avoid obstacles, and allocate communication resources to establish stable links and provide computational support in emergency situations, significantly enhancing system autonomy and response efficiency \cite{10495599}.

\item \textbf{System Designer:} Leveraging its strong task comprehension and complex system logic modeling capabilities, Agentic AI can automatically design system architectures and configure modules based on the functional requirements of communication systems. In AI-integrated communication tasks, Agentic AI can combine knowledge from federated learning, resource scheduling, protocol stacks, and other domains to understand the design intent and operational mechanisms of algorithms such as FedAvg, autonomously completing system-level design and structural optimization. Through prompt tuning and policy feedback, it iteratively optimizes communication system performance, demonstrating highly intelligent system design potential \cite{10384606}.  

\item \textbf{Decision Executor:} Agentic AI not only exhibits rapid learning and adaptation in uncertain environments, showcasing autonomous strategic capabilities, but can also invoke traditional algorithms and external tools to demonstrate robust decision execution. In wireless network slicing management, Agentic AI can integrate multi-dimensional inputs (e.g., QoS, user demands, task priorities) and employ RL, causal reasoning, meta-learning, or game-theoretic strategies to make optimal decisions balancing performance, energy efficiency, and fairness. Additionally, Agentic AI can invoke external network simulation tools (e.g., NS-3 or OMNeT++) for simulation and validation, as well as call Software Defined Network (SDN) controllers to execute slice creation and scheduling\cite{wu2025agentic}.  
\end{enumerate}

This tutorial, set against the backdrop of the intelligent evolution of communication systems toward the 6G era, systematically reviews and thoroughly examines the pivotal roles of LAMs and Agentic AI in future intelligent communication systems. It offers a comprehensive overview from multiple perspectives, including model classification, training methodologies, Agentic AI system design, application scenarios, and research challenges. The main contributions of this work are summarized in the following five aspects.

\begin{enumerate}
\item \textbf{Systematic Review of Core Components and Model Classification in LAMs}: A comprehensive synthesis of core components, including Transformer, Vision Transformer (ViT), Variational AutoEncoder (VAE), Diffusion models, Diffusion Transformer (DiT), and Mixture of Experts (MoE), is provided, along with a classification and comparative analysis of mainstream model types such as LLMs, LVMs, LMMs, LRMs, and lightweight LAMs. This synthesis clarifies the respective applicability and technical potential of each model category in communication systems.

\item \textbf{Design of Datasets and Learning Mechanisms for Communication-specific LAMs}: Addressing challenges such as scarce domain knowledge, high task complexity, and diverse application requirements in communications, we propose a design framework for communication-specific LAMs. This framework features methodologies for constructing communication-relevant datasets, internal learning mechanisms encompassing pre-training, fine-tuning, and alignment, as well as external learning mechanisms including RAG and KG. These components collectively ensure model efficacy and usability in communication scenarios.

\item \textbf{Construction of LAM-based Agentic AI Framework From Communication Perspective}: A systematic integration of LAMs with Agentic AI technologies to establish a communication-oriented Agentic AI architecture. First, we identify core system components including LAMs, planners, knowledge bases, tools, and memory modules. Then, we examine interaction patterns for single-agent and multi-agent systems. Finally, we propose an integrated framework featuring multi-agent data retrieval, Multi-agent Collaborative Planning (MCP), and multi-agent evaluative reflection to support the intelligent processing of complex communication tasks.

\item \textbf{Exploration of LAM and Agentic AI Applications Across Communication Scenarios}: A systematic examination of LAM applications is conducted across critical domains such as semantic communication, IoT, edge intelligence, network design and management, security and privacy, and resource allocation. In parallel, we explore the potential of Agentic AI in wireless communication, semantic communication, network management and optimization, network security, and UAV communication, aiming to enhance overall system intelligence and operational efficiency.

\item \textbf{Identification of Challenges and Future Directions for LAM and Agentic AI in Communications}: For LAMs, we address data scarcity, inadequate reasoning, poor interpretability, and deployment difficulties, proposing solutions through autonomous continual learning, RL-driven reasoning training, model visualization, and model compression/distillation. For Agentic AI, we examine communication knowledge gaps, scalability limitations, complex control mechanisms, and evaluation challenges, suggesting future research on dynamic knowledge-guided Agentic RAG, distributed control architectures, unified control protocols, and process-oriented evaluation frameworks to enable the evolution from "LAM-driven" to "Agentic AI-driven" intelligent communication systems.

\end{enumerate}

\begin{figure*}[htpb]
		\centering
		\includegraphics[width=0.9\textwidth,height=0.77\textwidth]{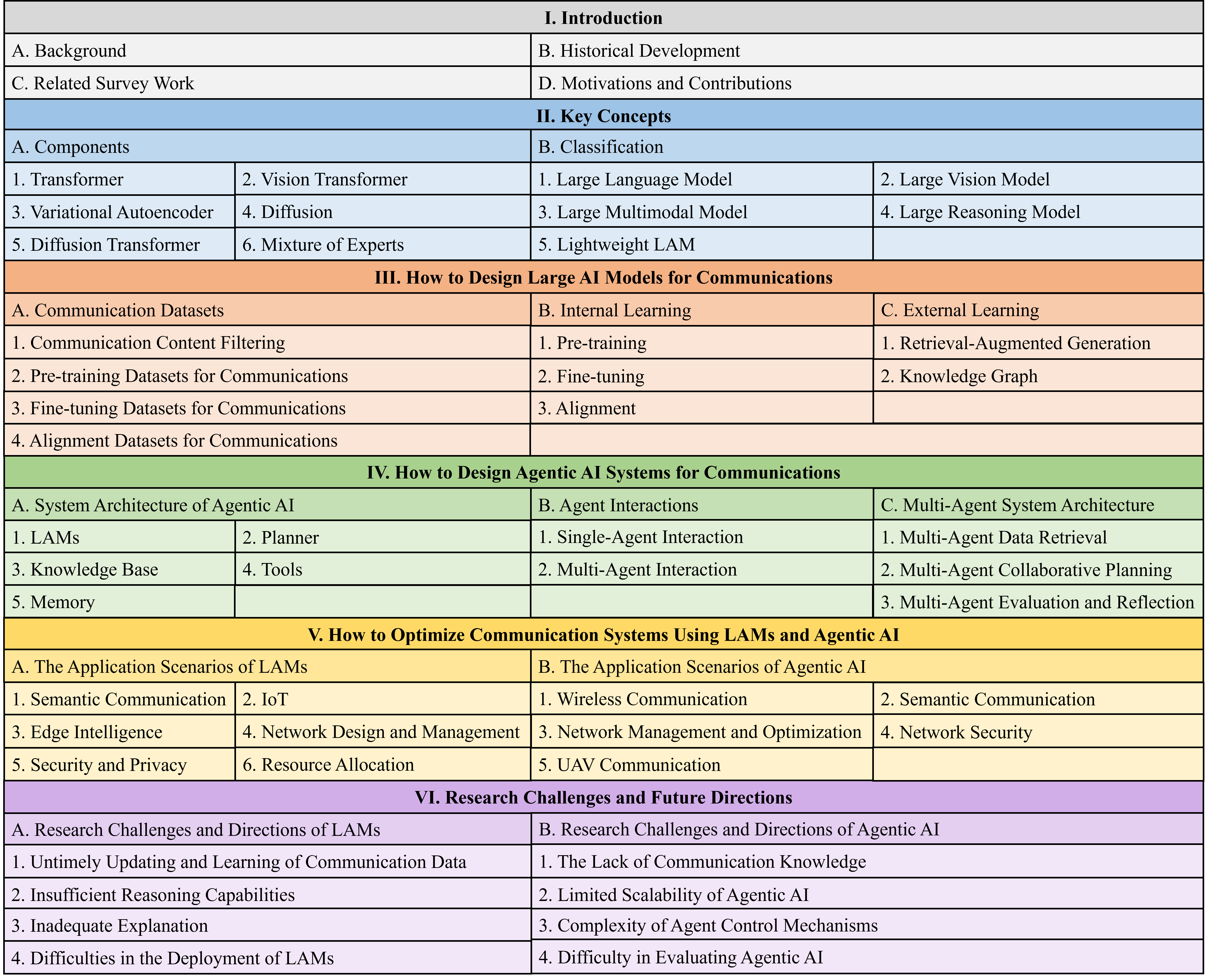}
		\caption{Overall organization of the tutorial.}
		\label{fig:a}
\end{figure*}

Fig.
\ref{fig:a} presents the organization of this tutorial. To ensure consistency in the formulation and model representation, this tutorial adopts the following unified symbolic notations, with definitions provided in Table \ref{tab:2}.

\begin{table}[htbp]
\centering
\caption{Acronyms and descriptions.}
\label{tab:2}
\small
\begin{tabularx}{\columnwidth}{ 
    >{\raggedright\arraybackslash}p{1.8cm}  
    >{\raggedright\arraybackslash}X  
    @{}  
} 
\toprule
\textbf{Acronym} & \textbf{Description} \\
\midrule
A2A & Agent-to-Agent Protocol \\
ACP & Agent Communication Protocol \\
AGI & Artificial General Intelligence \\ 
AI & Artificial Intelligence \\ 
CNN & Convolutional Neural Network \\
CoDi & Composable Diffusion \\ 
CoT & Chain of Thought \\ 
DiT & Diffusion Transformer \\ 
DPO & Direct Preference Optimization \\ 
FM & Foundation Model \\ 
FFN & Feed-Forward Network \\ 
GAN & Generative Adversarial Network \\ 
GQA & Grouped-Query Attention \\
GPT & Generative Pretrained Transformer \\ 
ICL & In-context Learning \\ 
KD & Knowledge Distillation \\ 
KG & Knowledge Graph \\ 
LAM & Large AI Model \\ 
LLM & Large Language Model \\ 
LLaMA & Large Language Model Meta AI \\ 
LMM & Large Multimodal Model \\ 
LLaVA & Large Language and Vision Assistant \\ 
LoRA & Low-Rank Adaptation \\ 
LRM & Large Reasoning Model \\ 
LVM & Large Vision Model \\ 
MAS & Multi-Agent System \\ 
MCP & Model Context Protocol \\
MHA & Multi-Head Attention \\
MoE & Mixture of Expert \\ 
MQA & Multi-Query Attention \\
PEFT & Parameter-Efficient Fine-Tuning \\ 
PPO & Proximal Policy Optimization \\ 
RAG & Retrieval-Augmented Generation \\ 
RL & Reinforcement Learning \\
RLHF & Reinforcement Learning from Human Feedback \\ 
SAM & Segment Anything Model \\ 
SFT & Supervised Fine-Tuning \\ 
SLM & Small Language Model \\ 
UAV & Unmanned Aerial Vehicle \\
VAE & Variational Autoencoder \\
ViT & Vision Transformer \\\bottomrule
\end{tabularx}
\end{table}

\section{Key Concepts}
\subsection{Components}
\subsubsection{Transformer}
The Transformer is a novel neural network architecture proposed by Google in 2017 \cite{vaswani2017attention}. Its core innovation relies entirely on the self-attention mechanism to capture dependencies within the input sequence, and the cross-attention mechanism to connect the encoder and decoder.

Self-attention is a key technique in the Transformer architecture. It enables the model to consider all other words (or tokens) in the sequence when processing a particular word, computing weighted representations based on relevance. Additionally, Google introduced multi-head self-attention to compute attention in parallel, allowing the model to learn information from different representation subspaces\cite{vaswani2017attention}. The computation of self-attention is given by the following formula:

\begin{equation}
\text{Attention}(Q, K, V) = \mathrm{softmax} \left( \frac{QK^\top}{\sqrt{d_k}} \right) V,
\end{equation}
where $Q$, $K$, and $V$ denote the Query, Key, and Value matrices, respectively; $d_k$ represents the dimensionality of the Key vectors; and $\mathrm{softmax}(\cdot)$ is the normalization function.

In addition, each layer within the encoder-decoder architecture of the Transformer includes a Feed-Forward Network (FFN), Layer Normalization (LayerNorm), and residual connections. These design choices significantly enhance the Transformer’s performance in modeling long-range dependencies, facilitating gradient propagation, and enabling efficient parallel training, outperforming traditional neural network architectures in these aspects.

The Transformer offers strong capabilities in modeling long-range dependencies and supports highly parallelized computation. However, a key limitation lies in the quadratic computational and memory complexity of the self-attention mechanism with respect to the sequence length $n$, i.e., $O\left(n^2\right)$, which constrains its ability to handle long-sequence data efficiently.

The Transformer is the cornerstone of LLMs. Its exceptional parallel computation capabilities and ability to capture long-range dependencies have enabled the scaling of models to tens or even hundreds of billions of parameters, as seen in models such as GPT and the LLaMA series. In communications, the Transformer has been widely applied to a range of tasks, including semantic communication \cite{yang2023witt}, signal processing \cite{leng2025unveiling}, multimodal perception \cite{tian2023multimodal}, and resource management \cite{zhang2025decision}, significantly advancing the level of intelligence in communication systems.

\subsubsection{ViT}
ViT was the first to demonstrate that a pure Transformer architecture can be directly applied to image recognition in 2020, achieving performance on large-scale datasets that matches or even surpasses that of state-of-the-art Convolutional Neural Networks (CNNs) \cite{dosovitskiy2020image}.

ViT first divides an image into fixed-size patches, linearly embeds these patches, adds positional encodings, and then feeds them into a standard Transformer encoder in the same way as word sequences are processed. Various visual tasks are then performed through different output layers, as illustrated below:
\begin{multline}
y = \mathrm{Encoder} \left( 
\mathrm{Concat}(\bm{z}_{\text{cls}},\mathrm{Flatten}(\mathrm{Patch}(\bm{I}))) + \bm{E}_{\text{pos}} 
\right),
\end{multline}
where $\bm{I}$ denotes the input image, $\text{Patch}(\cdot)$ and $\text{Flatten}(\cdot)$ refer to the patching and flattening operations, respectively, while $\text{Concat}(\cdot)$ represents the concatenation operation. $z_{\text{cls}}$ and $E_{\text{pos}}$ denote the CLS token and positional encoding, respectively, and $\text{Encoder}(\cdot)$ refers to the Transformer encoder.

The advantages of ViT lie in its strong capability for global information modeling and scalability, which aligns well with the extensibility of the Transformer architecture. However, ViT lacks certain inherent visual inductive biases present in traditional models, such as locality and translational invariance, which often necessitates pretraining on large-scale datasets to achieve competitive performance. Moreover, when handling high-resolution images, the increased sequence length leads to substantial computational overhead.

ViT has become one of the foundational architectures for LVMs (e.g., SAM, DINO) and serves as a critical component in many LMMs. In communications, ViT has been widely applied to tasks such as semantic communication \cite{mohsin2025vision}, line-of-sight blockage prediction \cite{gharsallah2024vit}, and automatic modulation recognition \cite{zheng2024fe}. Leveraging its powerful modeling capabilities, ViT enhances the system’s perception accuracy of spatially structured data and improves communication efficiency.

\subsubsection{VAE}
The VAE is a deep generative model based on variational Bayesian methods, integrating the architecture of autoencoders with the principles of probabilistic graphical models \cite{kingma2013auto}.

The VAE learns to encode input data into a low-dimensional latent space and enables sampling from this space to generate new data. Unlike standard autoencoders, VAEs learn a probabilistic distribution over the latent space, which facilitates the generation of diverse outputs. Specifically, the VAE encodes an input $x$ into a latent distribution $q_\phi\left(z \middle| x\right)$, typically assumed to be a multivariate Gaussian distribution, samples $z$ from this distribution, and then reconstructs an approximation of $x$ through $p_\theta\left(x \middle| z\right)$. The training objective is to maximize the Evidence Lower Bound (ELBO):
\begin{equation}
\mathcal{L} = \mathbb{E}_{q_\phi(z \mid x)} \left[ \log p_\theta(x \mid z) \right]
- \mathrm{KL}\left( q_\phi(z \mid x) \,\|\, p(z) \right),
\end{equation}
where \textcolor{black}{$\mathbb{E}$ denotes the expectation}, the reconstruction term $\mathbb{E}_{q\phi(z \mid x)}\left[\log p_\theta(x \mid z)\right]$ ensures the quality of the generated outputs, while the Kullback-Leibler term $\mathrm{KL}\left(q_\phi(z \mid x),|, p(z)\right)$ guides the latent space to align with the prior distribution $p(z)$, typically assumed to be a standard normal distribution.

The VAE offers a generative framework with a solid theoretical foundation. Its learned latent space is typically smooth, making it amenable to interpolation and interpretation. However, a common limitation is that the generated samples may appear blurry, and the ELBO represents only a lower bound of the true data likelihood.

VAE is commonly used for discrete image encoding and data compression, forming the core conceptual foundation of LVMs such as DALL-E and Stable Diffusion. As a powerful generative model, VAE has been widely applied in communication tasks such as CSI feedback \cite{hussien2022prvnet}, semantic communication \cite{bo2024joint}, and Multiple-Input Multiple-Output (MIMO) detection \cite{omondi2023variational}. By leveraging its latent variable modeling capability, VAE effectively enhances the operational efficiency of communication systems.

\subsubsection{Diffusion}
Diffusion models are probabilistic generative models based on Markov processes, proposed in 2020\cite{sohl2015deep}. Their core idea is to model data distributions through a "noise addition–denoising" process.

Diffusion models operate through two key processes: the forward diffusion process, which gradually adds Gaussian noise to the data (e.g., images, videos) until it becomes pure noise; and the reverse denoising process, which starts from pure noise and progressively removes the noise using a neural network to generate clear data samples.

\textbf{Forward Diffusion Process}: Noise is gradually added to the original sample $x_0$ over multiple steps until it becomes nearly Gaussian white noise. The $t$-th step can be expressed as:
\begin{equation}
q(x_t \mid x_{t-1}) = \mathcal{N} \left( 
\sqrt{1 - \beta_t} \, x_{t-1},\; \beta_t \, \mathbf{I}
\right),
\end{equation}
where \textcolor{black}{$q(x_t \mid x_{t-1})$ is a conditional probability distribution that defines the probability of the current noisy sample $x_t$ given the previous step’s noisy sample $x_{t-1}$}, $\beta_t$ is a noise variance parameter initialized to a small value, and \textcolor{black}{$\mathbf{I}$ is the identity matrix}, and $x_t$ and $x_{t-1}$ denote the images at step $t$ and step $t-1$, respectively.

\textbf{Reverse Denoising Process}: A neural network $\epsilon_\theta(x_t, t)$ is trained to estimate the noise component, and the image at step $t-1$ is updated using a fixed variance and the learned mean:
\begin{equation}
x_{t-1} = \frac{1}{\sqrt{1 - \beta_t}} \left( x_t - \beta_t \, \epsilon_\theta(x_t, t) \right)
+ \sigma_t z, \quad z \sim \mathcal{N}(0, \mathbf{I}),
\end{equation}
where $\sigma_t z$ represents noise with variance $\sigma_t$, which is used to maintain the diversity and stochasticity of the generated samples.

Diffusion models offer the advantages of generating high-quality and diverse samples, along with relatively stable training. However, their main drawbacks include slow generation speed due to the need for multi-step iterative sampling and relatively complex theoretical derivation.

Diffusion models have emerged as the dominant technology for high-quality image and video generation. LVMs such as Stable Diffusion \cite{rombach2022high}, DALL-E 2/3 \cite{ramesh2022hierarchical}\cite{betker2023improving}, and Imagen \cite{saharia2022photorealistic} are all based on the principles of diffusion models. In communications, diffusion models have been widely applied to tasks such as channel estimation \cite{fu2025conditional}, semantic communication \cite{grassucci2024diffusion}, channel enhancement \cite{zeng2024dmce}, and signal enhancement \cite{xu2024diffusion}, demonstrating high-fidelity generation and strong robustness under complex channel conditions.

\subsubsection{DiT}
DiT, proposed in 2022, is a specialized design that applies the Transformer architecture to diffusion models \cite{peebles2023scalable}. It typically replaces the previously common U-Net structure with a Transformer during the reverse denoising process of the diffusion model to predict noise\cite{peebles2023scalable}.

DiT maps the latent image representation $z_t$, the timestep encoding $t$, and an optional condition $y$ into a sequence of embeddings, which are then fed into a standard Transformer for global self-attention and feedforward processing. The final output is either the predicted noise $\epsilon_\theta(x_t, t, y)$ or a direct prediction of the denoised image sample, as illustrated below:
\begin{equation}
h = \mathrm{Encoder}\left( \mathrm{Embed}(z_t, t, y) \right),
\end{equation}
\begin{equation}
\epsilon_\theta(x_t, t, y) = \mathrm{Project}(h),
\end{equation}
where $\mathrm{Embed}(\cdot)$ denotes the embedding matrix, $\mathrm{Encoder}(\cdot)$ represents the Transformer encoder, and $\mathrm{Project}(\cdot)$ is the projection matrix used to output the predicted noise, $x_t$ denotes the noisy image at the current timestep during the denoising process.

DiT offers excellent scalability, enabling significant improvements in generation quality by increasing model size. However, it remains constrained by the inherently slow sampling speed of diffusion models, and the computational overhead of Transformers for high-dimensional inputs remains substantial.

The introduction of DiT represents a significant milestone in the development of diffusion models, demonstrating the powerful potential of the Transformer architecture in generative tasks. It has inspired the design of numerous subsequent large generative models, particularly world models such as OpenAI's Sora \cite{liu2024sora}, which also adopt the DiT architecture at their core to process spatiotemporal latent representation blocks. DiT has proven that the Transformer can serve as a universal and scalable backbone for a wide range of complex generative modeling tasks.

\subsubsection{MoE}
MoE is a model architecture paradigm designed to enhance model capacity through conditional computation while maintaining manageable computational costs \cite{jacobs1991adaptive}. It is not a standalone model but is typically integrated within specific LAMs.

In a standard Transformer module, the original FFN sublayer typically consists of two linear transformations and a non-linear activation function, applied independently to the representation of each position (token) in the output of the self-attention layer. To replace it with an MoE layer, one must first instantiate $N$ independent "expert" networks, each of which is itself an FFN with the same architecture as the original but with its own set of parameters. Additionally, a gating network is introduced to compute a probability score for each expert. Based on these scores, a sparse routing strategy is usually employed to select the top-$K$ scoring experts (with $K$ typically being small, such as 1 or 2) to process the current token. Finally, the outputs of the selected $K$ experts are combined through a weighted summation to produce the final MoE output for that token, as illustrated below:
\begin{equation}
g = \mathrm{softmax}\left(W_g x + b_g\right),
\end{equation}
\begin{equation}
y_i = \mathrm{Expert}_i(x), \quad \forall i \in \mathcal{I}_\mathcal{K},
\end{equation}
\begin{equation}
\mathrm{output} = \sum_{i \in \mathcal{I}_\mathcal{K}} g_i \cdot y_i,
\end{equation}
where $W_g$ and $b_g$ denote the weight matrix and bias vector of the gating network, respectively, and $\mathrm{softmax}(\cdot)$ is the normalization function. $\mathcal{I}_\mathcal{K}$ represents the index set of the top-$K$ experts with the highest scores; $\mathrm{Expert}_i(\cdot)$ denotes the $i$-th expert network; $y_i$ is its corresponding output; and $g_i$ is the score assigned to the $i$-th expert.

The key advantage of MoE lies in its effective decoupling of parameter scale from computational cost, enabling the training and deployment of LAMs that significantly exceed the size of dense models with comparable computational budgets. However, a major drawback is its substantial memory requirement: despite the sparsity in computation, all expert parameters must still be loaded into memory during inference. As a result, MoE models typically consume significantly more memory than dense models with equivalent computational complexity.

MoE is one of the key technologies enabling the development of state-of-the-art LLMs. Architectures such as Google’s GLaM\cite{du2022glam}, Mistral AI’s Mixtral 8x7B\cite{jiang2024mixtral}, and GPT-4\cite{achiam2023gpt} have all adopted the MoE framework. By leveraging MoE, these models can scale to larger parameter sizes while maintaining acceptable training and inference costs. In communications, MoE has been widely applied to tasks such as communication security \cite{zhao2025enhancing}, satellite communications \cite{zhang2024generative2}, and signal processing \cite{gao2023moe}, where the expert activation and parallel processing mechanisms contribute to enhanced system intelligence, improved inference efficiency, and greater robustness.

\subsection{Classification}

\subsubsection{LLM}
LLMs represent a significant branch in the field of deep learning, referring specifically to neural network models that are pretrained on massive text corpora and contain an extremely large number of parameters, typically in the tens or even hundreds of billions. Their core capability lies in understanding and generating human-like natural language, enabling them to perform a wide range of language-related tasks with remarkable generalization and adaptability. By learning grammar, semantics, and commonsense knowledge from large-scale corpora, LLMs have acquired human-level cognitive and reasoning abilities, establishing themselves as a foundational technology driving advancements in NLP and the broader field of AI.

Structurally, most state-of-the-art LLMs are built upon the Transformer decoder-only architecture, particularly leveraging its self-attention mechanism. This mechanism enables the model to effectively capture long-range dependencies within the input sequence. A typical Transformer building block consists of a multi-head self-attention layer and a FFN layer, combined with residual connections and LayerNorm to stabilize the training process. By stacking multiple such blocks, LLMs are capable of constructing deep networks that learn complex patterns and representations in language data, ranging from low-level features to high-level abstractions.

The GPT series developed by OpenAI (e.g., GPT-3 \cite{brown2020language}, ChatGPT \cite{achiam2023gpt}), the Gemma series by Google \cite{team2023gemini}\cite{team2024gemini}, the LLaMA series by Meta AI \cite{touvron2023llama}\cite{touvron2023llama2}\cite{dubey2024llama}, and Claude by Anthropic are all prominent representatives of LLMs. In communications, LLMs have been widely applied to tasks such as semantic communication \cite{jiang2024large5}, network management \cite{boateng2025survey}, multi-agent systems \cite{jiang2024large3}, and communication security \cite{yao2024survey}. Leveraging their powerful language understanding and generation capabilities, LLMs significantly enhance the intelligence, adaptability, and interaction efficiency of communication systems.

\subsubsection{LVM}
LVMs refer to deep neural network models trained on massive visual datasets consisting of billions of images and videos, and characterized by an extremely large number of parameters. These models are designed to learn general and powerful visual representations, enabling them to understand complex image and video content and generalize across a wide range of downstream vision tasks. LVMs significantly expand the performance ceiling and application scope of computer vision systems.

LVMs typically adopt CNNs or ViTs as their backbone architectures. These architectures extract visual features in a hierarchical manner by stacking multiple processing layers, progressively capturing representations ranging from low-level features (such as edges and textures) to high-level semantics (such as object parts and complete objects). In addition to pure Transformer-based designs, hybrid architectures that combine CNNs and Transformers are also common in LVM development, aiming to leverage the local feature extraction strength of CNNs and the long-range dependency modeling capabilities of Transformers.

Representative LVMs include Masked Autoencoders (MAE) \cite{he2022masked} , DINO \cite{oquab2023dinov2}, and SAM \cite{kirillov2023segment}. Their applications in communications are primarily focused on semantic communication \cite{jiang2024large2}\cite{tariq2023segment}\cite{jiang2025lightweight}. By incorporating LVMs such as SAM and MAE, lightweight knowledge bases and efficient semantic encoders/decoders are constructed, enabling the compression and sharing of image semantic information. This significantly improves communication efficiency and image reconstruction quality.

\subsubsection{LMM}
LMMs are designed to jointly process and understand data from multiple distinct modalities, such as text, images, audio, video, and potentially even code, point clouds, and sensor data\cite{huang2024large}. The goal of LMMs is to enable comprehensive modality fusion and interaction, thereby more effectively simulating how humans perceive and understand the world through various inputs. These models are capable of performing complex cross-modal reasoning, content generation, and seamless interaction. LMMs are widely regarded as a critical step toward achieving more general forms of AI.

In terms of architectural design, LMMs are typically built upon powerful unimodal backbones and incorporate more sophisticated cross-modal fusion and alignment mechanisms. Their core architecture often includes: modality-specific encoders for different input types; one or more projection modules that map modality-specific information into a shared representation space and perform deep alignment, potentially using multi-layer cross-attention, modality gating mechanisms, or dedicated fusion networks; and a central processing unit, usually based on LLMs, responsible for semantic understanding, reasoning, and task execution. To support multimodal output generation, the LMM may also integrate corresponding decoders.

OpenAI’s GPT-4 is a prominent representative of LMMs, natively supporting flexible combinations of text and image modalities for both input and output. It significantly reduces interaction latency and enhances performance on cross-modal tasks, demonstrating truly real-time multimodal dialogue capabilities. Similarly, Google’s Gemini series \cite{team2023gemini}\cite{team2024gemini} and the LLaVA series \cite{liu2024visual}\cite{zhang2024llava}\cite{guo2024llava} are also powerful LMMs. By integrating a broader range of sensory inputs, these models enable unprecedented levels of complex cross-modal understanding and generation, further blurring the boundary between digital intelligence and physical-world perception.

LMMs have been widely applied in communication scenarios such as semantic communication \cite{jiang2024large5}\cite{qiao2024latency}, wireless network intent management \cite{xu2024large1}, and multimodal task-oriented dialogue systems \cite{kawamoto2023application}. By integrating multimodal information such as images, text, and sensor data, these models significantly enhance the communication system’s capabilities in understanding, adaptability, and intelligent interaction.

\subsubsection{LRM}
LRMs are AI models focused on enhancing systematic reasoning capabilities in complex tasks. Their primary goal is to solve complex logical problems, such as those encountered in mathematics, programming, and scientific domains, through explicit multi-step logical reasoning. Compared to conventional LLMs, LRMs significantly improve their performance in planning, problem decomposition, and dynamic knowledge integration by incorporating techniques such as RL \cite{ghasemi2025comprehensivesurveyreinforcementlearning}, Supervised Fine-Tuning (SFT) \cite{zhang2023instruction}, and Chain-of-Thought (CoT) \cite{wei2022chain}.

Structurally, LRMs typically adopt a multi-stage training framework. Starting from a base pretrained model, they optimize the generation of reasoning chains through RL or hybrid training strategies (e.g., SFT + RL), and dynamically augment external knowledge using RAG mechanisms \cite{lewis2020retrieval}. For instance, DeepSeek R1 \cite{guo2025deepseek} employs the Group Relative Policy Optimization (GRPO) algorithm \cite{shao2024deepseekmath}, which balances answer accuracy and reasoning format consistency through a reward function. Additionally, it incorporates cold-start data and language consistency constraints, enabling the model to maintain high-quality reasoning while reducing redundant computation and optimizing resource consumption.

Representative LRMs include models such as OpenAI-o1, DeepSeek R1, and Qwen-QwQ \cite{yang2024qwen2}. As an early benchmark, OpenAI-o1 demonstrated complex reasoning capabilities through Monte Carlo Tree Search (MCTS) \cite{coulom2006efficient} and Procedural Reward Modeling (PRM). However, its closed-source nature and high computational cost have limited its widespread adoption. In contrast, DeepSeek R1 \cite{guo2025deepseek} stands out for its open-source availability, offering reasoning performance comparable to o1 but at a significantly lower cost, thereby greatly advancing the open-source community. Additionally, LRMs such as Qwen-QwQ \cite{yang2024qwen2} have exhibited reasoning abilities on par with o1-like models in domain-specific tasks such as code generation, further enriching the diversity of the LRM ecosystem.
In communications, LRMs are primarily applied to enhance system intelligence, adaptability, and security \cite{qu2025survey} by leveraging their powerful multi-step reasoning and abstraction capabilities.

\subsubsection{Lightweight LAM}
Lightweight LAMs refer to those LAMs that are specifically designed and optimized to reduce model complexity, minimize storage size, lower computational resource consumption, and accelerate inference speed. The core value of such models lies in their ability to operate efficiently in resource-constrained environments, such as mobile devices, embedded systems, IoT devices, and edge computing nodes. Additionally, they contribute to reducing the cost and latency of large-scale cloud deployments, thereby enabling LAMs to be more broadly and economically applied in various real-time or power-sensitive scenarios.

Lightweight LAMs require carefully optimized architectural designs and trade-offs to maximize performance within constrained resource “budgets.” Structural lightweight is typically reflected in several aspects: first, by directly reducing model depth (number of layers) and width (hidden dimensions, number of attention heads); second, by adopting more efficient component variants, such as Grouped-Query Attention (GQA)\cite{ainslie2023gqa} or Multi-Query Attention (MQA)\cite{shazeer2019fast} in place of standard Multi-Head Attention (MHA), which significantly reduces the required cache size during inference and improves efficiency. 

A series of representative lightweight LAMs have recently demonstrated their potential and value. For example, the TinyLLaMA project aims to replicate the architecture and training pipeline of LLaMA 2 with approximately 1.1 billion parameters, incorporating optimizations such as GQA to provide a foundational model for extremely resource-constrained research and development \cite{zhang2024tinyllama}. LiteLLaMA typically refers to official or community-optimized versions of the LLaMA series with fewer parameters, emphasizing a balance between performance and resource consumption \cite{xu2025evaluating}. MiniCPM is a multimodal model designed for edge deployment, with around 2 billion parameters, serving as a representative model tailored for mobile and edge devices\cite{hu2024minicpm}. In addition, Microsoft’s Phi series (e.g., Phi-2\cite{javaheripi2023phi}, Phi-3-mini/small/medium\cite{abdin2024phi}) is known for its “small size, high performance” characteristics, achieving results that surpass models of similar scale through training on high-quality datasets. The continual advancement of these lightweight LAMs is enabling powerful AI capabilities to be brought to a wider range of terminal devices and application scenarios.
In communications, lightweight LAMs play a critical role in semantic communications \cite{jiang2025lightweight}, where their low computational overhead and strong inference capabilities allow for efficient semantic information extraction, compression, and reconstruction directly on edge devices, significantly enhancing both communication efficiency and robustness.

The classification of LAMs and their corresponding application scenarios in communications are presented in Table \ref{tab:Classification}.

\begin{table*}[htbp]
\scriptsize
\caption{Classification of LAMs and their applications in communications.}
\label{tab:Classification}
\renewcommand{\arraystretch}{1.5}
\setlength{\tabcolsep}{6pt}
\begin{tabular}{|>{\raggedright\arraybackslash}p{2.6cm}|
                >{\raggedright\arraybackslash}p{4cm}|
                >{\raggedright\arraybackslash}p{3cm}|
                >{\raggedright\arraybackslash}p{6cm}|}
\hline
\textbf{LAM Category} & 
\textbf{Components} & 
\textbf{Specific Models} & 
\textbf{Application Scenarios} \\
\hline
Large Language Model & Transformer, MoE & GPT series, Gemma series, LLaMA series & Semantic Communication \cite{jiang2024large5}, Network Management \cite{boateng2025survey}, Multi-agent Systems \cite{jiang2024large3}, Communication Security \cite{yao2024survey} \\
\hline
Large Vision Model & ViT, Diffusion, DiT, MoE  & SAM series, DINOv2, MAE & Semantic Communication \cite{jiang2024large2} \\
\hline
Large Multimodal Model & Transformer, ViT, VAE, Diffusion, DiT, MoE & GPT-4o, Gemini series, LLaVA series & Semantic Communication \cite{jiang2024large5}, Intent Network Management \cite{xu2024large1}, Multimodal Task Dialogue \cite{kawamoto2023application} \\
\hline
Large Reasoning Model & Transformer, MoE & OpenAI o1, DeepSeek R1, Qwen-QwQ & Network Management \cite{qu2025survey} \\
\hline
Lightweight LAM & Transformer, ViT & TinyLlama, LiteLlama, MiniCPM, Phi series & Semantic Communication \cite{jiang2025lightweight} \\
\hline
\end{tabular}
\end{table*}

\subsection{Summary and Lessons Learned}
\subsubsection{Summary}
This chapter provides a systematic summary of the core components and model types involved in LAMs for communications. We detail the fundamental principles, architectural characteristics, and application scenarios in communications for typical modules such as Transformer, ViT, VAE, Diffusion, DiT, and MoE. Meanwhile, we review the definitions, structural features, and representative works of different categories of LAMs including LLM, LVM, LMM, LRM, and lightweight LAMs. This lays a comprehensive foundation for readers to understand the current technology system of LAMs.

\subsubsection{Lessons Learned}
\textcolor{black}{Although significant progress has been made in the model architectures, classification mechanisms, and communication applications of LAMs, multiple challenges remain. On one hand, models like Transformer, ViT, and Diffusion require heavy computation when processing long sequences and high-dimensional data, diffusion models have slow generation speeds, and MoE structures consume high memory, all of which affect the real-time performance and deployment efficiency of LAMs in communication scenarios. On the other hand, the stability and consistency of LAMs in multimodal fusion and complex reasoning tasks still need improvement. Future research should focus on optimizing model computation structures, enhancing sampling and reasoning efficiency, constructing unified and scalable multimodal model architectures, and promoting efficient deployment of lightweight LAMs in edge environments \cite{chen2024big}.}

\section{How to Design Large AI Models for Communications}

In the context of learning communication knowledge, LAMs primarily adopt two approaches. The first approach involves embedding communication knowledge directly into the model parameters through pre-training, fine-tuning, and alignment. However, this method is time-consuming and less suitable for knowledge that requires frequent updates. The second approach combines RAG and KG, leveraging external vector databases and graph databases to provide contextualized knowledge to LAMs without modifying their parameters. This approach is more adaptive to the demands of rapidly evolving knowledge. In the following sections, we first introduce the methodologies for constructing communication datasets, followed by a detailed discussion of both learning paradigms.
\subsection{Communication Datasets}
The construction of communication datasets serves as the foundation for training LAMs in communications. It encompasses three key components: pre-training, fine-tuning, and alignment datasets. The primary goal is to support the transition and enhancement of model capabilities from general-purpose intelligence to task-specific competence in communication tasks.

\subsubsection{Communication Content Filtering}
Currently, general-purpose datasets used for training LAMs contain substantial communication-related content. Representative datasets include Common Crawl\footnote{http://commoncrawl.org/the-data/get-started/}, Pile\footnote{https://github.com/togethercomputer/}, Dolma\footnote{https://huggingface.co/datasets/allenai/dolma}, and RedPajama-Data\footnote{https://github.com/togethercomputer/RedPajama-Data}. Accordingly, domain-specific communication datasets can be constructed by extracting relevant content from these sources. This process involves identifying a set of commonly used communication-related keywords, filtering the datasets based on these keywords to retain communication-relevant content, and applying deduplication techniques to remove redundant or repetitive entries. These steps aim to enhance the training efficiency of LAMs while maintaining data diversity. During content filtering, communication-specific keywords can be precisely selected based on the following criteria\textcolor{black}{\cite{zou2024telecomgpt}}:
\begin{itemize}
    \item \textbf{Technical Relevance: }Keywords should be closely associated with core communication theories. For example, “6G” represents the latest generation of mobile communication technologies, and “VoIP” refers to voice communication over IP networks, both indicating clear communication-specific contexts.
    \item \textbf{High Frequency: }Keywords should be commonly used terms in professional communication literature. For instance, “Broadband” describes high-speed internet access technologies, and “LTE,” an abbreviation for long-term evolution, frequently appears in mobile communication scenarios.
    \item \textbf{Uniqueness: }Keywords should possess distinctiveness and specificity in communications. Terms like “spectrum allocation,” a key concept in wireless communication, and “fiber-optic communication,” the foundation of modern high-speed transmission, exemplify such uniqueness.
    \item \textbf{Authoritativeness: }Keywords should originate from core communication standards and carry authoritative significance. For example, “3GPP” defines global mobile communication standards, and “IEEE 802.11” specifies wireless LAN standards, both accurately referencing formal communication technologies.
    \item \textbf{Timeliness: }Keywords should reflect cutting-edge trends in communications. “Network slicing” allows flexible allocation of network resources, while “quantum encryption,” based on quantum mechanics, enables secure communication, both representing advanced and emerging concepts.
    \item \textbf{Clarity: }Priority should be given to keywords that precisely describe frontier communication technologies, avoiding vague or overly generic terms. Examples include “VoLTE” for high-definition voice services over LTE networks, and “Non-Terrestrial Networks” refer to satellite communications and other non-ground-based networking technologies.
\end{itemize}

\subsubsection{Pre-training Datasets for Communications }
The pre-training of LAMs requires vast and diverse communication-related data that spans multiple technical domains and data sources to comprehensively acquire domain-specific knowledge and enhance the model's generalization and accuracy. The following are representative datasets that can be utilized for pre-training LAMs in communications:

\begin{itemize}
\item \textbf{TSpec-LLM} \cite{nikbakht2024tspec} is an open-source dataset targeting 3GPP documents, covering over 30,000 documents from 1999 to 2023 with a total size of 13.5 GB. The dataset is formatted in markdown while preserving the original structural hierarchy, facilitating comprehension and processing by LLMs. 

\item \textbf{OpenTelecom Dataset} \cite{zou2024telecomgpt} is a pre-training corpus designed for LAMs in telecommunications. It encompasses a wide range of sources, including communication standards, research papers, books, and patents, with a particular emphasis on authoritative documents from 3GPP and IEEE. This dataset ensures that models acquire comprehensive and credible telecommunications knowledge, providing rich and high-quality textual resources for communication-related NLP tasks.

\item \textbf{CommData-PT} \cite{jiang2025commgpt} is a high-quality pre-training corpus specifically constructed for LAMs in communications. It includes comprehensive content from 3GPP standards, IEEE protocols, communication-related patents, academic papers, source code, and Wikipedia entries, covering knowledge across all layers of communication systems. The data are processed through LAM-assisted recognition, keyword-based filtering, cleaning, and standardization, resulting in a corpus with high domain specificity and structural consistency. This dataset provides strong support for both the pre-training and downstream task performance of communication-oriented LAMs.

\end{itemize}

\subsubsection{Fine-tuning Datasets for Communications}
In communication, fine-tuning enables LAMs to learn and understand domain-specific instructions and perform corresponding tasks. Instruction-tuning datasets play a crucial role in this process by providing paired instruction-response samples, allowing the model to learn how to execute tasks based on given instructions. This significantly enhances the model's adaptability and accuracy in specific communication tasks. The following are several representative instruction-tuning datasets:

\begin{itemize}
\item \textbf{TelecomInstruct Dataset} \cite{zou2024telecomgpt} is an instruction-tuning dataset tailored for telecommunication tasks. It covers a wide range of task types, including question answering, document classification, code generation, and protocol interpretation. This dataset is designed to enhance the model’s ability to comprehend and execute telecom-specific instructions, thereby improving its effectiveness and generalization in complex communication scenarios.

\item \textbf{CommData-FT} \cite{jiang2025commgpt} is a high-quality instruction-tuning dataset specifically designed for fine-tuning LAMs in communications. Built upon the CommData-PT corpus, it contains well-structured instruction-response pairs covering tasks such as protocol-related question answering, document classification, and text summarization. The dataset adheres to standardized formatting and is manually curated to ensure data quality, providing strong support for model fine-tuning and specialization in communication tasks.

\end{itemize}

\subsubsection{Alignment Datasets for Communications }
Alignment datasets provide high-quality feedback to LAMs to optimize their behaviors and decision-making processes, enabling the models to better align with human preferences and values in complex environments. For example, the \textbf{TelecomAlign dataset} \cite{zou2024telecomgpt} is specifically designed for alignment fine-tuning in telecommunications. It aims to train LLMs to generate responses that better meet communication-specific requirements and human preferences while minimizing redundancy, verbosity, and irrelevant content. By favoring concise and accurate answers, this dataset helps reduce system latency and aligns with the principles of semantic communication, thereby enhancing the model's practicality and human-AI collaboration capabilities in communication scenarios.

\subsection{Internal Learning} 

Internal learning in communications typically involves three stages: pre-training, fine-tuning, and alignment. Pre-training enables the model to acquire general semantic and knowledge capabilities from large-scale data; fine-tuning adapts the model to the specific requirements of communication tasks; and alignment further refines the model's output behavior to ensure it aligns more closely with the practical standards and objectives of communication systems.

\subsubsection{Pre-training}
Pre-training LAMs in communications is a critical step in building foundational models equipped with both domain-specific knowledge and language understanding capabilities. It aims to address the limited adaptability of general-purpose LAMs to specialized communication tasks. Pre-training involves unsupervised learning on large-scale communication data, enabling the model to acquire fundamental concepts and structural knowledge relevant to the communication domain, thereby laying a solid foundation for downstream tasks. In communication scenarios, a continual pre-training strategy is typically adopted, where domain-specific data are introduced to further pre-train open-source LAMs (e.g., LLaMA, Gemma). The data sources include 3GPP standards, IEEE publications, communication patents, source code, and Wikipedia entries, forming high-quality corpora such as OpenTelecom \cite{zou2024telecomgpt} and CommData-PT \cite{jiang2025commgpt}. The training objective is to predict the next token, allowing the model to learn the semantics and logic in communication contexts. Upon completion, the model retains its general language capabilities while significantly enhancing its understanding of communication protocols, terminologies, and system structures, thereby establishing a strong foundation for subsequent instruction fine-tuning and alignment.

The specific pre-training approach involves continual pre-training on domain-specific communication datasets \cite{zou2024telecomgpt}. Unlike the initial pre-training phase of LAMs, this process offers a cost-effective method to adapt a general-purpose LAM into a communication-specialized model. During continual pre-training, the learning objective is based on causal language modeling, where the model predicts the next token given the preceding word sequence. Formally, let the input text be represented by a word sequence $x=(x_1,...,x_T)$ and $\theta$ denote the model parameters. The LAM is trained by minimizing the negative log-likelihood loss to enhance its understanding of communication knowledge. The loss function is defined as follows \cite{zou2024telecomgpt}:

\begin{equation}
 L{(x,\theta)}=-\sum_{t=1}^{T}\log\mathbb{P}{(x_t|(x_{<t}))}, 
\end{equation}
where $x_{<t}$ denotes the sequence of tokens preceding the token $x_t$, and $T$ is the length of the word sequence.

\subsubsection{Fine-tuning}
Fine-tuning is a critical post-pretraining stage for communication-oriented LAMs, aimed at enhancing their understanding and execution capabilities for specific communication tasks such as protocol parsing, question answering, and code generation. This process involves supervised training on high-quality, task-specific datasets to enable the model to generate professional and accurate outputs in response to given instructions. Fine-tuning typically leverages instruction datasets such as CommData-FT \cite{jiang2025commgpt} or Telecom-Instruct \cite{zou2024telecomgpt}, and optimizes the model using a cross-entropy loss function. To improve training efficiency, Parameter-Efficient Fine-Tuning (PEFT) techniques such as LoRA are often employed. As a result, the fine-tuned communication model exhibits stronger domain specificity and task adaptability, achieving superior performance in tasks like telecom question answering and code generation compared to general-purpose LAMs\cite{jiang2025commgpt}.

Instruction tuning performs SFT of communication-oriented LAMs using instruction-tuning datasets. These datasets consist of multiple instruction–response pairs, where each instruction $x^{(i)}$ is paired with a corresponding response $y^{(i)}$. The dataset can be formally represented as $I={\begin{Bmatrix}x^{(i)},y^{(i)}\end{Bmatrix}}_{i=1}^{N}$.
Using these instruction–response pairs, the LAM is trained to minimize the conditional negative log-likelihood loss of the response given the instruction, thereby enhancing its performance in zero-shot or few-shot scenarios and reducing refusal behaviors when responding to user requests. The loss function is defined as follows \cite{zou2024telecomgpt}:
   \begin{equation}
   L{(y^{(i)},\theta)}=-\sum_{t=1}^{\left|y^{(i)}\right|}\log\mathbb{P}{(y^{(i)}_t|y^{(i)}_{<t},x^{(i)})}    
   \end{equation}
where $x^{(i)}$ denotes the instruction in the $i$-th instruction–response pair, $y^{(i)}$ represents the corresponding correct response, $y^{(i)}_t$ is the $t$-th token in the response sequence of the $i$-th sample, and $y^{(i)}_{<t}$ denotes the subsequence of response tokens from the beginning up to the $(t-1)$-th token.

\subsubsection{Alignment}
In communications, alignment is a critical step in enhancing the practical utility of LAMs, aiming to ensure that the generated outputs better reflect the preferences and requirements of communication tasks. Although fine-tuned models possess a certain level of capability in handling such tasks, they may still produce redundant, inaccurate, or task-irrelevant responses, necessitating further optimization. Alignment addresses this by constructing preference-labeled datasets and applying methods such as Direct Preference Optimization (DPO), which encourages the model to generate concise, accurate, and highly relevant outputs. Compared to traditional RL approaches, DPO eliminates the need for complex reward models, resulting in more efficient and stable training. Ultimately, aligned models demonstrate performance that is more consistent with real-world expectations in tasks such as telecom-specific question answering and protocol interpretation, making them better suited for future intelligent communication systems.
The specific loss function of DPO is defined as follows \cite{zou2024telecomgpt}:
\begin{equation}
   \mathcal{L}_{\text{DPO}}(\pi_\theta; \pi_{\text{ref}})= - \mathbb{E}_{(x, y_w, y_l)}  
\end{equation}
\begin{equation}
\Big[ 
\log \sigma \Big( 
\beta \log \frac{\pi_\theta(y_w|x)}{\pi_{\text{ref}}(y_w|x)}- \beta \log \frac{\pi_\theta(y_l|x)}{\pi_{\text{ref}}(y_l|x)} \Big) 
\Big]
\end{equation}
where $x$ denotes the input prompt or task instruction, $y_w$ represents the preferred response according to human feedback, and $y_l$ denotes the less preferred response. $\pi_\theta$ is the current model being optimized, while $\pi_{\text{ref}}$ is the reference model, typically a fine-tuned model without preference alignment. $\pi_\theta(y_w|x)$ denotes the conditional probability of generating response $y_w$ given input $x$ under the model $\pi_\theta$.

\subsection{External Learning}
In communications, the external learning paradigm for LAMs primarily follows two approaches: RAG based on vectorized data and KG based on structured graph data. RAG enhances the model's information retrieval and generation capabilities by incorporating external semantic vector databases, while KG strengthens the model's knowledge reasoning and contextual understanding through structured representations. Together, these approaches significantly expand the applicability and effectiveness of LAMs in complex communication scenarios \cite{jiang2025commgpt}.

\subsubsection{Retrieval-Augmented Generation}
In the external learning paradigm of LAMs for communications, vector-based RAG serves as a key knowledge enhancement approach, addressing limitations such as slow knowledge updates and inaccurate responses in pre-trained models. RAG integrates information retrieval with generative modeling by retrieving relevant content from an external knowledge base before response generation. The retrieved content is combined with the original query and jointly fed into the LAM to produce more accurate and context-aware outputs. The typical workflow involves segmenting communication documents and converting them into high-dimensional vectors, storing them in a vector database with indexing. Upon receiving a user query, the model encodes the query into a vector and retrieves the most relevant document fragments. Finally, these retrieved segments are combined with the original query and passed to the LAM to produce an enhanced response.

In communication scenarios, vector-based RAG focuses on extracting semantic information from domain-specific communication documents to construct specialized communication vector databases. This approach enables the rapid integration of new knowledge without requiring updates to model parameters, making it particularly suitable for communication tasks characterized by frequent knowledge updates and high complexity. For example, the CommGPT system \cite{jiang2025commgpt} employs vector databases such as milvus \cite{wang2021milvus}, significantly improving the model's accuracy in understanding technical terminology and complex standards, and demonstrating strong scalability and practical utility.

\subsubsection{Knowledge Graph}
In the external learning paradigm of LAMs for communications, graph-based KGs serve as a vital approach for knowledge modeling and enhancement, enabling the model to better understand complex communication concepts and their interrelationships. A KG is a structured knowledge base that represents entities (e.g., protocols, technologies, parameters) as nodes and their semantic relationships as edges within a graph structure. KGs provide global, structured knowledge support for LAMs, thereby enhancing their reasoning, retrieval, and interpretability capabilities. The construction of a KG typically involves three main steps: first, extracting entities along with their attributes and relationships from communication documents; second, constructing a graph structure to capture the semantic associations among entities; and finally, storing the graph in a graph database (e.g., neo4j \cite{henriquezgraph}), which allows the model to query and reason over the graph to obtain relevant entities and contextual information.

In communication scenarios, graph-based KGs are primarily used to represent and organize key entities and their associations within communication documents. By constructing a KG, the model can identify not only explicit relationships (e.g., a protocol belonging to a specific standard) but also infer implicit ones (e.g., the relevance of a particular technology to multiple protocols). For instance, in the CommGPT system \cite{jiang2025commgpt}, KGs are integrated with RAG to form a multi-scale knowledge enhancement mechanism, enabling the model to generate more accurate and contextually consistent responses to communication tasks involving multiple entities and multi-hop relationships. This significantly improves the model's logical reasoning and global understanding capabilities in complex communication scenarios.

\begin{figure*}[htpb]
		\centering
		\includegraphics[width=0.8\textwidth,height=0.45\textwidth]{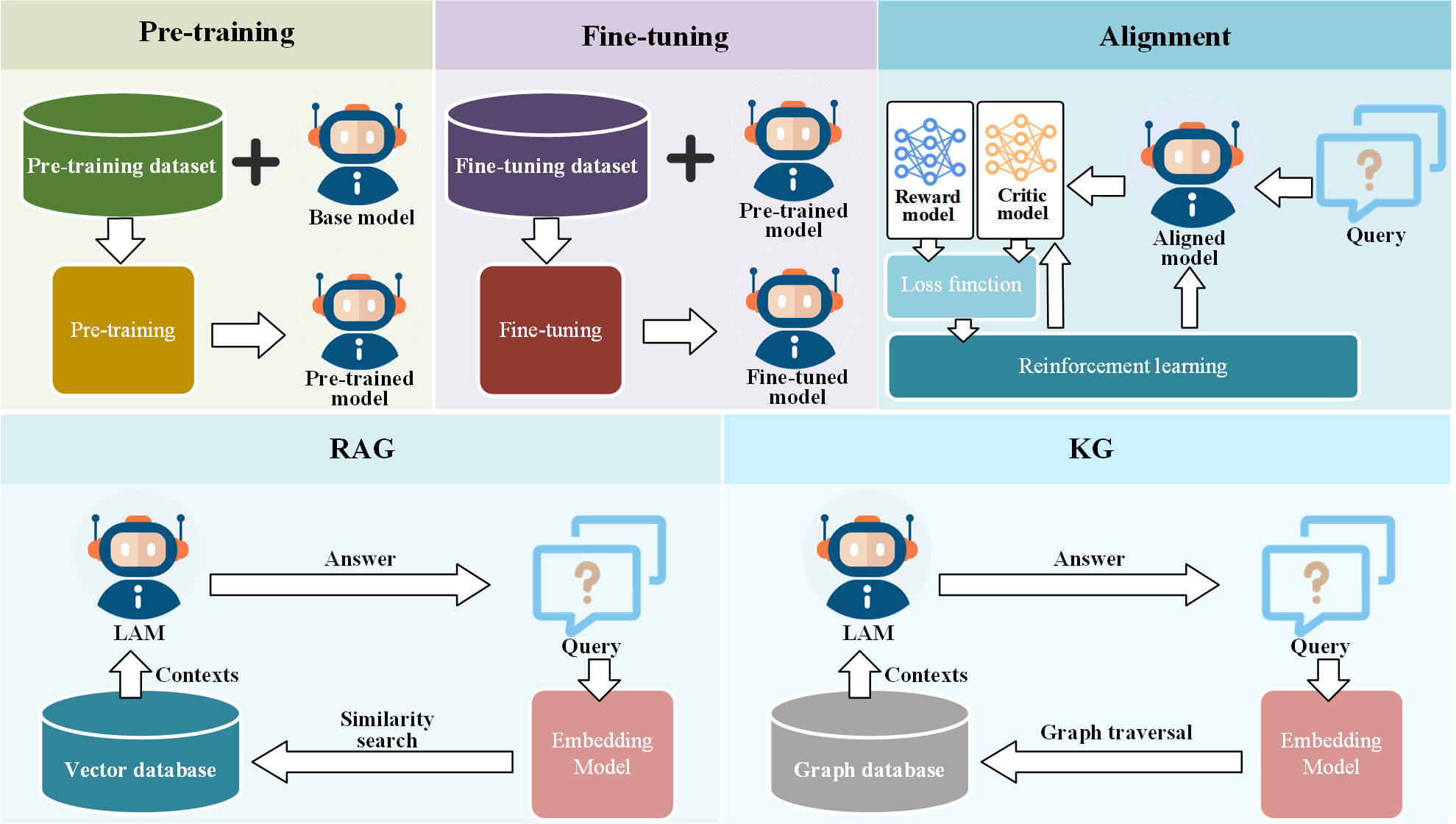}
		\caption{The structured design pipeline of LAMs for communications through various learning methods.}
		\label{fig:fig3}
	\end{figure*}

Table \ref{tab:comparison} presents the comparison of internal learning and external learning. 
As shown in Fig. \ref{fig:fig3}, a structured design pipeline is established to develop LAMs specifically optimized for communications through various learning methods.

\begin{table*}[htbp]
\centering
\caption{Comparison of internal learning and external learning.}
\label{tab:comparison}
\scriptsize
\renewcommand{\arraystretch}{1.2}
\begin{tabular}{|p{1.8cm}|p{2.4cm}|p{2.8cm}|p{2.8cm}|p{2.8cm}|p{3cm}|}
\hline
\textbf{Aspect} & \textbf{Pretraining} & \textbf{Fine-tuning} & \textbf{Alignment} & \textbf{RAG} & \textbf{Knowledge Graph}  \\
\hline
Goal & Learning general language patterns & Optimizing performance for specific tasks & Align outputs with  preference or task-specific constraints & Retrieving external documents to enhance LLM & Constructing structured knowledge to enhance LLM  \\
\hline
Data Type & Large-scale unlabeled text & Labeled instruction data & Preference data, synthetic alignment data & External documents/vector database & Structured graph database  \\
\hline
Learning Method & Unsupervised learning & Supervised learning & Reinforcement learning & Retrieval and in-context learn-ing & Graph reasoning  \\
\hline
Tuning Parameters & All parameters & Task-specific parameters & Alignment-specific parameters  & Parameters of retrieval modules & Graph embedding and structure \\
\hline
Time Complexity & High & Medium  & Medium & Low & Medium \\
\hline
Disadvantages & Lack of task-specific optimization & Labeled instruction data requirement & Costly reward modeling and subjective criteria  & Retrieval suboptimality & Difficulty in generating high-quality graph structure. \\
\hline
\end{tabular}
\end{table*}

\subsection{Summary and Lessons Learned}
\subsubsection{Summary}
This chapter systematically summarizes the key components and technical pathways for building LAMs for communications. Starting from the construction of communication datasets, we outline the filtering strategies for communication-specific data, methods for building pre-training, instruction fine-tuning, and preference alignment datasets, forming a comprehensive design scheme covering the entire data lifecycle. Regarding model training, we introduce internal learning and external learning separately, clarifying the objective functions and technical approaches at each stage. Internal learning includes key techniques such as pre-training, instruction fine-tuning, and preference alignment, while external learning involves vector-based RAG and graph-based KG. Through this content, we provide a complete technical reference and methodological summary for the systematic design of LAMs for communications.

\subsubsection{Lessons Learned}
\textcolor{black}{Although preliminary achievements have been made in dataset construction, pre-training, fine-tuning, and alignment of LAMs, internal learning still faces challenges such as knowledge update delays and high training costs, while external learning encounters difficulties including complex knowledge organization and system integration. Future efforts should focus on constructing high-quality, multi-level communication datasets, improving model adaptability and update capability for communication knowledge, and optimizing external enhancement mechanisms like RAG and KG, thereby promoting efficient and scalable structure design and development in communication scenarios \cite{jiang2025comprehensive}.}

\section{How to Design Agentic AI Systems for Communications}
The LAM-based Agentic AI system \cite{jiang2024large3} refers to an intelligent agent framework driven by LAMs and integrated with key modules such as a knowledge base, planners, tools, and memory modules. It is designed to autonomously comprehend, plan, and execute complex communication tasks. This system not only possesses natural language understanding and reasoning capabilities but also supports multi-agent collaboration to enable iterative optimization of communication tasks. Compared with traditional agent systems, the Agentic AI system, empowered by LAMs, exhibits greater generality, scalability, and adaptability, thereby unlocking the full potential of AI in 6G networks. In the following sections, we provide a detailed overview of the system’s core components, agent interaction mechanisms, and multi-agent system architecture.

\begin{figure*}[htpb]
		\centering
		\includegraphics[width=0.8\textwidth,height=0.47\textwidth]{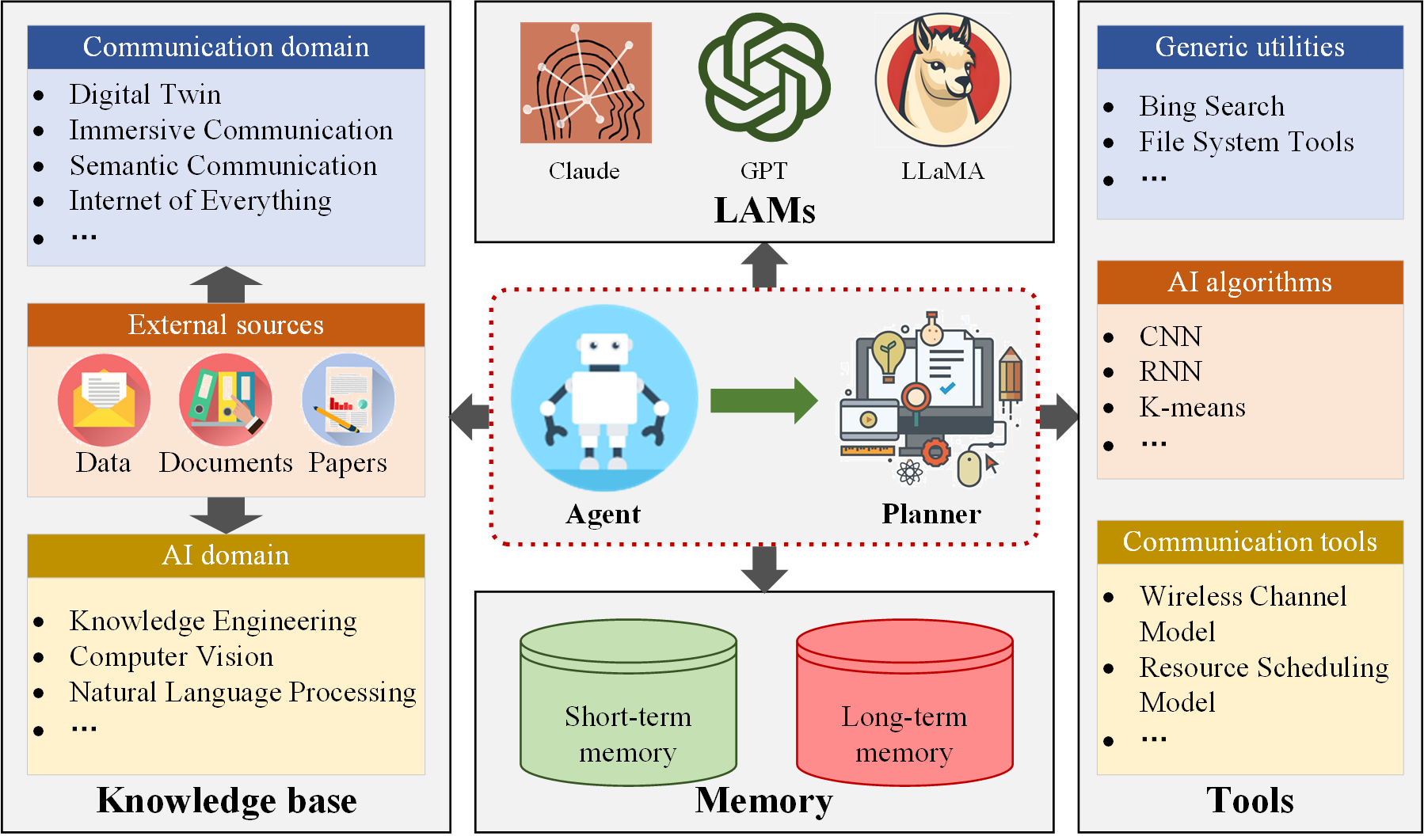}
		\caption{The architecture of the LAM-based Agentic AI system.}
		\label{fig:fig4}
	\end{figure*}

\subsection{System Architecture of Agentic AI}
The Agentic AI system is composed of LAMs, knowledge bases, planners, tools, and memory modules, collectively enabling natural language understanding, knowledge reasoning, and task execution. The agent leverages the knowledge base to acquire domain-specific knowledge in communications, utilizes the planner to reason about and decompose tasks, invokes external tools to perform operational steps, and employs the memory module to store and retrieve historical information for reflection and continuous task optimization. The system architecture of LAM-based Agentic AI is shown in Fig. \ref{fig:fig4}.

\subsubsection{LAMs}
In an Agentic AI system, LAMs serve as the core reasoning engine and central coordination hub, responsible for task comprehension, contextual modeling, tool invocation, and reflective evaluation. LMMs such as GPT-4 or Gemini are typically employed to enable multimodal perception and interpretation of external commands and environments, thereby orchestrating the system's overall operations. LAMs utilize prompt templates to structure task processing workflows, invoke the planner to dynamically generate subtask instructions, and dispatch them to other modules while actively monitoring execution states and coordinating outputs to ensure semantic consistency and logical coherence. Furthermore, LAMs can interface with external tools such as search engines, code executors, and file systems, as well as internal tools tailored to communication tasks, enabling a closed-loop cycle from reasoning to execution. As a unified decision-making engine connecting users, agents, tools, and the knowledge base, LAMs significantly enhance the system’s adaptability and scalability in handling complex communication tasks.

\subsubsection{Planner}
The planner module is a central component responsible for understanding tasks, decomposing objectives, and organizing execution in an Agentic AI system. It typically employs LRMs with strong language comprehension and inference capabilities, such as OpenAI-o1\cite{jaech2024openai} and DeepSeek R1\cite{guo2025deepseek}, which leverage slow thinking and CoT reasoning to break down complex tasks into executable subtask sequences. The planner generates feasible task chains, explicitly defining the order and dependencies among subtasks to guide downstream execution. The system adopts advanced reasoning strategies such as CoT\cite{wei2022chain}, Tree-of-Thought (ToT)\cite{yao2024tree}, Graph-of-Thought (GoT)\cite{besta2024graph}, and the Plan-and-Solve framework \cite{wang2023plan}, progressively refining tasks into actionable steps while iteratively improving plans based on execution feedback. This module often integrates key techniques including multi-agent collaboration, structured task modeling, and feedback-driven replanning, significantly enhancing the system’s efficiency and adaptability in managing complex communication tasks.

\subsubsection{Knowledge Base}
In an Agentic AI system, the knowledge base module serves as a foundational pillar supporting task comprehension and reasoning, providing agents with structured, retrievable, and highly relevant external knowledge. This module integrates communication-related documents (e.g., research papers, standards, protocols) and AI domain knowledge. It is constructed through semantic encoding and knowledge embedding, enabling efficient alignment with the knowledge retrieval requirements of LAMs. The knowledge base may consist of both vectorized and graph-structured data: vector data are retrieved via RAG querying \cite{gao2023retrieval}, while graph data are accessed through KG querying \cite{pan2024unifying}. When external knowledge is required for task execution, user queries are converted into retrieval-style prompts to search the knowledge base. Relevant segments are returned based on semantic similarity ranking, then filtered and reformulated by the agent to provide task-contextualized knowledge support. This module supports both structured knowledge acquisition and dynamic evolution with adaptive updates.

\subsubsection{Tools}
The tool module serves as a critical bridge between language understanding and task execution, enabling agents to perform “perception–reasoning–action” operations in an Agentic AI system. This module integrates various types of external tools, including general-purpose tools (e.g., search engines, file access, API invocation, data analyzers), communication-specific tools (e.g., channel models, beamforming algorithms, resource allocation algorithms), and AI-specific tools (e.g., clustering algorithms, feature extraction methods, deep learning algorithms), forming a comprehensive execution toolkit for the system. Agents comprehend the functionality of these tools through natural language descriptions and issue tool invocation commands, transmitting data to the appropriate tool for processing. Upon execution, the output is returned to guide subsequent steps in the task pipeline. Throughout this process, the tool module can be repeatedly invoked by multiple agents to support tasks such as code generation, data processing, and system modeling, acting as a core enabler in bridging reasoning and action. This module incorporates key technologies such as prompt engineering \cite{sahoo2024systematic}, Agent Communication Protocol (ACP)\cite{ehtesham2025survey}, and tool selection mechanisms \cite{qu2025tool}, thereby enhancing the intelligence and flexibility of tool utilization and improving system adaptability in multi-turn interactions and complex task scenarios.

\subsubsection{Memory}
In an Agentic AI system, the memory module is a core component that supports self-reflection, task optimization, and continual learning. It is responsible for recording and managing intermediate states and outcomes throughout the task execution process, thereby establishing both short-term and long-term memory mechanisms. Short-term memory captures semantically similar task experiences to facilitate comparative analysis and localized optimization, while long-term memory retains semantically distinct experiences to support system-level strategy updates and global optimization. After each task, the agent evaluates its execution plan and outcomes, classifies and stores them accordingly, and then leverages short-term memory to propose fine-grained workflow adjustments and long-term memory to perform broader workflow revisions. This module emulates the human cognitive process of “short-term recall and long-term accumulation,” enabling memory-driven self-optimization across multi-turn interactions and serving as a critical foundation for autonomous learning and task generalization. The memory module integrates key technologies such as short and long-term memory modeling \cite{lindemann2021survey}, vector databases \cite{han2023comprehensive}, semantic embeddings \cite{ueki2021survey}, and self-reflection mechanisms\cite{shinn2023reflexion}, significantly enhancing the system’s capacity for information organization and knowledge evolution in complex communication tasks.

\subsection{Agent Interaction}
As the core units of decision-making and execution, agents rely on the comprehension and generation capabilities of LAMs to engage in autonomous or collaborative interactions centered around task objectives and the external environment. Depending on the scope and nature of the interaction, agent interaction mechanisms can be categorized into two types: single-agent interaction and multi-agent collaborative interaction.

\subsubsection{Single-Agent Interaction}
Single-agent interaction encompasses task reasoning optimization and causal logic refinement, which effectively enhance the reasoning efficiency of LAMs and strengthen their understanding of causal relationships.

\begin{itemize}
    \item \textbf{Task Reasoning Optimization:}
Agents possess autonomous decision-making and self-learning capabilities, enabling them to monitor the state of the LAM in real time during the reasoning process, identify bottlenecks, and dynamically adjust reasoning paths. By leveraging contextual information from the knowledge base and memory modules, agents can adapt to task variations, thereby improving reasoning efficiency and accuracy. For instance, the AutoGPT system achieves automated reasoning by decomposing goals into subtasks and executing them iteratively \cite{yang2023auto}. Integrating agents with RL can further enhance their policy learning capabilities, leading to superior reasoning performance across a variety of tasks.

\item \textbf{Causal Logic Optimization:}
By incorporating causal logic, agents can significantly enhance their reasoning capabilities. Through the integration of KG, agents are able to analyze causal relationships among various factors, enabling them not only to identify complex patterns within the data but also to understand the underlying causal mechanisms. This allows for more targeted and context-aware reasoning. Agents can determine which variables exert direct or indirect influence on outcomes and accordingly adjust their reasoning strategies, thereby improving prediction accuracy and decision quality while reducing errors arising from the neglect of causal dependencies.
\end{itemize}

\subsubsection{Multi-Agent Interaction}
In multi-agent systems, each agent functions as an independent decision-making unit that collaborates with others to solve complex problems. Common optimization strategies include unordered complementary collaboration, ordered complementary collaboration, and adversarial collaboration.

\begin{itemize}
    \item \textbf{Unordered Complementary Collaboration:}
Unordered complementary collaboration emphasizes flexible interaction among agents without predefined sequences or rules. Each agent contributes information and strategies based on its own experience, enhancing system diversity and enabling the analysis of complex and uncertain problems from multiple perspectives. This approach mitigates the limitations of single-agent viewpoints, improves overall reasoning capabilities, and increases the system’s adaptability in dynamic environments.

\item \textbf{Ordered Complementary Collaboration:}
Ordered complementary collaboration follows a well-defined sequence and structured process among multiple agents. Information is passed along a predetermined path, and tasks are executed in a coordinated manner, forming a continuous chain of knowledge. The output of one agent serves as the input for the next, thereby improving collaborative efficiency, reducing information redundancy and interference, optimizing the reasoning process, and enhancing task execution efficiency and decision accuracy in complex scenarios.

\item \textbf{Adversarial Collaboration:}
Adversarial collaboration involves competitive interactions among agents, where they continuously challenge and critique each other’s reasoning strategies. This dynamic fosters self-reflection and strategic refinement, as agents learn from the strengths and weaknesses of their counterparts. Such game-theoretic learning not only enhances individual reasoning abilities but also strengthens the system’s overall robustness and problem-solving depth in complex environments.
\end{itemize}

\subsection{Multi-Agent System Architecture}
\begin{figure*}[htpb]
		\centering
		\includegraphics[width=1\textwidth,height=0.38\textwidth]{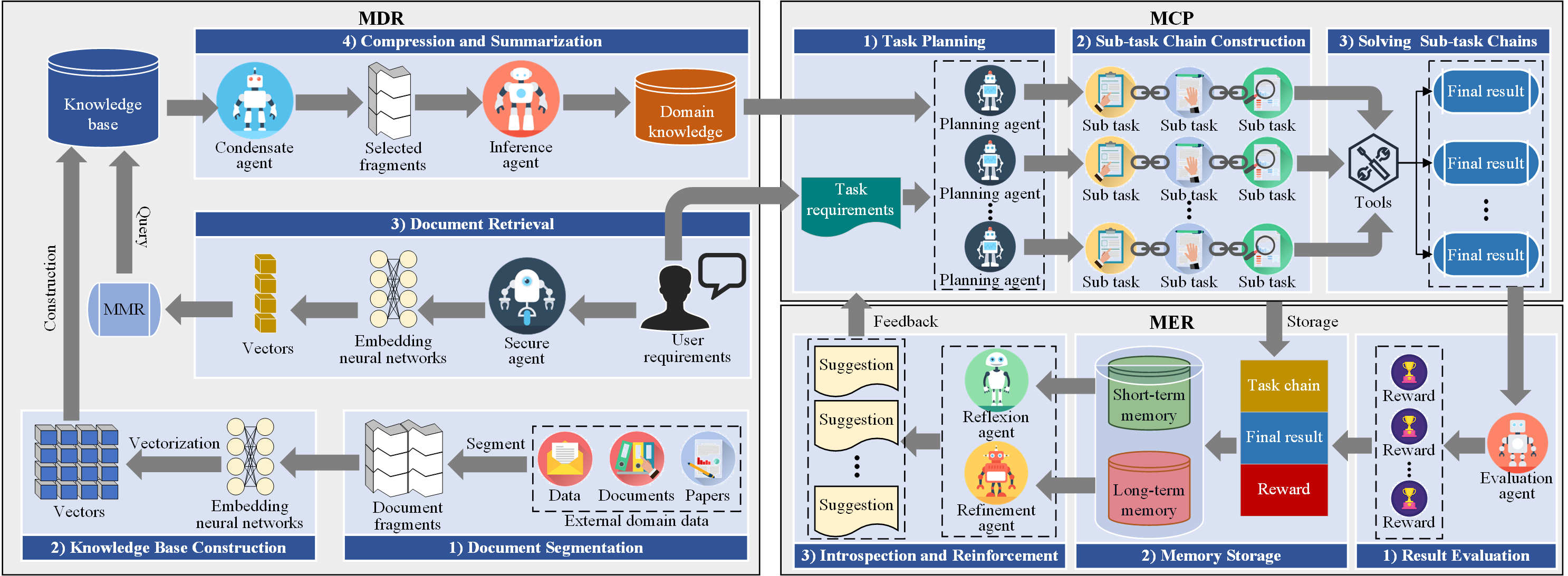}
		\caption{Schematic diagram of CommLLM \cite{jiang2024large3}.}
		\label{fig:fig5}
	\end{figure*}

The CommLLM framework \cite{jiang2024large3} establishes a LAM-centric, multi-agent collaborative system architecture for 6G communications. The schematic diagram of CommLLM is shown in Fig. \ref{fig:fig5}. This architecture integrates a knowledge base, planners, tools, and memory modules to support intelligent decision-making and task execution. Driven by natural language input, the system orchestrates the full process of knowledge retrieval, task planning, result evaluation, and self-optimization through distributed collaboration among agents. Each agent is assigned to specific subtasks, working in coordination to form a closed-loop workflow of “input–reasoning–feedback–optimization.” This enables the system to handle complex communication tasks, adapt to dynamic environments, and continuously learn, representing a critical technological pathway for the intelligent evolution of 6G communication systems. The architecture consists of the following three components:

\begin{itemize}
    \item \textbf{Multi-Agent Data Retrieval:}
The Multi-agent Data Retrieval (MDR) module is responsible for acquiring task-relevant information from external knowledge bases and serves as the foundational entry point for semantic modeling and knowledge support in the system. It comprises multiple function-specific agents, including a secure agent, a condensate agent, and an inference agent, forming a multi-stage data retrieval pipeline from input filtering to information reconstruction. Specifically, when a user submits a natural language task request, the secure agent first screens the input to eliminate potentially malicious instructions or requests that violate system constraints, ensuring the robustness and compliance of the task chain. The system then leverages embedded vector retrieval mechanisms to identify the most relevant knowledge segments from the pre-constructed knowledge base based on the processed query semantics. These retrieved segments are then compressed and cleaned by the condensate agent to remove redundancy and retain core content. Finally, the inference agent utilizes the LAM’s language understanding and generation capabilities to perform semantic abstraction, logical integration, and knowledge reconstruction on the compressed text, forming structured, task-ready knowledge representations. This module not only ensures the contextual accuracy and relevance of knowledge used in downstream reasoning but also, through collaboration with the Multi-agent Collaborative Planning (MCP) module, significantly enhances the system's domain adaptability, response efficiency, and interpretability in complex communication tasks. It constitutes the foundational gateway for establishing semantic closure and knowledge-enhanced reasoning in Agentic AI systems.

\item \textbf{Multi-Agent Collaborative Planning:}
The MCP module plays a central role in decomposing complex communication tasks and generating high-quality execution pathways. It operates by orchestrating multiple planning agents configured with diverse roles and reasoning strategies, enabling parallel, multi-perspective task analysis. After receiving user requests and domain-relevant information from the knowledge base, each planning agent applies reasoning frameworks such as CoT or the Plan-and-Solve strategy to semantically model the task and divide it into subtasks. These subtasks are organized into chains with defined execution order and dependency relationships. The resulting task chains are structured to ensure logical consistency and semantic coherence, and may be executed sequentially or in parallel. The system then calls upon the LAM’s intrinsic tool capabilities or external integrated tool modules to address each subtask individually and generate preliminary execution results. The multi-agent planning strategy of MCP significantly enhances the system’s responsiveness, reasoning robustness, and generation quality in multi-objective, constraint-rich scenarios. It also lays the groundwork for the evaluation and refinement process carried out by the Multi-agent Evaluation and Reflection (MER) module. MCP effectively addresses the “path limitation” and “single-mode bias” inherent in traditional single-agent systems, serving as the core planning engine for complex communication task execution in Agentic AI.

\item \textbf{Multi-Agent Evaluation and Reflection: }
The MER module is the core mechanism for evaluating solution quality, generating self-feedback, and enabling iterative optimization. It is designed to compensate for potential deviations, inefficiencies, or illogical reasoning that may occur during a single inference pass. This module comprises multiple functional agents responsible for evaluation, reflection, and optimization. In the operational workflow, MER receives multiple candidate task plan chains generated by the MCP module, each representing a distinct reasoning path and execution strategy proposed by different planning agents. Evaluation agents assess and rank these candidates across multiple dimensions, including accuracy, efficiency, and interpretability, based on predefined task objectives, contextual constraints, and prior experience. The reflexion agents then access short-term memory to trace contextual trajectories and intermediate variables, identify weaknesses or logical gaps in the reasoning process, and use the LAM’s reasoning abilities to propose fine-grained improvement suggestions. Refinement agents further leverage high-quality historical solutions and paradigms stored in long-term memory to perform structural strategy reconstruction and path rewriting, feeding the optimized plans back to the MCP module for the next iteration. Importantly, MER is not merely a result scoring mechanism; it is tightly integrated with the memory system and agent behavior configurations, forming a reflective and learnable feedback loop in the Agentic workflow. This greatly enhances the system’s stability, adaptability, and robustness in addressing complex and open-ended tasks.
\end{itemize}

\subsection{Summary and Lessons Learned}
\subsubsection{Summary}
This chapter summarizes the core concepts and system architecture of LAM-based Agentic AI Systems, systematically outlining their constituent modules, interaction mechanisms, and functional workflows. We focus on the design of the reasoning engine centered on LAMs, constructing an agent system framework for complex communication tasks around key modules such as knowledge base retrieval, task planning, tool invocation, and memory optimization. Through analysis of agent interaction methods and optimization mechanisms, we demonstrate the system’s capabilities in task modeling, reasoning execution, and self-evolution, providing a technical foundation and methodological reference for advancing research and implementation of Agentic AI Systems tailored for 6G communications.

\subsubsection{Lessons Learned}
\textcolor{black}{Although LAM-based Agentic AI systems have shown good generality and scalability across various communication tasks, challenges remain, such as complex multi-agent collaboration, instability in task planning, low efficiency in knowledge base retrieval, and limited flexibility in tool integration and invocation. Future research should aim to enhance collaboration strategies and consistency modeling among multiple agents, build more efficient knowledge management mechanisms, strengthen the intelligence and generalization capability of tool invocation, and promote the continuous development of autonomous adaptation and multi-turn optimization capabilities of the Agentic AI system in complex communication scenarios \cite{sapkota2025ai}.}

\section{How to Optimize Communication Systems Using LAMs and Agentic AI}

\subsection{The Application Scenarios of LAMs }

\subsubsection{LAMs for Semantic Communication}
With the remarkable capabilities of LAMs in natural language understanding and cross-modal reasoning tasks, semantic communication is gradually evolving from the traditional bit transmission paradigm toward an intelligence-driven, semantic-centric communication paradigm.

In semantic modeling and optimization, introducing LLMs enables direct modeling and efficient decoding of semantic information. The general knowledge acquired through pretraining supports end-to-end semantic communication even without fine-tuning \cite{wang2024large}.
Moreover, LLMs enhance feature extraction and reconstruction in semantic communication by constructing general-purpose knowledge bases, enabling intelligent optimization across task identification, semantic processing, and physical transmission. By integrating the SC-GPT approach, LLMs further improve channel adaptability and transmission efficiency, thereby significantly enhancing the accuracy and robustness of semantic communication \cite{jiang2024semantic}.
In end-to-end semantic communication systems, the integration of KG with LLMs in the KG-LLM framework significantly enhances transmission efficiency and semantic reconstruction quality by enabling structured semantic extraction and context-aware compression encoding, thereby greatly improving the overall performance of semantic communication \cite{salehi2025llm}.
In talking-head video semantic communication, LLMs construct private knowledge bases to enable semantic error correction and disambiguation and participate in joint semantic–channel encoding, thereby improving the accuracy and robustness of text transmission. Simultaneously, they facilitate high-quality reconstruction of speech and video, enabling more efficient and accurate semantic transmission and recovery under low-bandwidth conditions \cite{jiang2024large6}.

In image semantic communication systems, the incorporation of LLM-designed semantic encoders and context-aware semantic decoders enables the extraction of high-density semantic information from raw images and the reconstruction of contextually consistent representations at the receiver. This significantly enhances the efficiency of semantic communication \cite{ribouh2025large}.
In multi-user image semantic communication systems, the SAM leverages a lightweight knowledge base to identify key semantic regions within images and employs efficient semantic encoding for image compression and reconstruction. Through a shared semantics mechanism, similar information is transmitted uniformly across users, reducing redundancy and enhancing both the efficiency and robustness of semantic communication \cite{jiang2025lightweight}.
In cross-modal semantic communication systems, a Vision-Language Model-based Cross-modal Semantic Communication (VLM-CSC) framework is constructed based on BLIP and Stable Diffusion models. This system performs high semantic density extraction from images to text and enables controllable image reconstruction. Under dynamic channel conditions, it integrates memory-enhanced continual learning and noise-aware attention modulation modules, significantly improving the completeness of semantic representation and the robustness of semantic communication transmission \cite{jiang2024visual}.
In 3D semantic communication systems, the Generative AI Model-assisted 3D Semantic Communication (GAM-3DSC) framework, integrating SAM, NeRF, and diffusion models, performs task-oriented 3D semantic extraction, adaptive semantic compression, and channel estimation aided by a conditional Generative Adversarial Network (GAN) combined with diffusion models. This design enables efficient compression, transmission, and reconstruction of 3D semantic information under uncertain channel conditions, significantly enhancing multi-view semantic representation and the robustness of semantic communication \cite{jiang2024large1}.

In multimodal semantic communication systems, the proposed LAM-based Multimodal Semantic Communication (LAM-MSC) framework integrates the CoDi model for modality transformation and leverages a personalized knowledge base built on GPT-4 for semantic extraction and reconstruction. Additionally, Conditional GANs (CGANs) are employed for channel estimation, significantly improving the transmission efficiency of the semantic communication system under complex channel conditions and multimodal data scenarios \cite{jiang2024large5}.
Moreover, in 6G semantic communication systems, the Privacy-preserving Semantic Communication scheme based on Multimodal LLM (MLLM-PSC), built upon the GPT-4V architecture, incorporates semantic extraction with user-profile-driven few-shot prompt learning and sensitive semantic encryption mechanisms. This framework enhances multimodal information compression and semantic consistency while achieving end-to-end privacy protection and low-overhead transmission of sensitive content \cite{cao2024multimodal}.
The M4SC system integrates the reasoning and generalization capabilities of LLMs to construct a unified semantic space via Kernelized Alignment Networks (KAN) for multimodal alignment. It optimizes multi-task representation and transfers through natural language instruction templates and separates shared and private semantic transmissions in multi-user scenarios, significantly improving the compression ratio, bandwidth utilization, and adaptability of semantic communication systems \cite{jiang2025m4sc}.
In bandwidth-constrained environments, LLMs enhance the efficiency and robustness of underwater image semantic communication by identifying key information within images and performing semantic compression. By incorporating diffusion models and ControlNet, high-quality image reconstruction is achieved, while language-based LLMs are employed to recover textual information. This approach significantly reduces data volume while ensuring semantic consistency and reconstruction fidelity \cite{chen2024semantic}. Meanwhile, LLMs are employed in semantic communication systems within edge IoT networks, where they enable user intent recognition, semantic extraction, and reconstruction at the edge. This enhances information transmission efficiency and intelligent interaction capabilities in resource-constrained environments \cite{kalita2024large}.

In summary, LAMs empowered semantic communication systems are fundamentally reshaping the foundational units of information transmission, shifting from the transfer of symbols and bits to the accurate conveyance of meaning. This paradigm shift provides strong support for the development of the next generation of ubiquitous, intelligent, and efficient communication systems with enhanced reliability.

\begin{figure*}[htpb]
		\centering
		\includegraphics[width=1\textwidth,height=0.5\textwidth]{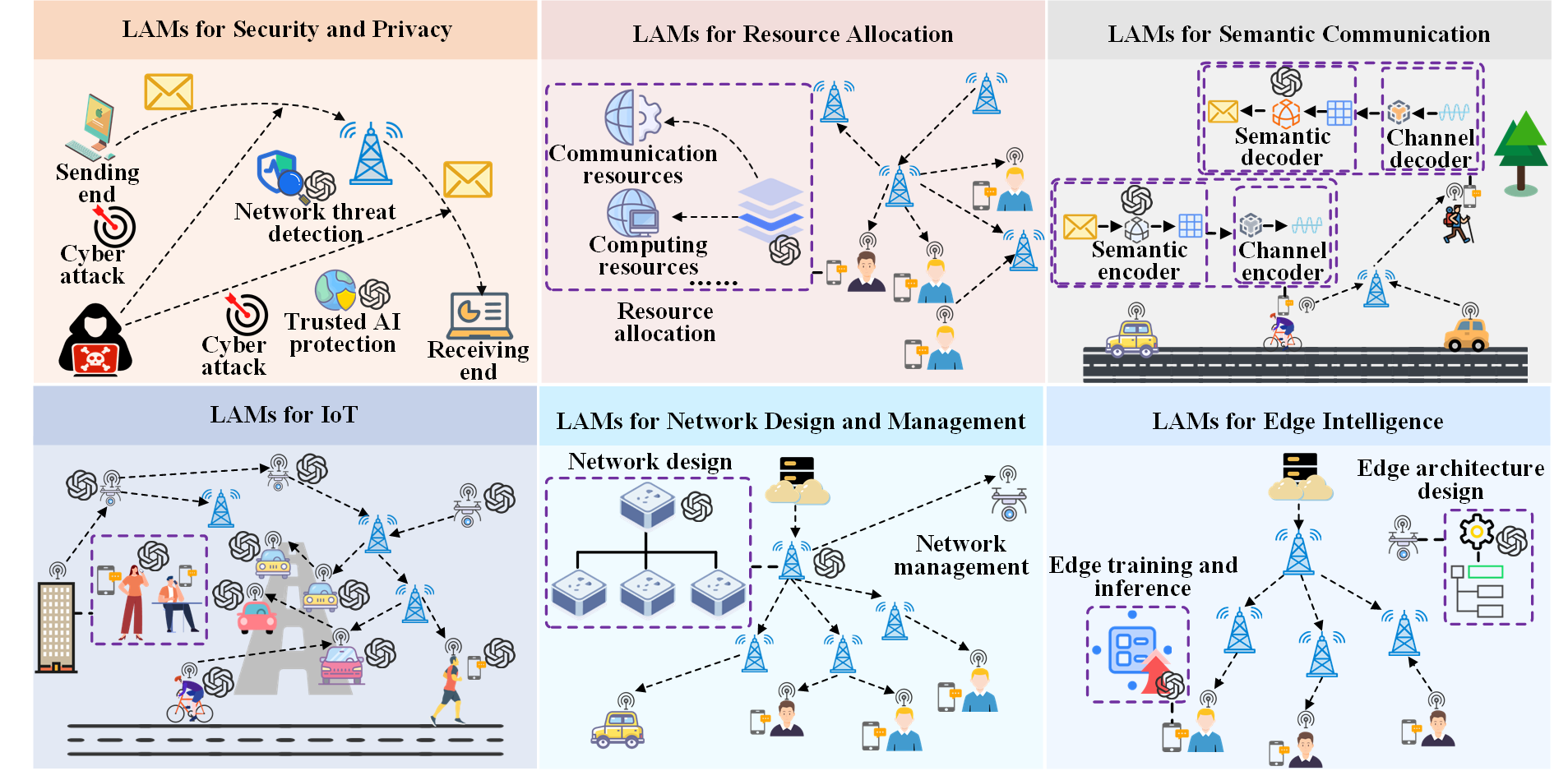}
		\caption{The application scenarios of LAMs.}
		\label{fig:fig6}
	\end{figure*}

\subsubsection{LAMs for IoT}
With the widespread deployment of the IoT and the continuous expansion of its application scenarios, traditional IoT systems face significant bottlenecks in task complexity, semantic understanding, device collaboration, and real-time interaction. The introduction of LAMs offers new perspectives for developing more intelligent and adaptive IoT systems.

In task orchestration and control automation, the LLMind framework introduces an LLM-centric orchestration mechanism to enable collaborative control across devices and functional modules. By employing a multi-stage transformation process—from natural language, to finite state machines, and ultimately to executable code—it supports dynamic orchestration and feedback-driven execution across multiple devices, thereby equipping IoT systems with the capabilities for continual learning and adaptive evolution \cite{cui2024llmind}.
Moreover, the AutoIoT system integrates LLMs into the development of Artificial intelligence \& IoT (AIoT) applications by translating user intentions into locally executable code through natural language programming. This approach offers strong interpretability and flexible interaction while avoiding privacy and latency issues commonly associated with remote model invocation. It presents a new paradigm for low-barrier, high-efficiency AIoT application development \cite{shen2025autoiot}.
LLMs can also enable on-device intelligence enhancement through edge inference and PEFT mechanisms. Furthermore, by leveraging collaborative reasoning and distributed learning, a resource-adaptive optimization framework can be established, significantly improving the responsiveness, privacy preservation, and cross-device generalization capabilities of IoT systems in applications such as smart healthcare, intelligent cities, and home automation \cite{chen2025llm}.

In privacy and security protection, LLMs combine local deployment with prompt engineering to enable understanding, task analysis, and code generation from complex data uploaded by IoT devices. Without the need for fine-tuning, they can perform reasoning and self-correction, thereby significantly enhancing the system’s data processing capabilities and intelligent responsiveness under privacy-sensitive, resource-constrained, and task-diverse scenarios. This demonstrates the critical role of LLMs in optimizing intelligent services within IoT systems \cite{xiao2024efficient}.
LLMs are also employed in the development of real-time threat detection and prevention systems for IoT networks. By leveraging fine-tuned lightweight models in combination with IoT-specific datasets, these systems enable rapid responses to unknown attacks and efficient defense under constrained resource conditions, highlighting the promising potential of LLMs in intelligent security systems \cite{otoum2025llm}.
Moreover, LLMs demonstrate significant potential in enhancing the security, manageability, and intelligent data processing capabilities of IoT systems by enabling high-precision few-shot recognition in Distributed Denial of Service (DDoS) attack detection, automatically generating multi-scenario control scripts within macro-programming frameworks, and providing high-quality, interpretable responses for large-scale sensor data processing tasks \cite{zong2025integrating}.

In semantic reasoning and multimodal modeling, LLMs enable the abstraction and fusion of multi-source sensor data to automatically identify and generate high-level activity events from raw logs, without relying on extensive manual rules or domain-specific knowledge. This significantly enhances the efficiency of IoT data processing and intelligent analysis, making it particularly suitable for applications such as smart healthcare and long-term monitoring. It highlights the critical role of LLMs in optimizing event understanding and unified log generation \cite{shirali2024llm}.
To enhance the reasoning capabilities of LLMs in the real physical world, the IoT-LLM framework integrates IoT sensing data with CoT prompting and incorporates IoT knowledge augmentation and retrieval mechanisms. This enables unified modeling and reasoning across various tasks, such as human activity recognition and industrial anomaly detection, significantly improving the depth of LLMs’ understanding of physical laws and sensor semantics \cite{an2024iot}.
For more complex IoT sensing scenarios, the IOT-LM framework employs a multimodal, multi-task perception encoder and instruction tuning mechanism to map multi-source heterogeneous sensor data into the input space of LLMs. This enables cross-modal knowledge fusion and shared task modeling, significantly enhancing the intelligence and generalization capabilities of IoT systems in multi-task recognition, interactive question answering, and real-time reasoning \cite{mo2024iot}.

In summary, the deep integration of LAMs with IoT not only significantly enhances semantic understanding and reasoning capabilities but also drives comprehensive upgrades in the intelligence, scalability, and security of IoT systems. This lays a solid foundation for building a new generation of IoT ecosystems characterized by cognitive capabilities, autonomous task execution, and natural human-machine interaction.

\subsubsection{LAMs for Edge Intelligence}
With the continuous advancement of LAMs, deploying them at the edge to support real-time, low-power, and privacy-preserving intelligent services is emerging as a key direction in the development of edge intelligence.

In edge training and inference optimization, to address the inefficiencies of federated learning in training LAMs at the edge, LLMs are utilized with forward gradient training and PEFT. This enables low-memory, on-device inference and training on mobile devices, supports Neural Processing Unit‌ (NPU) acceleration, and allows for multi-device parallelism. As a result, the efficiency of federated learning and the convergence speed of models are significantly improved, highlighting the critical role of LLMs in enhancing edge intelligence under resource-constrained conditions\cite{xu2023fwdllm}.
To address the resource constraints of edge devices, the Edge-LLM framework employs Layer-wise Unified Compression (LUC) and adaptive layer-tuning mechanisms to enable efficient fine-tuning and inference of LLMs on edge devices. This significantly reduces memory usage and computational complexity while maintaining performance, thereby improving deployment feasibility \cite{yu2024edge}.
Furthermore, the EdgeShard system partitions LLM inference tasks across multiple edge devices and the cloud, enabling fragmented processing of compute-intensive workloads. It introduces a joint device selection and model partitioning optimization algorithm, which markedly improves inference throughput and reduces latency, offering a new paradigm for LLM inference in collaborative edge environments \cite{zhang2024edgeshard}.
At the deployment level, the edge-device collaborative LLM system assigns the generation process to be jointly executed by the serial components on the device and the parallel components at the edge, thereby optimizing inference latency and energy consumption. Additionally, an integer programming algorithm is employed to allocate computational and transmission resources, reflecting a systematic design approach to device–edge collaborative optimization \cite{zhao2024edge}.

In edge architecture design, LLMs are incorporated into Mobile Edge Intelligence (MEI) through the MEI4LLM framework, which integrates techniques such as PEFT, distributed training, and inference. This framework enhances computational efficiency and resource utilization at the edge, effectively addressing challenges related to limited device capabilities and privacy protection. It demonstrates the critical role of LLMs in optimizing edge intelligence services \cite{qu2025mobile}.
Moreover, in decentralized deployment scenarios, LLMs enable distributed inference across energy-harvesting edge devices. By incorporating scheduling optimization and dynamic power control, they significantly improve energy efficiency and task throughput at the edge while reducing the risk of device failure. This highlights the critical role of LLMs in enhancing sustainable edge intelligence \cite{khoshsirat2024decentralized}.
From the perspective of 6G network evolution, LLMs support low-latency, on-device inference and training on 6G edge devices through techniques such as PEFT, quantization, and model partitioning. These methods effectively reduce resource consumption while improving system responsiveness and intelligent service capabilities, further underscoring the importance of LLMs in advancing edge intelligence \cite{lin2023pushing}.
In addition, a comprehensive survey emphasizes that LLM-powered edge intelligence architectures should encompass efficient deployment, security protection, and trustworthy development mechanisms. It also outlines key technical pathways including model compression, autonomous optimization, and cross-domain scenario adaptation, thereby providing a systematic panorama of LLM-driven edge intelligence development \cite{friha2024llm}.

In summary, LAMs are rapidly advancing their deep integration with edge intelligence, progressively overcoming computational bottlenecks, energy constraints, and communication overhead. This provides strong support for building low-latency, privacy-preserving, and highly generalizable edge intelligence systems.

\subsubsection{LAMs for Network Design and Management}
As 6G networks evolve toward greater intelligence and autonomy, traditional rule-based approaches to network design and management face limitations in flexibility and generalization. Leveraging their powerful capabilities in semantic understanding, intent recognition, task planning, and program generation, LLMs are increasingly being integrated into key components of network management architectures, facilitating the development of more efficient, intelligent, and versatile future networks.

In network intelligence design and management, LLMs play a pivotal role by enabling the ChatNet framework, which facilitates the automatic translation from natural language to network-specific language. This supports tasks such as network planning, configuration, and security policy formulation, thereby improving design efficiency and automation levels, and demonstrating the critical role of LLMs in enhancing intelligent network operations and management \cite{huang2024large}.
Building on this, the NetLM framework leverages LLMs to translate natural language intents into executable policies. By integrating multimodal representation learning and knowledge space construction, it provides intelligent support for resource scheduling, policy generation, and network configuration, enhancing the design efficiency and automation of 6G networks, while improving autonomy and flexibility in multi-task network management \cite{wang2023network}.
Moreover, LLMs enable the unified access and modeling of multi-source heterogeneous data for intelligent operations, resource scheduling, and configuration strategies. By combining natural language understanding, instruction generation, and knowledge augmentation techniques, they support intent parsing, fault diagnosis, and policy generation, thus improving the automation and generalization capabilities of 6G network management \cite{yue2023ai}.
In addition, LLMs support the automatic generation of configuration commands and network topology diagrams from natural language descriptions, enabling seamless transitions from textual input to executable network configurations. By incorporating prompt engineering techniques involving both textual and visual modalities, this approach significantly enhances response accuracy and design visibility, thereby improving the reliability and practicality of LLMs in network architecture design, generation, and configuration automation \cite{komanduri2025optimizing}.

In adaptive optimization for network management, LLMs exhibit program synthesis capabilities tailored to graph-structured tasks. They can directly generate high-quality graph manipulation code from natural language descriptions, providing more efficient interfaces for network lifecycle management and communication graph analysis \cite{mani2023enhancing}.
Compared to traditional approaches that rely on scenario modeling and parameter tuning, LLMs demonstrate CoT reasoning and In-Context Learning(ICL) in knowledge-free network management. By incorporating multi-model collaboration mechanisms, they support adaptive strategy generation for tasks such as resource scheduling and power control, even in the absence of prior scenario knowledge \cite{lee2024large}.
Through the NetLLM framework, LLMs achieve unified adaptation by employing multimodal encoders to process non-textual inputs (e.g., time-series data and graph structures), and task-specific output heads to directly respond to configuration tasks such as bitrate selection and scheduling strategies. This significantly enhances the efficiency and generalization ability of network task management \cite{wu2024netllm}.
Additionally, LLMs parse user intent and generate function-to-service mappings to collaboratively execute cross-domain network slicing resource configurations and task deployments. This enables intelligent end-to-end management from design to execution, greatly advancing the level of automation in network administration \cite{dandoush2024large}.
Furthermore, under the Network Algorithm Design Automation (NADA) via LLMs framework, LLMs are used to generate and optimize Adaptive Bitrate (ABR) algorithms by automatically designing state models and neural network structures. Coupled with prompt engineering and filtering mechanisms, this approach enables efficient algorithm customization and performance enhancement across diverse network environments such as 4G, 5G, and satellite networks. It significantly simplifies the development process of network algorithms and strengthens the intelligence and automation capabilities of network management \cite{he2024designing}.

In summary, LAMs are profoundly empowering network design and management tasks, driving the evolution of intelligent network systems toward higher autonomy, stronger generalization capabilities, and improved task adaptability across key dimensions such as network architecture, resource scheduling, configuration deployment, and algorithm generation.

\subsubsection{LAMs for Security and Privacy}
With the widespread deployment of LAMs in communication systems, network security, and user privacy protection are facing unprecedented challenges. Issues such as outsourced training, user data exposure, and model misuse necessitate that intelligent communication systems not only deliver efficient services but also possess robust security and privacy-preserving capabilities. In recent years, LAMs have demonstrated significant advantages in areas such as input encryption, adversarial attack defense, differential privacy mechanism design, and the construction of trusted execution paths, gradually emerging as a foundational pillar for the next generation of secure communication systems.

In the analysis of backdoor attacks and privacy leakage risks, the application of LLMs within communication networks has made them a novel vehicle for carrying out such attacks. Representative methods such as input interference, prompt manipulation, instruction injection, and demonstration contamination are classified based on the attack medium, triggering mechanism, and the intended target. By incorporating the unique characteristics of communication scenarios, the study further explores the concealment of these attacks and the difficulties in defending against them, offering valuable insights into enhancing the security of communication networks \cite{yang2024comprehensive}.
The deployment of LLMs in Zero touch network and Service Management (ZSM) environments raises urgent concerns regarding privacy leakage. Focusing on membership inference attacks, this study systematically evaluates the leakage risks and attack effectiveness of mainstream pre-trained models. It further proposes a mechanism that integrates trust evaluation modules, few-shot fine-tuning, and adversarial training, offering a concrete implementation pathway for ensuring the trustworthiness of LLM-based services in communication networks \cite{khowaja2024pathway}.
Moreover, the application of LLMs in network systems exhibits both offensive and defensive characteristics. On the positive side, they contribute to data integrity verification, anomaly detection, and multi-level privacy protection. However, they also face challenges such as hallucinated outputs, training data leakage, and vulnerability to adversarial examples. These issues highlight critical research directions for the development of trustworthy intelligent communication agents \cite{yao2024survey}.

In the context of privacy protection and secure deployment, the BlockChain for LLM (BC4LLM) framework enhances the attack resistance and trustworthiness of LLMs during distributed training by incorporating blockchain-based data ownership verification, identity authentication, and privacy-preserving computation. This framework enables end-to-end privacy protection and security enhancement for training data, learning processes, and generated content within communication networks \cite{luo2023bc4llm}.
LLMs can also achieve irreversible end-to-end data protection from input to inference by employing encrypted vocabulary reordering, geometric transformations in the embedding space, and client-side pre-encryption mechanisms. These techniques effectively prevent input leakage and the risk of model parameter reconstruction during communication, thereby enhancing security and privacy in cross-task inference scenarios \cite{mishra2024sentinellms}.
Within client–server communication frameworks, LLMs utilize activation-guided privacy restoration mechanisms and $d_X$-privacy-protected meta-vector construction to remove and reconstruct sensitive fragments in user inputs. This design prevents privacy leakage during data transmission while maintaining inference accuracy and computational efficiency \cite{zeng2024privacyrestore}.
In heterogeneous edge–cloud collaborative communication scenarios, LLMs can leverage PrivateLoRA by introducing low-rank residual transformation mechanisms and distributed parameter fine-tuning strategies. This approach enables a privacy-preserving information flow that transmits only non-reversible activation values and gradients, ensuring data locality while significantly reducing communication overhead. It facilitates the efficient deployment of privacy-enhanced generative services on mobile devices \cite{wang2023privatelora}.
When processing sensitive data, LLMs integrate user trust modeling, information sensitivity detection, and adaptive output control mechanisms. By applying Role-Based Access Control (RBAC), Attribute-Based Access Control (ABAC), differential privacy training, and semantic-level filtering, they enable output regulation based on user trust levels and strengthen privacy protection. This supports the development of intelligent inference systems that balance usability and security in communication-intensive domains such as healthcare and finance \cite{feretzakis2024trustworthy}.

In secure task collaboration and intelligent defense systems, LLMs demonstrate strong capabilities in task generalization, security knowledge modeling, and enhanced code generation. These abilities enable effective situational awareness and coordinated responses across a range of critical security tasks, including threat intelligence, vulnerability detection, program repair, and malicious behavior identification. Such integration offers a unified model support and a reference framework for intelligent defense and privacy protection in network environments \cite{zhang2025llms}.
Moreover, LLMs enhance system resilience against backdoors, data poisoning, and adversarial attacks through adversarial training, anomaly detection, and secure model updating mechanisms. By incorporating privacy-preserving technologies such as differential privacy, federated learning, and secure multi-party computation, they support the construction of end-to-end security defense systems across the inference process and model lifecycle in communication scenarios, laying the foundation for highly trustworthy intelligent systems \cite{zhang2025large}.

In summary, LAMs are continuously advancing communication networks from traditional security architectures toward intelligent perception, secure decision-making, and autonomous defense. This transformation is driven by a range of mechanisms, including input reconstruction protection, differential privacy, semantic filtering, secure model updating, and access control strategies.

\subsubsection{LAMs for Resource Allocation}
With the widespread deployment of wireless communication, edge computing, and multi-agent systems, resource allocation problems have become increasingly complex. Challenges such as high-dimensional non-convexity, dynamic environmental variations, and multi-objective coordination have exposed the limitations of traditional optimization methods. Leveraging their powerful semantic modeling and reasoning capabilities, LAMs have demonstrated significant advantages in resource allocation scenarios, emerging as a key driver for intelligent resource scheduling.

In channel awareness and power control optimization, a few-shot learning-based LLM inference framework is constructed to generate power allocation strategies using channel gain prompts. Combined with a binary power control mechanism to enhance robustness, this approach enables joint optimization of spectral and energy efficiency objectives \cite{lee2024llm}.
In intelligent network resource scheduling, a MoE architecture integrated with the GPT family of LLMs replaces traditional gating networks. By interpreting user goals from natural language inputs and selecting the optimal combination of experts, this method effectively supports adaptive optimization for resource allocation tasks such as power control, service selection, and load balancing, significantly improving decision-making efficiency and system flexibility in multi-task environments \cite{10592370}.
Through the LLM-OptiRA framework, LLMs perform modeling, transformation, and solution of non-convex resource allocation problems. By incorporating error correction and feasibility domain validation mechanisms, the framework enables efficient and robust resource scheduling in wireless communication systems \cite{peng2025llm}.

In edge-side collaborative resource scheduling, a RAG optimization framework is proposed by integrating the generative capabilities of LLMs with real-time information retrieval. Within MEC systems, this framework jointly schedules task offloading ratios, computing resources, and transmission power allocation, significantly enhancing the adaptability and interpretability of dynamic resource allocation decisions \cite{ren2024retrieval}.
In integrated space–air-ground networks, LLMs are utilized as cacheable resources. By incorporating the concept of cognitive age and RL-based auction mechanisms, the system coordinates the allocation of model storage, communication bandwidth, and computational capacity, thereby improving inference service timeliness and overall resource utilization \cite{xu2024cachedmodelasaresourceprovisioninglarge}.
In edge–cloud collaborative architectures, LLMs are scheduled using a multi-armed bandit algorithm driven by constraint satisfaction objectives. This approach enables dynamic resource allocation for diverse LLM-based services, significantly enhancing inference throughput and energy efficiency while maintaining latency constraints \cite{yang2024perllm}.
In MEC scenarios, LLM training tasks are distributed between users and edge servers through a hierarchical coordination strategy. By incorporating PEFT and stability-aware optimization objectives, the system achieves multi-objective joint resource allocation that balances training energy consumption, latency, and model reliability \cite{liu2024resource}.

In task-aware and adaptive computing resource allocation, LLMs introduce dynamic decision modules and KV-cache pruning strategies to adaptively skip Transformer layers based on task complexity and token importance. This enables on-demand allocation and substantial compression of both computational and memory resources \cite{jiang2024dllm}.
In multi-agent systems, LLMs autonomously perform task assignment and resource coordination by integrating task planning mechanisms with agent capability awareness strategies. This approach improves resource utilization and inference efficiency in collaborative multi-model environments \cite{amayuelas2025self}.

In summary, LAMs are driving the evolution of resource allocation mechanisms from static, rule-based approaches toward dynamic, semantics-driven strategies. These models effectively address a range of critical scenarios, including training scheduling, inference compression, model caching, and multi-agent coordination. As a result, they significantly enhance resource allocation efficiency and generalization capabilities under complex conditions, laying a solid foundation for the development of the next generation of intelligent and highly adaptive network systems.

\subsection{The Application Scenarios of Agentic AI}

\begin{figure*}[htpb]
		\centering
		\includegraphics[width=1\textwidth,height=0.5\textwidth]{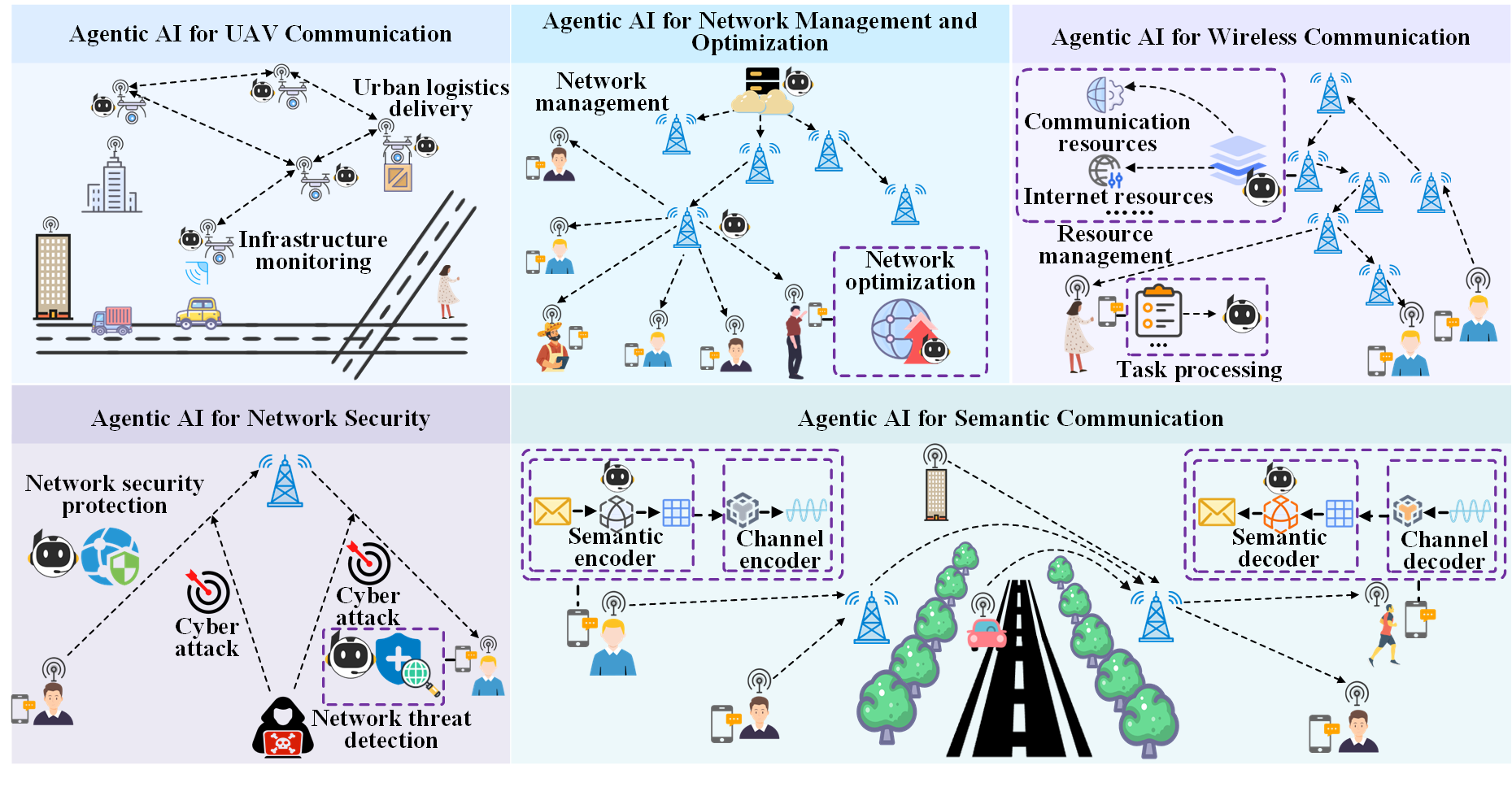}
		\caption{The application scenarios of Agentic AI.}
		\label{fig:fig7}
	\end{figure*}

\subsubsection{Agentic AI for Wireless Communication}
With the rapid development of wireless communication technologies, especially in the 6G era, the intelligence and automation of network architectures have become key factors in improving communication efficiency, optimizing network resources, and achieving effective management. By introducing agents into wireless communications, it is possible to realize more complex task collaboration, decision optimization, and network performance enhancement, thereby meeting the demands of 6G networks for efficient, intelligent, and personalized services.

In network optimization, utilizing LAMs as agents can significantly enhance the intelligence and optimization performance of networks. \textcolor{black}{In 6G networks, using LLMs as agents to collaboratively complete tasks and optimize network performance enables collective intelligence of multi-agent systems, thereby facilitating task decomposition and decision execution in edge environments \cite{zou2023wireless}}. Furthermore, applying LLMs for network operation optimization and health assessment provides more efficient fault diagnosis, resource scheduling, and load balancing, thus improving the intelligence level of 6G networks \cite{long20246g}. In wireless sensing systems, using LLMs as agents combined with Model Context Protocols (MCP) and expert systems enhances the perception and reasoning capabilities of LLMs in complex wireless communication environments, allowing more accurate resolution of dynamic issues in wireless networks \cite{liu2025model}.

Regarding task collaboration, employing LAMs as agents can effectively achieve task-oriented collaboration and automated workflows to drive task execution. In 6G networks, an AI agent architecture centered on LLMs utilizes key technologies such as multimodal collaboration, dynamic resource management, and edge computing to realize network automation, personalized services, and intelligent multi-device collaboration \cite{chen2024enabling}. Additionally, using LLMs as agents to realize task-oriented physical layer automation, a two-stage pre-training and fine-tuning scheme has been proposed to build specialized LLM agents adapted to different communication tasks, and a retrieval-based reasoning framework is employed to effectively invoke existing communication functions \cite{xiao2024llm}. In 6G networks, by leveraging LLMs as agents with perception, reasoning, and alignment modules, a split learning system has been designed to address device resource limitations, collaboratively complete complex tasks via edge computing, and ensure low latency and high-efficiency task execution \cite{xu2024large}.

In summary, using LAMs as agents can significantly enhance the intelligence level and task collaboration capabilities of wireless communication systems. These technologies provide more efficient and flexible solutions for 6G communication systems, driving communication networks toward greater intelligence and higher efficiency.

\subsubsection{Agentic AI for Semantic Communication}
With the rapid development of agent technologies in communications, semantic communication is gradually transitioning from the traditional bit transmission model to an intelligent communication paradigm centered on “semantics.” The introduction of agents provides semantic communication with adaptive capabilities and efficient data processing methods, enabling the system not only to handle traditional bitstream data but also to extract, transmit, and reconstruct semantic information, thereby improving the efficiency and accuracy of communication systems.

In semantic extraction and transmission, agents can dynamically adjust the encoding, transmission, and decoding methods based on the communication environment, user requirements, and information content, achieving optimal information transfer. In the Agent-driven Generative Semantic Communication (A-GSC) framework, agents use RL to adaptively adjust semantic sampling and extraction, enabling dynamic optimization of data transmission according to task demands \cite{10815060}. 
\textcolor{black}{In multi-user systems, leveraging LLMs as agents supports complex task decomposition, normalization of semantic representation, and semantic translation mapping, thereby improving communication efficiency in multi-user scenarios, optimizing computational resource management, and addressing the limitations of traditional semantic communication frameworks in 6G networks \cite{10972177}.} Moreover, in multi-user semantic communication, employing LLMs as agents to establish the Shared KB (SKB) enhances communication efficiency, and the proposed Multi-User Generative Semantic Communication (M-GSC) framework extends the capabilities of LLMs to handle complex multi-user tasks, optimize network resource usage, and overcome challenges in semantic encoding and decoding \cite{10972177}.

In resource allocation, agents can collaboratively optimize resources and communication strategies, improving network resource utilization and task transmission efficiency. In multi-cell semantic communication systems, agents dynamically optimize resource allocation through DRL, enhancing the overall network performance \cite{10845514}. \textcolor{black}{Additionally, in task-oriented semantic communication, a two-layer agent framework based on DRL jointly optimizes power, time slots, and semantic compression rates, achieving intelligent resource allocation under energy harvesting and cognitive radio environments, thereby improving user experience and spectrum utilization efficiency \cite{zhang2023toward}.}

In summary, agent-driven semantic communication achieves efficient information transmission by dynamically adjusting the encoding, transmission, and decoding of information. Meanwhile, with the help of DRL techniques, agents can optimize resource allocation and communication strategies, improving network resource utilization and task transmission efficiency, thereby advancing semantic communication systems toward greater intelligence and efficiency.

\subsubsection{Agentic AI for Network Management and Optimization}
With the application of agents in network management and optimization, management models are gradually evolving from traditional rule-based approaches toward more intelligent and automated paradigms. Agents are capable of autonomous learning, adapting to dynamic environments, and collaboratively completing complex tasks, significantly enhancing network management efficiency and adaptability.

In network management, leveraging LAMs as agents through autonomous decision-making and collaboration markedly improves the automation of network resource management. \textcolor{black}{In wireless networks, integrating LLMs as agents within multi-agent generative AI enables multiple agents to collaboratively plan and solve tasks to achieve network objectives. Through task planning, reasoning, and cooperative problem-solving, these agents can operate efficiently in edge networks, significantly elevating the intelligence level of network management \cite{zou2023wireless}. In 6G networks, LLMs as core agents work in coordination with other network components to provide powerful language understanding and generation capabilities for network health assessment and fault diagnosis, supporting automated and intelligent network operations \cite{long20246g}.} Meanwhile, in 6G-enabled Digital Twin (DT) networks, LLMs optimize data retrieval and Radio Resource Management (RRM) through intelligent reasoning and autonomous learning, realizing more automated and efficient network management \cite{jiang2024links}.

In network optimization, using LAM agents with collaboration and self-learning capabilities enables real-time optimization of network resource allocation, thereby enhancing network performance. The WirelessAgent framework employs LLMs to create autonomous Agentic AI that integrates four core modules—perception, memory, planning, and action—simulating human cognitive processes to manage complex wireless tasks. This enables LLMs to optimize resources in dynamic network environments, improving network resilience and performance \cite{tong2025wirelessagent}. Additionally, agents can autonomously adjust network resource configurations through RL and other methods, enhancing optimization efficiency and reliability \cite{xiao2025towards}. In Open Radio Access Network (O-RAN) systems, LLMs as agents dynamically optimize resource allocation to ensure network performance stability and adaptability. By analyzing network load and user demands in real time, LLMs automatically adjust resource allocation strategies to optimize resource utilization and improve reliability \cite{wu2025llm}.

In summary, the application of agents in network management and optimization greatly enhances network adaptability and intelligence. Through autonomous learning, collaboration, and task decomposition, agents make network management more efficient and network optimization more precise, driving the intelligent development of next-generation network systems.

\subsubsection{Agentic AI for Network Security}

With the continuous advancement of agent technology in network security, agent-driven security defense systems are gradually replacing traditional protective measures. Through autonomous learning, collaboration, and adaptation, agents can efficiently respond to complex security threats and provide real-time network protection and defense strategies.

Utilizing LAMs as agents enables efficient identification and response to network threats through real-time analysis and collaborative work, thereby enhancing the system’s security capabilities. \textcolor{black}{In network defense (such as blue team operations), employing LLMs as agents allows for real-time threat analysis, automatic generation of defense strategies, and dynamic adjustment of security measures based on network status and potential threats, significantly improving overall system security and defense capabilities. Particularly in 6G network environments, LLMs can effectively identify complex security threats and strengthen network security by generating corresponding defense strategies \cite{nguyen2024large}. In the scenario of Internet of Autonomous Defense Vehicles (IoADV), LLMs as core agents, leveraging powerful language understanding and generation capabilities along with multimodal data processing and predictive analytics, significantly enhance agent perception, decision-making, and real-time navigation in complex environments. They also play a key role in communication optimization and decision-making, improving the intelligence level of defense vehicles and their ability to operate in complex combat scenarios \cite{onsu2024leveraging}.}

Regarding defense, using LAMs as agents enables dynamic adjustment of security measures based on network conditions and potential threats, thereby strengthening overall system security. In Space-Air-Ground Integrated Networks (SAGIN) under zero-trust architecture, employing LLMs as agents facilitates collaboration among multiple LLM agents to perform security assessment and defense strategy generation via an LLM-based Situation Awareness (LLM-SA) method. This approach effectively handles vast heterogeneous threat information and, through adaptive learning and cooperative analysis, enhances network security defense capabilities, ensuring the system can rapidly respond and optimize defenses against complex and evolving attacks \cite{cao2025exploring}.

\textcolor{black}{In summary, leveraging LAMs as agents provides novel solutions for modern network security. Agents are capable not only of real-time response in complex security environments but also of optimizing defense strategies through collaboration, thereby offering robust guarantees for the long-term stability and security of networks.}

\subsubsection{Agentic AI for UAV Communication}

With the continuous advancement of agent technology in UAV communications, the architecture of aerial networks is gradually shifting from traditional centralized control to a novel structure centered on “distributed agent collaboration.” Agent systems leverage advanced mechanisms such as RL, state awareness, and behavior modeling, enabling UAVs to possess stronger task-driven capabilities, autonomous decision-making, and efficient collaboration. This provides intelligent support for resource scheduling and system control in complex and dynamic communication environments.

\textcolor{black}{In task execution of multi-UAV systems, agents improve UAVs’ reasoning and decision-making abilities through LAMs, allowing them to better adapt to complex environments and perform tasks. In Urban Air Mobility (UAM), the AirVista framework employs Multimodal LLMs (MLLMs) as agents and enhances UAVs’ 3D spatial reasoning and task execution efficiency by integrating Artificial systems, Computational experiments, and Parallel execution (ACP) methods. This framework particularly assists UAVs in urban environments to better accomplish complex tasks such as infrastructure monitoring, urban logistics delivery and security patrols \cite{lin2024airvista}. In UAV mission reliability assessment, a novel decision model combining LLMs and RAG techniques has been proposed. This enables UAVs to adapt in real time within complex dynamic environments, improving decision accuracy and responsiveness in multi-UAV networks, thereby enhancing system reliability \cite{sezgin2025scenario}.}

Regarding multi-agent collaboration and task planning optimization, agents utilize LAMs to optimize collaboration and task planning among multiple UAVs, improving task completion rates and system efficiency. In UAV-assisted edge computing environments, combining LLMs with a Multi-Agent DRL (MADRL) framework, and introducing the QTRAN algorithm, optimizes task offloading and trajectory planning. This method effectively addresses the correlation between local observations and global states in multi-UAV systems, thereby improving task completion rates and convergence speed \cite{zhu2024task}. In UAV task generation and planning, the UAV-CodeAgents framework employs LLMs and Vision-Language Models (VLMs), combined with the ReAct (Reason + Act) paradigm, enabling UAVs to plan tasks based on satellite imagery and natural language instructions, precisely locate targets, and dynamically adjust task goals, thereby improving task execution efficiency and adaptability \cite{sautenkov2025uav}.

In summary, agent technology assisted by LAMs significantly enhances UAVs’ capabilities in task execution, decision-making, multi-agent collaboration, and task planning. These advances not only drive the development of UAV technology but also provide intelligent support for the efficient operation of future communication systems in dynamic environments.

\subsection{Summary and Lessons Learned}
\subsubsection{Summary}
This chapter summarizes the typical application scenarios of LAMs and Agentic AI in communication systems. LAMs are primarily used in semantic communication, IoT, edge intelligence, network design and management, security and privacy, and resource allocation, while Agentic AI is widely applied in wireless communication, semantic communication, network management and optimization, network security, and UAV communication. Their integration accelerates the intelligent and autonomous evolution of communication systems.

\subsubsection{Lessons Learned}
\textcolor{black}{Although LAMs and Agentic AI have demonstrated great potential in semantic communication, IoT, edge intelligence, network management, and security protection, current research still faces challenges such as complex model deployment, high resource consumption, insufficient multimodal understanding capabilities, and security risks. Future research should focus on lightweight model design, cross-modal semantic fusion, system collaborative optimization, and privacy protection mechanisms to promote the continuous evolution of communication systems toward efficient, secure, and adaptive intelligent architectures \cite{jiang2025comprehensive}.}


\section{Research Challenges and Future Directions}
\subsection{Research Challenges and Directions of LAMs}
\subsubsection{Untimely Updating and Learning of Communication Data}
The lack of high-quality communication data represents a critical challenge limiting the development of LAMs in communications. Due to the rapid evolution of communication technologies and the vast, complex nature of domain-specific knowledge, LAMs must continuously learn and adapt. However, obtaining timely and reliable communication data remains highly difficult. On one hand, relevant knowledge is scattered across research papers, standards, and patents, making data collection and processing costly. On the other hand, real-world communication data is highly dynamic and often involves privacy-sensitive or confidential information, restricting its availability for public training. Furthermore, data in 6G systems is frequently affected by noise, incompleteness, and errors, requiring models to exhibit strong robustness and fault tolerance. The difficulty in data acquisition and the lack of timely learning have become major bottlenecks impeding the performance improvement and practical deployment of LAMs in communications.

To address the challenge of data scarcity faced by LAMs in communications, continual learning offers an effective development pathway. By incorporating autonomous continual learning strategies \cite{liu2023ai}, LAMs are able to independently acquire and update communication knowledge in dynamic environments. Multimodal continual learning \cite{yu2024recent} further enhances their ability to process diverse communication data across modalities such as text, images, and audio, thereby expanding the scope of knowledge acquisition. Robust continual learning \cite{ru2024maintaining} improves model stability and adaptability in complex communication scenarios. When combined with evaluation metrics such as forward transfer, backward transfer, and forgetting measures \cite{chaudhry2018riemannian}, the learning process can be effectively monitored and optimized. The integration of continual learning significantly enhances LAMs’ capabilities in knowledge acquisition, adaptation, and reasoning for communication tasks, thereby mitigating the challenges posed by delayed updates in communication data.

\subsubsection{Insufficient Reasoning Capabilities}
The limited logical reasoning capabilities of LAMs have emerged as a significant challenge for their application in 6G systems \cite{friedman2023large}. As 6G communication environments require models to handle complex tasks such as signal processing and resource allocation, the ability to perform accurate logical and causal reasoning becomes essential. However, current LAMs are primarily data-driven and lack a deep understanding of causal relationships, making them inadequate for handling multi-hop and counterfactual reasoning. This often results in the generation of inaccurate or illogical solutions in complex communication scenarios, thereby compromising system stability and user experience. For instance, an LAM may fail to recognize that network congestion is caused by a sudden surge in users or that signal degradation stems from channel fading. Moreover, the inherent ambiguity of natural language and the limitations of formal logic further constrain the reasoning capabilities of LAMs. Enhancing logical understanding and reasoning in communication-specific contexts is a critical direction for future research.

To address the insufficient reasoning capabilities of LAMs, future advancements may focus on three key directions: process-based reward models\cite{lightman2023let}, long-chain reasoning mechanisms\cite{yao2024tree}, and RL-driven reasoning training\cite{xi2024training}. Process-based reward models provide feedback at each step of the reasoning process, guiding the model to prioritize the coherence and validity of its reasoning paths. Long-chain reasoning mechanisms help maintain logical consistency across multi-step tasks, thereby enhancing the depth of reasoning in complex problem settings. RL-driven reasoning training leverages environmental feedback to enable the model to iteratively refine its reasoning strategies through trial and error, learning more effective reasoning trajectories from experience and improving generalization to novel tasks. The synergistic integration of these three approaches is expected to significantly enhance the logical reasoning and causal inference capabilities of LAMs in communications.

\subsubsection{Inadequate Explanation}
Current LAMs exhibit significant limitations in interpretability, which has become a critical barrier to their application and further development. As complex black-box systems, LAMs lack effective mechanisms to reveal the underlying logic behind their decisions, resulting in limited transparency and controllability in communication tasks. Whether addressing signal processing, resource allocation, or managing diverse communication protocols and network scenarios, LAMs often fail to clearly articulate the rationale behind their outputs, thereby constraining user understanding and system-level optimization. The absence of both local and global interpretability not only undermines model trustworthiness but may also pose risks to the security and stability of communication systems. Therefore, enhancing interpretability and developing systematic explanation frameworks are essential directions for enabling the reliable deployment of LAMs in communications.

To address the issue of limited interpretability in LAMs, future efforts may leverage techniques such as causal learning \cite{9709543}, neuro-symbolic integration \cite{wagner2021neural}, feature attribution \cite{ribeiro2016should}, and model visualization \cite{bau2018gan} to enhance model transparency. Causal learning facilitates the identification of underlying causal relationships behind model decisions, improving logical clarity; neuro-symbolic integration combines the strengths of neural networks and symbolic reasoning, enabling more traceable inference processes; feature attribution highlights the key input features influencing model predictions, helping users understand the basis of the output; and model visualization offers intuitive insights into the model’s internal mechanisms, increasing user comprehension and trust. Together, these approaches contribute to making LAMs more interpretable and controllable in communication tasks.

\subsubsection{Difficulties in Deployment of LAMs}
LAMs face significant challenges in real-world deployment, primarily due to constraints in hardware and communication resources. Current communication devices, particularly edge devices such as mobile terminals and IoT nodes, often have limited computational and storage capabilities, which makes it challenging to meet the high computational demands of LAMs. This results in performance degradation and increased deployment costs. Additionally, the integration of LAMs increases the volume of communication data, especially in semantic communication scenarios where the synchronization of model parameters and knowledge bases imposes higher demands on network bandwidth and latency. The scarcity of spectrum resources further exacerbates the complexity of data compression and reconstruction. Therefore, improving model efficiency under resource-constrained conditions and advancing lightweight LAM design and compression techniques are critical for enabling the practical deployment of LAMs.

To address the resource constraints associated with deploying LAMs, future efforts should focus on developing efficient model compression and acceleration techniques to reduce computational complexity and communication overhead. Pruning methods \cite{zhu2024survey} can eliminate redundant structures, thereby reducing model size and computational demands. Quantization techniques \cite{xiao2023smoothquant} compress high-precision parameters into lower-bit representations, minimizing memory usage and inference latency, particularly when combined with activation-aware mechanisms that help maintain high accuracy. Knowledge Distillation (KD) \cite{xu2024survey} transfers knowledge from a large model to a smaller one, enabling lightweight deployment while preserving core capabilities, making it especially suitable for edge environments. The integration of these techniques offers a practical pathway for the efficient deployment of LAMs in 6G communications.

\subsection{Research Challenges and Directions of Agentic AI}
\subsubsection{The Lack of Communication Knowledge}
In the development of Agentic AI, the lack of communication knowledge presents a significant challenge to its application in 6G systems. Due to the complexity and rapid evolution of communication technologies, agent systems rely heavily on high-quality communication knowledge bases to support task understanding and reasoning-based decision-making. However, current knowledge bases offer insufficient coverage of core concepts, protocols, and standards in communications, leading to misinterpretations and inaccurate reasoning in tasks such as channel estimation and resource allocation. Moreover, communication knowledge is dispersed across research papers, standards, and patents, making its collection and integration both challenging and labor-intensive, requiring domain expertise to ensure quality. While expanding the knowledge base can enhance model capabilities, it also introduces higher storage and computational costs, potentially reducing system efficiency. Therefore, constructing an efficient and comprehensive communication knowledge infrastructure is a key direction for enabling the successful deployment of Agentic AI in communications.

To address the issue of communication knowledge scarcity in Agentic AI systems for 6G communications, the introduction of the Agentic RAG mechanism \cite{singh2025agentic} offers a promising solution. This approach integrates agent collaboration with semantic retrieval techniques, enabling the system to dynamically extract relevant knowledge from communication literature, standards, and patents in real-time based on the task context. The retrieved information is then synthesized by agents to generate context-aware outputs, effectively overcoming the limitations of static knowledge base coverage. Agentic RAG not only enhances the system’s ability to understand and reason over complex communication protocols and dynamic environments but also supports continual knowledge updates and transfer learning, thereby improving the system’s adaptability and intelligent decision-making capabilities in communication scenarios.

\subsubsection{Limited Scalability of Agentic AI}
Agentic AI faces significant scalability challenges when deployed in large-scale, multi-task, or multi-agent collaborative communication environments. Most existing systems still rely on centralized control architectures, where a central model manages task scheduling and resource allocation. While effective in small-scale scenarios, such architectures are prone to resource bottlenecks, latency, and even deadlocks under high concurrency or multi-system collaboration, thereby limiting overall system performance. Additionally, current multi-agent coordination mechanisms lack distributed scheduling and autonomous capabilities. Communication and task allocation among agents often depend on static configurations, which are insufficiently flexible or adaptive to handle heterogeneous tasks and dynamic environments. For example, in complex task chains, a single agent failure can lead to system-wide cascading failures, and conventional systems lack the mechanisms for rapid recovery or adaptive restructuring. These limitations expose critical weaknesses in scalability and robustness under dynamic environments.

To address the scalability limitations of Agentic AI, future efforts should focus on developing distributed and hierarchical architectures \cite{yang2025agentnet}\cite{saleh2025usercentrix}, enhancing multi-agent collaboration mechanisms\cite{tran2025multi}, and optimizing runtime resource scheduling. Architecturally, the integration of distributed control with hierarchical structures allows agents at different levels to take on decision-making, coordination, and execution roles, thereby improving the system’s parallel processing capabilities and response efficiency. Regarding coordination mechanisms, incorporating federated learning enables knowledge sharing among agents, while swarm intelligence techniques facilitate self-organizing collaboration based on local observations, enhancing system flexibility and adaptability. At the operational level, the use of self-optimizing algorithms and load-balancing strategies allows the system to dynamically adjust resource allocation and task routing, effectively mitigating bottlenecks and improving overall stability and efficiency. The synergy of these approaches provides a robust foundation for building efficient, stable, and scalable Agentic AI systems.

\subsubsection{Complexity of Agent Control Mechanisms}
In Agentic AI systems, the complexity of coordination among agents and control interactions between agents and other system components presents a major research challenge. As the number of agents increases and task environments evolve dynamically, the system faces significant pressure in task decomposition, information exchange, and behavioral coordination. These challenges often lead to issues such as component scheduling conflicts and communication inconsistencies, resulting in low collaboration efficiency, high task execution latency, and difficulty in meeting the real-time, reliability, and adaptability demands of complex scenarios. Moreover, existing systems often lack standardized collaboration protocols and structured control workflows, making it difficult to support efficient organization and dynamic adaptation among large-scale heterogeneous agents. The immaturity of current control mechanisms has become a core bottleneck limiting the broader application of Agentic AI in communication systems.

To address the complexity of agent control mechanisms in Agentic AI systems, three core Protocols, namely MCP \cite{krishnan2025advancing}, Agent-to-Agent Protocol (A2A) \cite{habler2025building}, and ACP\cite{ehtesham2025survey}, offer a layered and complementary technical pathway. MCP standardizes the interaction between models and external resources, enhancing system controllability and modularity. A2A enables dynamic discovery and task collaboration among agents through mechanisms such as agent cards and the task–artifact framework, thereby simplifying cross-system coordination. ACP, built upon a REST-native architecture, supports multimodal asynchronous communication and task tracking, strengthening collaborative capabilities among agents within local environments. Together, these protocols establish a secure, flexible, and scalable agent control infrastructure.

\subsubsection{Difficulty in Evaluating Agentic AI}
Current Agentic AI systems face significant challenges in evaluation, primarily due to the absence of unified and systematic assessment frameworks, which makes it difficult to comprehensively reflect agent capabilities in dynamic task settings. Existing approaches often rely on static datasets and outcome-oriented single metrics, overlooking the agent’s performance throughout multi-step reasoning, tool invocation, and strategy planning processes, thereby failing to identify latent issues. Moreover, the inherent diversity and uncertainty of agent behavior make traditional evaluation methods based on fixed reference answers inadequate for assessing flexibility and generalization capability. This limitation significantly hinders accurate performance validation and system optimization \cite{zhuge2024agent}.

To address the evaluation challenges of Agentic AI, future research should focus on developing a unified and scalable evaluation framework capable of assessing agent performance at a fine-grained level across task planning, tool invocation, and reasoning processes. The evaluation paradigm should shift from outcome-based metrics to process-oriented assessment to enhance understanding of the agent's behavioral trajectory. Additionally, automated evaluation methods such as LAM-based evaluators should be incorporated to reduce human effort and enhance evaluation efficiency. It is also essential to develop general-purpose evaluation tools that can accommodate the diversity and uncertainty of agent behaviors, thereby providing robust support for system optimization and reliable deployment \cite{yehudai2025survey}.

\subsection{Chapter Summary}
This chapter provides a comprehensive summary of the key research challenges and future directions for LAMs and Agentic AI in future intelligent communication systems. It systematically reviews the core limitations hindering their performance and deployment, along with potential solutions. For LAMs, we highlight major challenges such as delayed communication data updates and learning, insufficient reasoning capabilities, limited interpretability, and deployment difficulties. We propose that these issues can be addressed through the adoption of techniques such as continual learning, long-chain reasoning, explainable AI, and model compression and distillation, thereby enhancing the model’s capabilities in knowledge acquisition, logical reasoning, interpretability, and edge deployment in dynamic communication environments.
 For Agentic AI, we summarize the challenges related to insufficient coverage of communication knowledge, weak system scalability, complex agent control mechanisms, and the absence of robust evaluation methodologies. We emphasize that advancing Agentic RAG mechanisms guided by dynamic knowledge, distributed control architectures, unified control protocols (MCP, A2A, ACP), and process-oriented evaluation frameworks will be crucial for future development. Collectively, these insights offer a systematic analysis and forward-looking guidance to drive intelligent communication systems toward greater autonomy, interpretability, and practical utility.

\section{Conclusion}
This tutorial provides a systematic review of the development trajectory and key technological pathways from LAMs to Agentic AI in future intelligent communication systems. First, we present a comprehensive overview of the core components and classification methods of LAMs, covering models such as Transformer, ViT, VAE, Diffusion, DiT, and MoE, and differentiating the applicability of LLMs, LVMs, LMMs, LRMs, and lightweight LAMs in communication tasks. We then propose a LAM construction paradigm tailored for communication systems, encompassing three critical aspects: dataset construction, internal training mechanisms (e.g., pre-training, fine-tuning, and alignment), and external learning mechanisms (e.g., RAG and KG), to guide effective model learning in communication scenarios. Building on this foundation, we construct a LAM-based Agentic AI system framework by defining its core modules, including planners, knowledge bases, tools, and memory modules, and by outlining the interaction mechanisms for both single-agent and multi-agent settings. We further propose a multi-agent system for data retrieval, collaborative planning, and reflective evaluation. At the application level, we systematically summarize the practical value and potential of LAMs and Agentic AI in key communication tasks, including semantic communication, the IoT, edge intelligence, network management, network security, UAV communication, and other emerging applications. Finally, we identify the core challenges currently facing LAMs and Agentic AI in communications and outline future research directions. This work offers a systematic reference and theoretical foundation to support the evolution of intelligent communication systems from model-driven to agent-driven paradigms.

\bibliographystyle{IEEEtran}
\bibliography{ref}

\begin{thebibliography}{100}
\providecommand{\url}[1]{#1}
\csname url@samestyle\endcsname
\providecommand{\newblock}{\relax}
\providecommand{\bibinfo}[2]{#2}
\providecommand{\BIBentrySTDinterwordspacing}{\spaceskip=0pt\relax}
\providecommand{\BIBentryALTinterwordstretchfactor}{4}
\providecommand{\BIBentryALTinterwordspacing}{\spaceskip=\fontdimen2\font plus
\BIBentryALTinterwordstretchfactor\fontdimen3\font minus
  \fontdimen4\font\relax}
\providecommand{\BIBforeignlanguage}[2]{{%
\expandafter\ifx\csname l@#1\endcsname\relax
\typeout{** WARNING: IEEEtran.bst: No hyphenation pattern has been}%
\typeout{** loaded for the language `#1'. Using the pattern for}%
\typeout{** the default language instead.}%
\else
\language=\csname l@#1\endcsname
\fi
#2}}
\providecommand{\BIBdecl}{\relax}
\BIBdecl

\bibitem{zhang20196g}
Z.~Zhang, Y.~Xiao, Z.~Ma, M.~Xiao, Z.~Ding, X.~Lei, G.~K. Karagiannidis, and
  P.~Fan, ``6g wireless networks: Vision, requirements, architecture, and key
  technologies,'' \emph{IEEE vehicular technology magazine}, vol.~14, no.~3,
  pp. 28--41, 2019.

\bibitem{kenton2019bert}
J.~D. M.-W.~C. Kenton and L.~K. Toutanova, ``Bert: Pre-training of deep
  bidirectional transformers for language understanding,'' in \emph{Proceedings
  of naacL-HLT}, vol.~1, 2019, p.~2.

\bibitem{radford2018improving}
A.~Radford, ``Improving language understanding by generative pre-training,''
  2018.

\bibitem{radford2019language}
A.~Radford, J.~Wu, R.~Child, D.~Luan, D.~Amodei, I.~Sutskever \emph{et~al.},
  ``Language models are unsupervised multitask learners,'' \emph{OpenAI blog},
  vol.~1, no.~8, p.~9, 2019.

\bibitem{brown2020language}
T.~B. Brown, ``Language models are few-shot learners,'' \emph{arXiv preprint
  arXiv:2005.14165}, 2020.

\bibitem{raffel2020exploring}
C.~Raffel, N.~Shazeer, A.~Roberts, K.~Lee, S.~Narang, M.~Matena, Y.~Zhou,
  W.~Li, and P.~J. Liu, ``Exploring the limits of transfer learning with a
  unified text-to-text transformer,'' \emph{Journal of machine learning
  research}, vol.~21, no. 140, pp. 1--67, 2020.

\bibitem{ouyang2022training}
L.~Ouyang, J.~Wu, X.~Jiang, D.~Almeida, C.~Wainwright, P.~Mishkin, C.~Zhang,
  S.~Agarwal, K.~Slama, A.~Ray \emph{et~al.}, ``Training language models to
  follow instructions with human feedback,'' \emph{Advances in neural
  information processing systems}, vol.~35, pp. 27\,730--27\,744, 2022.

\bibitem{kirillov2023segment}
A.~Kirillov, E.~Mintun, N.~Ravi, H.~Mao, C.~Rolland, L.~Gustafson, T.~Xiao,
  S.~Whitehead, A.~C. Berg, W.-Y. Lo \emph{et~al.}, ``Segment anything,'' in
  \emph{Proceedings of the IEEE/CVF International Conference on Computer
  Vision}, Paris, France, October 2023, pp. 4015--4026.

\bibitem{achiam2023gpt}
J.~Achiam, S.~Adler, S.~Agarwal, L.~Ahmad, I.~Akkaya, F.~L. Aleman, D.~Almeida,
  J.~Altenschmidt, S.~Altman, S.~Anadkat \emph{et~al.}, ``Gpt-4 technical
  report,'' \emph{arXiv preprint arXiv:2303.08774}, 2023.

\bibitem{geminiteam2024geminifamilyhighlycapable}
\BIBentryALTinterwordspacing
Team, Gemini \emph{et~al.}, ``Gemini: A family of highly capable multimodal
  models,'' 2024. [Online]. Available: \url{https://arxiv.org/abs/2312.11805}
\BIBentrySTDinterwordspacing

\bibitem{hayawi2024cross}
K.~Hayawi and S.~Shahriar, ``A cross-domain performance report of open ai
  chatgpt o1 model,'' 2024.

\bibitem{guo2025deepseek}
D.~Guo, D.~Yang, H.~Zhang, J.~Song, R.~Zhang, R.~Xu, Q.~Zhu, S.~Ma, P.~Wang,
  X.~Bi \emph{et~al.}, ``Deepseek-r1: Incentivizing reasoning capability in
  llms via reinforcement learning,'' \emph{arXiv preprint arXiv:2501.12948},
  2025.

\bibitem{yang2023auto}
H.~Yang, S.~Yue, and Y.~He, ``Auto-gpt for online decision making: Benchmarks
  and additional opinions,'' \emph{arXiv preprint arXiv:2306.02224}, 2023.

\bibitem{nakajima2023babyagi}
Y.~Nakajima, ``Babyagi,'' \emph{GitHub repository}, 2023.

\bibitem{xie2023openagents}
T.~Xie, F.~Zhou, Z.~Cheng, P.~Shi, L.~Weng, Y.~Liu, T.~J. Hua, J.~Zhao, Q.~Liu,
  C.~Liu \emph{et~al.}, ``Openagents: An open platform for language agents in
  the wild,'' \emph{arXiv preprint arXiv:2310.10634}, 2023.

\bibitem{ferrag2025llm}
M.~A. Ferrag, N.~Tihanyi, and M.~Debbah, ``From llm reasoning to autonomous ai
  agents: A comprehensive review,'' \emph{arXiv preprint arXiv:2504.19678},
  2025.

\bibitem{acharya2025agentic}
D.~B. Acharya, K.~Kuppan, and B.~Divya, ``Agentic ai: Autonomous intelligence
  for complex goals--a comprehensive survey,'' \emph{IEEE Access}, 2025.

\bibitem{gridach2025agentic}
M.~Gridach, J.~Nanavati, K.~Z.~E. Abidine, L.~Mendes, and C.~Mack, ``Agentic ai
  for scientific discovery: A survey of progress, challenges, and future
  directions,'' \emph{arXiv preprint arXiv:2503.08979}, 2025.

\bibitem{jiang2025comprehensive}
F.~Jiang, C.~Pan, L.~Dong, K.~Wang, M.~Debbah, D.~Niyato, and Z.~Han, ``A
  comprehensive survey of large ai models for future communications:
  Foundations, applications and challenges,'' \emph{arXiv preprint
  arXiv:2505.03556}, 2025.

\bibitem{zhou2024large}
H.~Zhou, C.~Hu, Y.~Yuan, Y.~Cui, Y.~Jin, C.~Chen, H.~Wu, D.~Yuan, L.~Jiang,
  D.~Wu \emph{et~al.}, ``Large language model (llm) for telecommunications: A
  comprehensive survey on principles, key techniques, and opportunities,''
  \emph{arXiv preprint arXiv:2405.10825}, 2024.

\bibitem{boateng2025survey}
G.~O. Boateng, H.~Sami, A.~Alagha, H.~Elmekki, A.~Hammoud, R.~Mizouni,
  A.~Mourad, H.~Otrok, J.~Bentahar, S.~Muhaidat \emph{et~al.}, ``A survey on
  large language models for communication, network, and service management:
  Application insights, challenges, and future directions,'' \emph{IEEE
  Communications Surveys \& Tutorials}, 2025.

\bibitem{chen2024big}
Z.~Chen, Z.~Zhang, and Z.~Yang, ``Big ai models for 6g wireless networks:
  Opportunities, challenges, and research directions,'' \emph{IEEE Wireless
  Communications}, vol.~31, no.~5, pp. 164--172, 2024.

\bibitem{huang2024large}
Y.~Huang, H.~Du, X.~Zhang, D.~Niyato, J.~Kang, Z.~Xiong, S.~Wang, and T.~Huang,
  ``Large language models for networking: Applications, enabling techniques,
  and challenges,'' \emph{IEEE Network}, vol.~39, no.~1, pp. 235--242, 2025.

\bibitem{chowdhury20206g}
M.~Z. Chowdhury, M.~Shahjalal, S.~Ahmed, and Y.~M. Jang, ``6g wireless
  communication systems: Applications, requirements, technologies, challenges,
  and research directions,'' \emph{IEEE Open Journal of the Communications
  Society}, vol.~1, pp. 957--975, 2020.

\bibitem{jiang2024large3}
F.~Jiang, Y.~Peng, L.~Dong, K.~Wang, K.~Yang, C.~Pan, D.~Niyato, and O.~A.
  Dobre, ``Large language model enhanced multi-agent systems for 6g
  communications,'' \emph{IEEE Wireless Communications}, vol.~31, no.~6, pp.
  48--55, 2024.

\bibitem{jiang2024large2}
F.~Jiang, Y.~Peng, L.~Dong, K.~Wang, K.~Yang, C.~Pan, and X.~You, ``Large ai
  model-based semantic communications,'' \emph{IEEE Wireless Communications},
  vol.~31, no.~3, pp. 68--75, 2024.

\bibitem{10495599}
J.~Zhong, M.~Li, Y.~Chen, Z.~Wei, F.~Yang, and H.~Shen, ``A safer vision-based
  autonomous planning system for quadrotor uavs with dynamic obstacle
  trajectory prediction and its application with llms,'' in \emph{2024 IEEE/CVF
  Winter Conference on Applications of Computer Vision Workshops (WACVW)},
  2024, pp. 920--929.

\bibitem{10384606}
Y.~Shen, J.~Shao, X.~Zhang, Z.~Lin, H.~Pan, D.~Li, J.~Zhang, and K.~B. Letaief,
  ``Large language models empowered autonomous edge ai for connected
  intelligence,'' \emph{IEEE Communications Magazine}, vol.~62, no.~10, pp.
  140--146, 2024.

\bibitem{wu2025agentic}
J.~Wu, J.~Zhu, and Y.~Liu, ``Agentic reasoning: Reasoning llms with tools for
  the deep research,'' \emph{arXiv preprint arXiv:2502.04644}, 2025.

\bibitem{vaswani2017attention}
A.~Vaswani, N.~Shazeer, N.~Parmar, J.~Uszkoreit, L.~Jones, A.~N. Gomez,
  {\L}.~Kaiser, and I.~Polosukhin, ``Attention is all you need,''
  \emph{Advances in neural information processing systems}, vol.~30, 2017.

\bibitem{yang2023witt}
K.~Yang, S.~Wang, J.~Dai, K.~Tan, K.~Niu, and P.~Zhang, ``Witt: A wireless
  image transmission transformer for semantic communications,'' in \emph{ICASSP
  2023-2023 IEEE International Conference on Acoustics, Speech and Signal
  Processing (ICASSP)}.\hskip 1em plus 0.5em minus 0.4em\relax IEEE, 2023, pp.
  1--5.

\bibitem{leng2025unveiling}
Y.~Leng, Q.~Lin, L.-Y. Yung, J.~Lei, Y.~Li, and Y.-C. Wu, ``Unveiling the power
  of complex-valued transformers in wireless communications,'' \emph{arXiv
  preprint arXiv:2502.11151}, 2025.

\bibitem{tian2023multimodal}
Y.~Tian, Q.~Zhao, F.~Boukhalfa, K.~Wu, F.~Bader \emph{et~al.}, ``Multimodal
  transformers for wireless communications: A case study in beam prediction,''
  \emph{arXiv preprint arXiv:2309.11811}, 2023.

\bibitem{zhang2025decision}
J.~Zhang, J.~Li, Z.~Wang, L.~Shi, S.~Jin, W.~Chen, and H.~V. Poor, ``Decision
  transformers for wireless communications: A new paradigm of resource
  management,'' \emph{IEEE Wireless Communications}, 2025.

\bibitem{dosovitskiy2020image}
A.~Dosovitskiy, L.~Beyer, A.~Kolesnikov, D.~Weissenborn, X.~Zhai,
  T.~Unterthiner, M.~Dehghani, M.~Minderer, G.~Heigold, S.~Gelly \emph{et~al.},
  ``An image is worth 16x16 words: Transformers for image recognition at
  scale,'' \emph{arXiv preprint arXiv:2010.11929}, 2020.

\bibitem{mohsin2025vision}
M.~A. Mohsin, M.~Jazib, Z.~Alam, M.~F. Khan, M.~Saad, and M.~A. Jamshed,
  ``Vision transformer based semantic communications for next generation
  wireless networks,'' \emph{arXiv preprint arXiv:2503.17275}, 2025.

\bibitem{gharsallah2024vit}
G.~Gharsallah and G.~Kaddoum, ``Vit los v2x: Vision transformers for
  environment-aware los blockage prediction for 6g vehicular networks,''
  \emph{IEEE Access}, 2024.

\bibitem{zheng2024fe}
G.~Zheng, B.~Zang, P.~Yang, W.~Zhang, and B.~Li, ``Fe-skvit: A feature-enhanced
  vit model with skip attention for automatic modulation recognition.''
  \emph{Remote Sensing}, vol.~16, no.~22, 2024.

\bibitem{kingma2013auto}
D.~P. Kingma and M.~Welling, ``Auto-encoding variational bayes,'' \emph{arXiv
  preprint arXiv:1312.6114}, 2013.

\bibitem{hussien2022prvnet}
M.~Hussien, K.~K. Nguyen, and M.~Cheriet, ``Prvnet: A novel
  partially-regularized variational autoencoders for massive mimo csi
  feedback,'' in \emph{2022 IEEE wireless communications and networking
  conference (WCNC)}.\hskip 1em plus 0.5em minus 0.4em\relax IEEE, 2022, pp.
  2286--2291.

\bibitem{bo2024joint}
Y.~Bo, Y.~Duan, S.~Shao, and M.~Tao, ``Joint coding-modulation for digital
  semantic communications via variational autoencoder,'' \emph{IEEE
  Transactions on Communications}, 2024.

\bibitem{omondi2023variational}
G.~Omondi and T.~O. Olwal, ``Variational autoencoder-enhanced deep neural
  network-based detection for mimo systems,'' \emph{e-Prime-Advances in
  Electrical Engineering, Electronics and Energy}, vol.~6, p. 100335, 2023.

\bibitem{sohl2015deep}
J.~Sohl-Dickstein, E.~Weiss, N.~Maheswaranathan, and S.~Ganguli, ``Deep
  unsupervised learning using nonequilibrium thermodynamics,'' in
  \emph{International conference on machine learning}.\hskip 1em plus 0.5em
  minus 0.4em\relax Lille, France: PMLR, July 2015, pp. 2256--2265.

\bibitem{rombach2022high}
R.~Rombach, A.~Blattmann, D.~Lorenz, P.~Esser, and B.~Ommer, ``High-resolution
  image synthesis with latent diffusion models,'' in \emph{Proceedings of the
  IEEE/CVF conference on computer vision and pattern recognition}, New Orleans,
  LA, USA, June 2022, pp. 10\,684--10\,695.

\bibitem{ramesh2022hierarchical}
A.~Ramesh, P.~Dhariwal, A.~Nichol, C.~Chu, and M.~Chen, ``Hierarchical
  text-conditional image generation with clip latents,'' \emph{arXiv preprint
  arXiv:2204.06125}, vol.~1, no.~2, p.~3, 2022.

\bibitem{betker2023improving}
J.~Betker, G.~Goh, L.~Jing, T.~Brooks, J.~Wang, L.~Li, L.~Ouyang, J.~Zhuang,
  J.~Lee, Y.~Guo \emph{et~al.}, ``Improving image generation with better
  captions,'' \emph{Computer Science. https://cdn. openai. com/papers/dall-e-3.
  pdf}, vol.~2, no.~3, p.~8, 2023.

\bibitem{saharia2022photorealistic}
C.~Saharia, W.~Chan, S.~Saxena, L.~Li, J.~Whang, E.~L. Denton, K.~Ghasemipour,
  R.~Gontijo~Lopes, B.~Karagol~Ayan, T.~Salimans \emph{et~al.},
  ``Photorealistic text-to-image diffusion models with deep language
  understanding,'' \emph{Advances in neural information processing systems},
  vol.~35, pp. 36\,479--36\,494, 2022.

\bibitem{fu2025conditional}
H.~Fu, W.~Si, and R.~Liu, ``Conditional denoising diffusion-based channel
  estimation for fast time-varying mimo-ofdm systems,'' \emph{Digital Signal
  Processing}, p. 105283, 2025.

\bibitem{grassucci2024diffusion}
E.~Grassucci, C.~Marinoni, A.~Rodriguez, and D.~Comminiello, ``Diffusion models
  for audio semantic communication,'' in \emph{ICASSP 2024-2024 IEEE
  International Conference on Acoustics, Speech and Signal Processing
  (ICASSP)}.\hskip 1em plus 0.5em minus 0.4em\relax Seoul, Korea: IEEE, April
  2024, pp. 13\,136--13\,140.

\bibitem{zeng2024dmce}
Y.~Zeng, X.~He, X.~Chen, H.~Tong, Z.~Yang, Y.~Guo, and J.~Hao, ``Dmce:
  Diffusion model channel enhancer for multi-user semantic communication
  systems,'' in \emph{ICC 2024-IEEE International Conference on
  Communications}.\hskip 1em plus 0.5em minus 0.4em\relax IEEE, 2024, pp.
  855--860.

\bibitem{xu2024diffusion}
Y.~Xu, L.~Huang, L.~Zhang, L.~Qian, and X.~Yang, ``Diffusion-based radio signal
  augmentation for automatic modulation classification,'' \emph{Electronics},
  vol.~13, no.~11, p. 2063, 2024.

\bibitem{peebles2023scalable}
W.~Peebles and S.~Xie, ``Scalable diffusion models with transformers,'' in
  \emph{Proceedings of the IEEE/CVF international conference on computer
  vision}, 2023, pp. 4195--4205.

\bibitem{liu2024sora}
Y.~Liu, K.~Zhang, Y.~Li, Z.~Yan, C.~Gao, R.~Chen, Z.~Yuan, Y.~Huang, H.~Sun,
  J.~Gao \emph{et~al.}, ``Sora: A review on background, technology,
  limitations, and opportunities of large vision models,'' \emph{arXiv preprint
  arXiv:2402.17177}, 2024.

\bibitem{jacobs1991adaptive}
R.~A. Jacobs, M.~I. Jordan, S.~J. Nowlan, and G.~E. Hinton, ``Adaptive mixtures
  of local experts,'' \emph{Neural computation}, vol.~3, no.~1, pp. 79--87,
  1991.

\bibitem{du2022glam}
N.~Du, Y.~Huang, A.~M. Dai, S.~Tong, D.~Lepikhin, Y.~Xu, M.~Krikun, Y.~Zhou,
  A.~W. Yu, O.~Firat \emph{et~al.}, ``Glam: Efficient scaling of language
  models with mixture-of-experts,'' in \emph{International conference on
  machine learning}.\hskip 1em plus 0.5em minus 0.4em\relax PMLR, 2022, pp.
  5547--5569.

\bibitem{jiang2024mixtral}
A.~Q. Jiang, A.~Sablayrolles, A.~Roux, A.~Mensch, B.~Savary, C.~Bamford, D.~S.
  Chaplot, D.~d.~l. Casas, E.~B. Hanna, F.~Bressand \emph{et~al.}, ``Mixtral of
  experts,'' \emph{arXiv preprint arXiv:2401.04088}, 2024.

\bibitem{zhao2025enhancing}
C.~Zhao, H.~Du, D.~Niyato, J.~Kang, Z.~Xiong, D.~I. Kim, X.~S. Shen, and K.~B.
  Letaief, ``Enhancing physical layer communication security through generative
  ai with mixture of experts,'' \emph{IEEE Wireless Communications}, 2025.

\bibitem{zhang2024generative2}
R.~Zhang, H.~Du, Y.~Liu, D.~Niyato, J.~Kang, Z.~Xiong, A.~Jamalipour, and
  D.~In~Kim, ``Generative ai agents with large language model for satellite
  networks via a mixture of experts transmission,'' \emph{IEEE Journal on
  Selected Areas in Communications}, vol.~42, no.~12, pp. 3581--3596, 2024.

\bibitem{gao2023moe}
J.~Gao, Q.~Cao, and Y.~Chen, ``Moe-amc: Enhancing automatic modulation
  classification performance using mixture-of-experts,'' \emph{arXiv preprint
  arXiv:2312.02298}, 2023.

\bibitem{team2023gemini}
G.~Team, R.~Anil, S.~Borgeaud, J.-B. Alayrac, J.~Yu, R.~Soricut, J.~Schalkwyk,
  A.~M. Dai, A.~Hauth, K.~Millican \emph{et~al.}, ``Gemini: a family of highly
  capable multimodal models,'' \emph{arXiv preprint arXiv:2312.11805}, 2023.

\bibitem{team2024gemini}
G.~Team, P.~Georgiev, V.~I. Lei, R.~Burnell, L.~Bai, A.~Gulati, G.~Tanzer,
  D.~Vincent, Z.~Pan, S.~Wang \emph{et~al.}, ``Gemini 1.5: Unlocking multimodal
  understanding across millions of tokens of context,'' \emph{arXiv preprint
  arXiv:2403.05530}, 2024.

\bibitem{touvron2023llama}
H.~Touvron, T.~Lavril, G.~Izacard, X.~Martinet, M.-A. Lachaux, T.~Lacroix,
  B.~Rozi{\`e}re, N.~Goyal, E.~Hambro, F.~Azhar \emph{et~al.}, ``Llama: Open
  and efficient foundation language models,'' \emph{arXiv preprint
  arXiv:2302.13971}, 2023.

\bibitem{touvron2023llama2}
H.~Touvron, L.~Martin, K.~Stone, P.~Albert, A.~Almahairi, Y.~Babaei,
  N.~Bashlykov, S.~Batra, P.~Bhargava, S.~Bhosale \emph{et~al.}, ``Llama 2:
  Open foundation and fine-tuned chat models,'' \emph{arXiv preprint
  arXiv:2307.09288}, 2023.

\bibitem{dubey2024llama}
A.~Dubey, A.~Jauhri, A.~Pandey, A.~Kadian, A.~Al-Dahle, A.~Letman, A.~Mathur,
  A.~Schelten, A.~Yang, A.~Fan \emph{et~al.}, ``The llama 3 herd of models,''
  \emph{arXiv preprint arXiv:2407.21783}, 2024.

\bibitem{jiang2024large5}
F.~Jiang, L.~Dong, Y.~Peng, K.~Wang, K.~Yang, C.~Pan, and X.~You, ``Large ai
  model empowered multimodal semantic communications,'' \emph{IEEE
  Communications Magazine}, vol.~63, no.~1, pp. 76--82, 2025.

\bibitem{yao2024survey}
Y.~Yao, J.~Duan, K.~Xu, Y.~Cai, Z.~Sun, and Y.~Zhang, ``A survey on large
  language model (llm) security and privacy: The good, the bad, and the ugly,''
  \emph{High-Confidence Computing}, p. 100211, 2024.

\bibitem{he2022masked}
K.~He, X.~Chen, S.~Xie, Y.~Li, P.~Doll{\'a}r, and R.~Girshick, ``Masked
  autoencoders are scalable vision learners,'' in \emph{Proceedings of the
  IEEE/CVF conference on computer vision and pattern recognition}, 2022, pp.
  16\,000--16\,009.

\bibitem{oquab2023dinov2}
M.~Oquab, T.~Darcet, T.~Moutakanni, H.~Vo, M.~Szafraniec, V.~Khalidov,
  P.~Fernandez, D.~Haziza, F.~Massa, A.~El-Nouby \emph{et~al.}, ``Dinov2:
  Learning robust visual features without supervision,'' \emph{arXiv preprint
  arXiv:2304.07193}, 2023.

\bibitem{tariq2023segment}
S.~Tariq, B.~E. Arfeto, C.~Zhang, and H.~Shin, ``Segment anything meets
  semantic communication,'' \emph{arXiv preprint arXiv:2306.02094}, 2023.

\bibitem{jiang2025lightweight}
F.~Jiang, S.~Tu, L.~Dong, K.~Wang, K.~Yang, R.~Liu, C.~Pan, and J.~Wang,
  ``Lightweight vision model-based multi-user semantic communication systems,''
  \emph{arXiv preprint arXiv:2502.16424}, 2025.

\bibitem{liu2024visual}
H.~Liu, C.~Li, Q.~Wu, and Y.~J. Lee, ``Visual instruction tuning,''
  \emph{Advances in neural information processing systems}, vol.~36, 2024.

\bibitem{zhang2024llava}
H.~Zhang, H.~Li, F.~Li, T.~Ren, X.~Zou, S.~Liu, S.~Huang, J.~Gao, Leizhang,
  C.~Li \emph{et~al.}, ``Llava-grounding: Grounded visual chat with large
  multimodal models,'' in \emph{European Conference on Computer Vision}.\hskip
  1em plus 0.5em minus 0.4em\relax Springer, 2024, pp. 19--35.

\bibitem{guo2024llava}
Z.~Guo, R.~Xu, Y.~Yao, J.~Cui, Z.~Ni, C.~Ge, T.-S. Chua, Z.~Liu, and G.~Huang,
  ``Llava-uhd: an lmm perceiving any aspect ratio and high-resolution images,''
  in \emph{European Conference on Computer Vision}.\hskip 1em plus 0.5em minus
  0.4em\relax Springer, 2024, pp. 390--406.

\bibitem{qiao2024latency}
L.~Qiao, M.~B. Mashhadi, Z.~Gao, C.~H. Foh, P.~Xiao, and M.~Bennis,
  ``Latency-aware generative semantic communications with pre-trained diffusion
  models,'' \emph{arXiv preprint arXiv:2403.17256}, 2024.

\bibitem{xu2024large1}
S.~Xu, C.~K. Thomas, O.~Hashash, N.~Muralidhar, W.~Saad, and N.~Ramakrishnan,
  ``Large multi-modal models (lmms) as universal foundation models for
  ai-native wireless systems,'' \emph{arXiv preprint arXiv:2402.01748}, 2024.

\bibitem{kawamoto2023application}
T.~Kawamoto, T.~Suzuki, K.~Miyama, T.~Meguro, and T.~Takagi, ``Application of
  frozen large-scale models to multimodal task-oriented dialogue,'' \emph{arXiv
  preprint arXiv:2310.00845}, 2023.

\bibitem{ghasemi2025comprehensivesurveyreinforcementlearning}
\BIBentryALTinterwordspacing
M.~Ghasemi, A.~H. Moosavi, and D.~Ebrahimi, ``A comprehensive survey of
  reinforcement learning: From algorithms to practical challenges,'' 2025.
  [Online]. Available: \url{https://arxiv.org/abs/2411.18892}
\BIBentrySTDinterwordspacing

\bibitem{zhang2023instruction}
S.~Zhang, L.~Dong, X.~Li, S.~Zhang, X.~Sun, S.~Wang, J.~Li, R.~Hu, T.~Zhang,
  F.~Wu \emph{et~al.}, ``Instruction tuning for large language models: A
  survey,'' \emph{arXiv preprint arXiv:2308.10792}, 2023.

\bibitem{wei2022chain}
J.~Wei, X.~Wang, D.~Schuurmans, M.~Bosma, F.~Xia, E.~Chi, Q.~V. Le, D.~Zhou
  \emph{et~al.}, ``Chain-of-thought prompting elicits reasoning in large
  language models,'' \emph{Advances in neural information processing systems},
  vol.~35, pp. 24\,824--24\,837, 2022.

\bibitem{lewis2020retrieval}
P.~Lewis, E.~Perez, A.~Piktus, F.~Petroni, V.~Karpukhin, N.~Goyal,
  H.~K{\"u}ttler, M.~Lewis, W.-t. Yih, T.~Rockt{\"a}schel \emph{et~al.},
  ``Retrieval-augmented generation for knowledge-intensive nlp tasks,''
  \emph{Advances in Neural Information Processing Systems}, vol.~33, pp.
  9459--9474, 2020.

\bibitem{shao2024deepseekmath}
Z.~Shao, P.~Wang, Q.~Zhu, R.~Xu, J.~Song, X.~Bi, H.~Zhang, M.~Zhang, Y.~Li,
  Y.~Wu \emph{et~al.}, ``Deepseekmath: Pushing the limits of mathematical
  reasoning in open language models,'' \emph{arXiv preprint arXiv:2402.03300},
  2024.

\bibitem{yang2024qwen2}
A.~Yang, B.~Yang, B.~Zhang, B.~Hui, B.~Zheng, B.~Yu, C.~Li, D.~Liu, F.~Huang,
  H.~Wei \emph{et~al.}, ``Qwen2. 5 technical report,'' \emph{arXiv preprint
  arXiv:2412.15115}, 2024.

\bibitem{coulom2006efficient}
R.~Coulom, ``Efficient selectivity and backup operators in monte-carlo tree
  search,'' in \emph{International conference on computers and games}.\hskip
  1em plus 0.5em minus 0.4em\relax Springer, 2006, pp. 72--83.

\bibitem{qu2025survey}
X.~Qu, Y.~Li, Z.~Su, W.~Sun, J.~Yan, D.~Liu, G.~Cui, D.~Liu, S.~Liang, J.~He
  \emph{et~al.}, ``A survey of efficient reasoning for large reasoning models:
  Language, multimodality, and beyond,'' \emph{arXiv preprint
  arXiv:2503.21614}, 2025.

\bibitem{ainslie2023gqa}
J.~Ainslie, J.~Lee-Thorp, M.~De~Jong, Y.~Zemlyanskiy, F.~Lebr{\'o}n, and
  S.~Sanghai, ``Gqa: Training generalized multi-query transformer models from
  multi-head checkpoints,'' \emph{arXiv preprint arXiv:2305.13245}, 2023.

\bibitem{shazeer2019fast}
N.~Shazeer, ``Fast transformer decoding: One write-head is all you need,''
  \emph{arXiv preprint arXiv:1911.02150}, 2019.

\bibitem{zhang2024tinyllama}
P.~Zhang, G.~Zeng, T.~Wang, and W.~Lu, ``Tinyllama: An open-source small
  language model,'' \emph{arXiv preprint arXiv:2401.02385}, 2024.

\bibitem{xu2025evaluating}
B.~Xu, Y.~Chen, Z.~Wen, W.~Liu, and B.~He, ``Evaluating small language models
  for news summarization: Implications and factors influencing performance,''
  \emph{arXiv preprint arXiv:2502.00641}, 2025.

\bibitem{hu2024minicpm}
S.~Hu, Y.~Tu, X.~Han, C.~He, G.~Cui, X.~Long, Z.~Zheng, Y.~Fang, Y.~Huang,
  W.~Zhao \emph{et~al.}, ``Minicpm: Unveiling the potential of small language
  models with scalable training strategies,'' \emph{arXiv preprint
  arXiv:2404.06395}, 2024.

\bibitem{javaheripi2023phi}
M.~Javaheripi, S.~Bubeck, M.~Abdin, J.~Aneja, S.~Bubeck, C.~C.~T. Mendes,
  W.~Chen, A.~Del~Giorno, R.~Eldan, S.~Gopi \emph{et~al.}, ``Phi-2: The
  surprising power of small language models,'' \emph{Microsoft Research Blog},
  vol.~1, no.~3, p.~3, 2023.

\bibitem{abdin2024phi}
M.~Abdin, J.~Aneja, H.~Awadalla, A.~Awadallah, A.~A. Awan, N.~Bach, A.~Bahree,
  A.~Bakhtiari, J.~Bao, H.~Behl \emph{et~al.}, ``Phi-3 technical report: A
  highly capable language model locally on your phone,'' \emph{arXiv preprint
  arXiv:2404.14219}, 2024.

\bibitem{zou2024telecomgpt}
H.~Zou, Q.~Zhao, Y.~Tian, L.~Bariah, F.~Bader, T.~Lestable, and M.~Debbah,
  ``Telecomgpt: A framework to build telecom-specfic large language models,''
  \emph{arXiv preprint arXiv:2407.09424}, 2024.

\bibitem{nikbakht2024tspec}
R.~Nikbakht, M.~Benzaghta, and G.~Geraci, ``Tspec-llm: An open-source dataset
  for llm understanding of 3gpp specifications,'' \emph{arXiv preprint
  arXiv:2406.01768}, 2024.

\bibitem{jiang2025commgpt}
F.~Jiang, W.~Zhu, L.~Dong, K.~Wang, K.~Yang, C.~Pan, and O.~A. Dobre,
  ``Commgpt: A graph and retrieval-augmented multimodal communication
  foundation model,'' \emph{arXiv preprint arXiv:2502.18763}, 2025.

\bibitem{wang2021milvus}
J.~Wang, X.~Yi, R.~Guo, H.~Jin, P.~Xu, S.~Li, X.~Wang, X.~Guo, C.~Li, X.~Xu
  \emph{et~al.}, ``Milvus: A purpose-built vector data management system,'' in
  \emph{Proceedings of the 2021 International Conference on Management of
  Data}, 2021, pp. 2614--2627.

\bibitem{henriquezgraph}
S.~D.~M. Henriquez, E.~R. Tito, J.~F.~I. Loo, L.~Ruth, and H.~Rondón, ``Graph
  database with neo4j and the cypher language: An application in mining
  companies.''

\bibitem{jaech2024openai}
A.~Jaech, A.~Kalai, A.~Lerer, A.~Richardson, A.~El-Kishky, A.~Low, A.~Helyar,
  A.~Madry, A.~Beutel, A.~Carney \emph{et~al.}, ``Openai o1 system card,''
  \emph{arXiv preprint arXiv:2412.16720}, 2024.

\bibitem{yao2024tree}
S.~Yao, D.~Yu, J.~Zhao, I.~Shafran, T.~Griffiths, Y.~Cao, and K.~Narasimhan,
  ``Tree of thoughts: Deliberate problem solving with large language models,''
  \emph{Advances in Neural Information Processing Systems}, vol.~36, 2024.

\bibitem{besta2024graph}
M.~Besta, N.~Blach, A.~Kubicek, R.~Gerstenberger, M.~Podstawski, L.~Gianinazzi,
  J.~Gajda, T.~Lehmann, H.~Niewiadomski, P.~Nyczyk \emph{et~al.}, ``Graph of
  thoughts: Solving elaborate problems with large language models,'' in
  \emph{Proceedings of the AAAI Conference on Artificial Intelligence},
  vol.~38, no.~16, 2024, pp. 17\,682--17\,690.

\bibitem{wang2023plan}
L.~Wang, W.~Xu, Y.~Lan, Z.~Hu, Y.~Lan, R.~K.-W. Lee, and E.-P. Lim,
  ``Plan-and-solve prompting: Improving zero-shot chain-of-thought reasoning by
  large language models,'' \emph{arXiv preprint arXiv:2305.04091}, 2023.

\bibitem{gao2023retrieval}
Y.~Gao, Y.~Xiong, X.~Gao, K.~Jia, J.~Pan, Y.~Bi, Y.~Dai, J.~Sun, H.~Wang, and
  H.~Wang, ``Retrieval-augmented generation for large language models: A
  survey,'' \emph{arXiv preprint arXiv:2312.10997}, vol.~2, p.~1, 2023.

\bibitem{pan2024unifying}
S.~Pan, L.~Luo, Y.~Wang, C.~Chen, J.~Wang, and X.~Wu, ``Unifying large language
  models and knowledge graphs: A roadmap,'' \emph{IEEE Transactions on
  Knowledge and Data Engineering}, vol.~36, no.~7, pp. 3580--3599, 2024.

\bibitem{sahoo2024systematic}
P.~Sahoo, A.~K. Singh, S.~Saha, V.~Jain, S.~Mondal, and A.~Chadha, ``A
  systematic survey of prompt engineering in large language models: Techniques
  and applications,'' \emph{arXiv preprint arXiv:2402.07927}, 2024.

\bibitem{ehtesham2025survey}
A.~Ehtesham, A.~Singh, G.~K. Gupta, and S.~Kumar, ``A survey of agent
  interoperability protocols: Model context protocol (mcp), agent communication
  protocol (acp), agent-to-agent protocol (a2a), and agent network protocol
  (anp),'' \emph{arXiv preprint arXiv:2505.02279}, 2025.

\bibitem{qu2025tool}
C.~Qu, S.~Dai, X.~Wei, H.~Cai, S.~Wang, D.~Yin, J.~Xu, and J.-R. Wen, ``Tool
  learning with large language models: A survey,'' \emph{Frontiers of Computer
  Science}, vol.~19, no.~8, p. 198343, 2025.

\bibitem{lindemann2021survey}
B.~Lindemann, T.~M{\"u}ller, H.~Vietz, N.~Jazdi, and M.~Weyrich, ``A survey on
  long short-term memory networks for time series prediction,'' \emph{Procedia
  Cirp}, vol.~99, pp. 650--655, 2021.

\bibitem{han2023comprehensive}
Y.~Han, C.~Liu, and P.~Wang, ``A comprehensive survey on vector database:
  Storage and retrieval technique, challenge,'' \emph{arXiv preprint
  arXiv:2310.11703}, 2023.

\bibitem{ueki2021survey}
K.~Ueki, ``Survey of visual-semantic embedding methods for zero-shot image
  retrieval,'' in \emph{2021 20th IEEE International Conference on Machine
  Learning and Applications (ICMLA)}.\hskip 1em plus 0.5em minus 0.4em\relax
  IEEE, 2021, pp. 628--634.

\bibitem{shinn2023reflexion}
N.~Shinn, F.~Cassano, A.~Gopinath, K.~Narasimhan, and S.~Yao, ``Reflexion:
  Language agents with verbal reinforcement learning,'' \emph{Advances in
  Neural Information Processing Systems}, vol.~36, pp. 8634--8652, 2023.

\bibitem{sapkota2025ai}
R.~Sapkota, K.~I. Roumeliotis, and M.~Karkee, ``Ai agents vs. agentic ai: A
  conceptual taxonomy, applications and challenge,'' \emph{arXiv preprint
  arXiv:2505.10468}, 2025.

\bibitem{wang2024large}
Z.~Wang, L.~Zou, S.~Wei, F.~Liao, J.~Zhuo, H.~Mi, and R.~Lai, ``Large language
  model enabled semantic communication systems,'' \emph{arXiv preprint
  arXiv:2407.14112}, 2024.

\bibitem{jiang2024semantic}
P.~Jiang, C.-K. Wen, X.~Yi, X.~Li, S.~Jin, and J.~Zhang, ``Semantic
  communications using foundation models: Design approaches and open issues,''
  \emph{IEEE Wireless Communications}, vol.~31, no.~3, pp. 76--84, 2024.

\bibitem{salehi2025llm}
S.~Salehi, M.~Erol-Kantarci, and D.~Niyato, ``Llm-enabled data transmission in
  end-to-end semantic communication,'' \emph{arXiv preprint arXiv:2504.07431},
  2025.

\bibitem{jiang2024large6}
F.~Jiang, S.~Tu, L.~Dong, C.~Pan, J.~Wang, and X.~You, ``Large generative
  model-assisted talking-face semantic communication system,'' \emph{arXiv
  preprint arXiv:2411.03876}, 2024.

\bibitem{ribouh2025large}
S.~Ribouh and O.~Saleem, ``Large language model-based semantic communication
  system for image transmission,'' \emph{arXiv preprint arXiv:2501.12988},
  2025.

\bibitem{jiang2024visual}
F.~Jiang, C.~Tang, L.~Dong, K.~Wang, K.~Yang, and C.~Pan, ``Visual language
  model based cross-modal semantic communication systems,'' \emph{IEEE
  Transactions on Wireless Communications}, pp. 1--1, Mar. 2025.

\bibitem{jiang2024large1}
F.~Jiang, Y.~Peng, L.~Dong, K.~Wang, K.~Yang, C.~Pan, and X.~You, ``Large
  generative model assisted 3d semantic communication,'' \emph{arXiv preprint
  arXiv:2403.05783}, 2024.

\bibitem{cao2024multimodal}
D.~Cao, J.~Wu, and A.~K. Bashir, ``Multimodal large language models driven
  privacy-preserving wireless semantic communication in 6g,'' in \emph{2024
  IEEE International Conference on Communications Workshops (ICC
  Workshops)}.\hskip 1em plus 0.5em minus 0.4em\relax IEEE, 2024, pp. 171--176.

\bibitem{jiang2025m4sc}
F.~Jiang, S.~Tu, L.~Dong, K.~Wang, K.~Yang, and C.~Pan, ``M4sc: An mllm-based
  multi-modal, multi-task and multi-user semantic communication system,''
  \emph{arXiv preprint arXiv:2502.16418}, 2025.

\bibitem{chen2024semantic}
W.~Chen, W.~Xu, H.~Chen, X.~Zhang, Z.~Qin, Y.~Zhang, and Z.~Han, ``Semantic
  communication based on large language model for underwater image
  transmission,'' \emph{arXiv preprint arXiv:2408.12616}, 2024.

\bibitem{kalita2024large}
A.~Kalita, ``Large language models (llms) for semantic communication in
  edge-based iot networks,'' \emph{arXiv preprint arXiv:2407.20970}, 2024.

\bibitem{cui2024llmind}
H.~Cui, Y.~Du, Q.~Yang, Y.~Shao, and S.~C. Liew, ``Llmind: Orchestrating ai and
  iot with llm for complex task execution,'' \emph{IEEE Communications
  Magazine}, vol.~63, no.~4, pp. 214--220, 2025.

\bibitem{shen2025autoiot}
L.~Shen, Q.~Yang, Y.~Zheng, and M.~Li, ``Autoiot: Llm-driven automated natural
  language programming for aiot applications,'' \emph{arXiv preprint
  arXiv:2503.05346}, 2025.

\bibitem{chen2025llm}
X.~Chen, W.~Wu, Z.~Li, L.~Li, and F.~Ji, ``Llm-empowered iot for 6g networks:
  Architecture, challenges, and solutions,'' \emph{arXiv preprint
  arXiv:2503.13819}, 2025.

\bibitem{xiao2024efficient}
B.~Xiao, B.~Kantarci, J.~Kang, D.~Niyato, and M.~Guizani, ``Efficient prompting
  for llm-based generative internet of things,'' \emph{IEEE Internet of Things
  Journal}, 2024.

\bibitem{otoum2025llm}
Y.~Otoum, A.~Asad, and A.~Nayak, ``Llm-based threat detection and prevention
  framework for iot ecosystems,'' \emph{arXiv preprint arXiv:2505.00240}, 2025.

\bibitem{zong2025integrating}
M.~Zong, A.~Hekmati, M.~Guastalla, Y.~Li, and B.~Krishnamachari, ``Integrating
  large language models with internet of things: applications,'' \emph{Discover
  Internet of Things}, vol.~5, no.~1, p.~2, 2025.

\bibitem{shirali2024llm}
M.~Shirali, M.~F. Sani, Z.~Ahmadi, and E.~Serral, ``Llm-based event abstraction
  and integration for iot-sourced logs,'' in \emph{International Conference on
  Business Process Management}.\hskip 1em plus 0.5em minus 0.4em\relax
  Springer, 2024, pp. 138--149.

\bibitem{an2024iot}
T.~An, Y.~Zhou, H.~Zou, and J.~Yang, ``Iot-llm: Enhancing real-world iot task
  reasoning with large language models,'' \emph{arXiv preprint
  arXiv:2410.02429}, 2024.

\bibitem{mo2024iot}
S.~Mo, R.~Salakhutdinov, L.-P. Morency, and P.~P. Liang, ``Iot-lm: Large
  multisensory language models for the internet of things,'' \emph{arXiv
  preprint arXiv:2407.09801}, 2024.

\bibitem{xu2023fwdllm}
M.~Xu, D.~Cai, Y.~Wu, X.~Li, and S.~Wang, ``Fwdllm: Efficient fedllm using
  forward gradient,'' \emph{arXiv preprint arXiv:2308.13894}, 2023.

\bibitem{yu2024edge}
Z.~Yu, Z.~Wang, Y.~Li, R.~Gao, X.~Zhou, S.~R. Bommu, Y.~Zhao, and Y.~Lin,
  ``Edge-llm: Enabling efficient large language model adaptation on edge
  devices via unified compression and adaptive layer voting,'' in
  \emph{Proceedings of the 61st ACM/IEEE Design Automation Conference}, San
  Francisco, Jun. 2024, pp. 1--6.

\bibitem{zhang2024edgeshard}
M.~Zhang, X.~Shen, J.~Cao, Z.~Cui, and S.~Jiang, ``Edgeshard: Efficient llm
  inference via collaborative edge computing,'' \emph{IEEE Internet of Things
  Journal}, pp. 1--1, Dec. 2024.

\bibitem{zhao2024edge}
W.~Zhao, W.~Jing, Z.~Lu, and X.~Wen, ``Edge and terminal cooperation enabled
  llm deployment optimization in wireless network,'' in \emph{IEEE/CIC
  International Conference on Communications (ICCC Workshops)}, Hangzhou, Aug.
  2024, pp. 220--225.

\bibitem{qu2025mobile}
G.~Qu, Q.~Chen, W.~Wei, Z.~Lin, X.~Chen, and K.~Huang, ``Mobile edge
  intelligence for large language models: A contemporary survey,'' \emph{IEEE
  Communications Surveys \& Tutorials}, pp. 1--1, Mar. 2025.

\bibitem{khoshsirat2024decentralized}
A.~Khoshsirat, G.~Perin, and M.~Rossi, ``Decentralized llm inference over edge
  networks with energy harvesting,'' \emph{arXiv preprint arXiv:2408.15907},
  2024.

\bibitem{lin2023pushing}
Z.~Lin, G.~Qu, Q.~Chen, X.~Chen, Z.~Chen, and K.~Huang, ``Pushing large
  language models to the 6g edge: Vision, challenges, and opportunities,''
  \emph{arXiv preprint arXiv:2309.16739}, 2023.

\bibitem{friha2024llm}
O.~Friha, M.~Amine~Ferrag, B.~Kantarci, B.~Cakmak, A.~Ozgun, and
  N.~Ghoualmi-Zine, ``Llm-based edge intelligence: A comprehensive survey on
  architectures, applications, security and trustworthiness,'' \emph{IEEE Open
  Journal of the Communications Society}, vol.~5, pp. 5799--5856, 2024.

\bibitem{wang2023network}
J.~Wang, L.~Zhang, Y.~Yang, Z.~Zhuang, Q.~Qi, H.~Sun, L.~Lu, J.~Feng, and
  J.~Liao, ``Network meets chatgpt: Intent autonomous management, control and
  operation,'' \emph{Journal of Communications and Information Networks},
  vol.~8, no.~3, pp. 239--255, 2023.

\bibitem{yue2023ai}
L.~Yue and T.~Chen, ``Ai large model and 6g network,'' in \emph{2023 IEEE
  Globecom Workshops (GC Wkshps)}.\hskip 1em plus 0.5em minus 0.4em\relax Kuala
  Lumpur, Malaysia: IEEE, December 2023, pp. 2049--2054.

\bibitem{komanduri2025optimizing}
V.~Komanduri, S.~Estropia, S.~Alessio, G.~Yerdelen, T.~Ferreira, G.~P. Roldan,
  Z.~Dong, and R.~Rojas-Cessa, ``Optimizing llm prompts for automation of
  network management: A user's perspective,'' in \emph{2025 International
  Conference on Artificial Intelligence in Information and Communication
  (ICAIIC)}.\hskip 1em plus 0.5em minus 0.4em\relax IEEE, 2025, pp. 0958--0963.

\bibitem{mani2023enhancing}
S.~K. Mani, Y.~Zhou, K.~Hsieh, S.~Segarra, T.~Eberl, E.~Azulai, I.~Frizler,
  R.~Chandra, and S.~Kandula, ``Enhancing network management using code
  generated by large language models,'' in \emph{Proceedings of the 22nd ACM
  Workshop on Hot Topics in Networks}, 2023, pp. 196--204.

\bibitem{lee2024large}
H.~Lee, M.~Kim, S.~Baek, N.~Lee, M.~Debbah, and I.~Lee, ``Large language models
  for knowledge-free network management: Feasibility study and opportunities,''
  \emph{arXiv preprint arXiv:2410.17259}, 2024.

\bibitem{wu2024netllm}
D.~Wu, X.~Wang, Y.~Qiao, Z.~Wang, J.~Jiang, S.~Cui, and F.~Wang, ``Netllm:
  Adapting large language models for networking,'' in \emph{Proceedings of the
  ACM SIGCOMM 2024 Conference}, 2024, pp. 661--678.

\bibitem{dandoush2024large}
A.~Dandoush, V.~Kumarskandpriya, M.~Uddin, and U.~Khalil, ``Large language
  models meet network slicing management and orchestration,'' \emph{arXiv
  preprint arXiv:2403.13721}, 2024.

\bibitem{he2024designing}
Z.~He, A.~Gottipati, L.~Qiu, X.~Luo, K.~Xu, Y.~Yang, and F.~Y. Yan, ``Designing
  network algorithms via large language models,'' in \emph{Proceedings of the
  23rd ACM Workshop on Hot Topics in Networks}, 2024, pp. 205--212.

\bibitem{yang2024comprehensive}
H.~Yang, K.~Xiang, M.~Ge, H.~Li, R.~Lu, and S.~Yu, ``A comprehensive overview
  of backdoor attacks in large language models within communication networks,''
  \emph{IEEE Network}, 2024.

\bibitem{khowaja2024pathway}
S.~A. Khowaja, P.~Khuwaja, K.~Dev, H.~A. Hamadi, and E.~Zeydan, ``Pathway to
  secure and trustworthy 6g for llms: Attacks, defense, and opportunities,''
  \emph{arXiv preprint arXiv:2408.00722}, 2024.

\bibitem{luo2023bc4llm}
H.~Luo, J.~Luo, and A.~V. Vasilakos, ``Bc4llm: Trusted artificial intelligence
  when blockchain meets large language models,'' \emph{arXiv preprint
  arXiv:2310.06278}, 2023.

\bibitem{mishra2024sentinellms}
A.~Mishra, M.~Li, and S.~Deo, ``Sentinellms: Encrypted input adaptation and
  fine-tuning of language models for private and secure inference,'' in
  \emph{Proceedings of the AAAI Conference on Artificial Intelligence},
  vol.~38, no.~19, 2024, pp. 21\,403--21\,411.

\bibitem{zeng2024privacyrestore}
Z.~Zeng, J.~Wang, J.~Yang, Z.~Lu, H.~Zhuang, and C.~Chen, ``Privacyrestore:
  Privacy-preserving inference in large language models via privacy removal and
  restoration,'' \emph{arXiv preprint arXiv:2406.01394}, 2024.

\bibitem{wang2023privatelora}
Y.~Wang, Y.~Lin, X.~Zeng, and G.~Zhang, ``Privatelora for efficient privacy
  preserving llm,'' \emph{arXiv preprint arXiv:2311.14030}, 2023.

\bibitem{feretzakis2024trustworthy}
G.~Feretzakis and V.~S. Verykios, ``Trustworthy ai: Securing sensitive data in
  large language models,'' \emph{AI}, vol.~5, no.~4, pp. 2773--2800, 2024.

\bibitem{zhang2025llms}
J.~Zhang, H.~Bu, H.~Wen, Y.~Liu, H.~Fei, R.~Xi, L.~Li, Y.~Yang, H.~Zhu, and
  D.~Meng, ``When llms meet cybersecurity: A systematic literature review,''
  \emph{Cybersecurity}, vol.~8, no.~1, pp. 1--41, 2025.

\bibitem{zhang2025large}
R.~Zhang, H.-W. Li, X.-Y. Qian, W.-B. Jiang, and H.-X. Chen, ``On large
  language models safety, security, and privacy: A survey,'' \emph{Journal of
  Electronic Science and Technology}, p. 100301, 2025.

\bibitem{lee2024llm}
W.~Lee and J.~Park, ``Llm-empowered resource allocation in wireless
  communications systems,'' \emph{arXiv preprint arXiv:2408.02944}, 2024.

\bibitem{10592370}
H.~Du, G.~Liu, Y.~Lin, D.~Niyato, J.~Kang, Z.~Xiong, and D.~I. Kim, ``Mixture
  of experts for intelligent networks: A large language model-enabled
  approach,'' in \emph{2024 International Wireless Communications and Mobile
  Computing (IWCMC)}, Ayia Napa, Cyprus, May 2024, pp. 531--536.

\bibitem{peng2025llm}
X.~Peng, Y.~Liu, Y.~Cang, C.~Cao, and M.~Chen, ``Llm-optira: Llm-driven
  optimization of resource allocation for non-convex problems in wireless
  communications,'' \emph{arXiv preprint arXiv:2505.02091}, 2025.

\bibitem{ren2024retrieval}
R.~Ren, Y.~Wu, X.~Zhang, J.~Ren, Y.~Shen, S.~Wang, and K.-F. Tsang,
  ``Retrieval-augmented generation for mobile edge computing via large language
  model,'' \emph{arXiv preprint arXiv:2412.20820}, 2024.

\bibitem{xu2024cachedmodelasaresourceprovisioninglarge}
\BIBentryALTinterwordspacing
M.~Xu, D.~Niyato, H.~Zhang, J.~Kang, Z.~Xiong, S.~Mao, and Z.~Han, ``Cached
  model-as-a-resource: Provisioning large language model agents for edge
  intelligence in space-air-ground integrated networks,'' 2024. [Online].
  Available: \url{https://arxiv.org/abs/2403.05826}
\BIBentrySTDinterwordspacing

\bibitem{yang2024perllm}
Z.~Yang, Y.~Yang, C.~Zhao, Q.~Guo, W.~He, and W.~Ji, ``Perllm: Personalized
  inference scheduling with edge-cloud collaboration for diverse llm
  services,'' \emph{arXiv preprint arXiv:2405.14636}, 2024.

\bibitem{liu2024resource}
C.~Liu and J.~Zhao, ``Resource allocation for stable llm training in mobile
  edge computing,'' in \emph{Proceedings of the Twenty-fifth International
  Symposium on Theory, Algorithmic Foundations, and Protocol Design for Mobile
  Networks and Mobile Computing}, Athens Greece, October 2024, pp. 81--90.

\bibitem{jiang2024dllm}
\BIBentryALTinterwordspacing
Y.~Jiang, H.~Wang, L.~Xie, H.~Zhao, H.~Qian, and J.~C.~S. Lui, ``D-llm: A token
  adaptive computing resource allocation strategy for large language models,''
  in \emph{Advances in Neural Information Processing Systems (NeurIPS)},
  vol.~37, 2024, pp. 1725--1749. [Online]. Available:
  \url{https://openreview.net/forum?id=UIOjGTKHQG}
\BIBentrySTDinterwordspacing

\bibitem{amayuelas2025self}
A.~Amayuelas, J.~Yang, S.~Agashe, A.~Nagarajan, A.~Antoniades, X.~E. Wang, and
  W.~Wang, ``Self-resource allocation in multi-agent llm systems,'' \emph{arXiv
  preprint arXiv:2504.02051}, 2025.

\bibitem{zou2023wireless}
H.~Zou, Q.~Zhao, L.~Bariah, M.~Bennis, and M.~Debbah, ``Wireless multi-agent
  generative ai: From connected intelligence to collective intelligence,''
  \emph{arXiv preprint arXiv:2307.02757}, 2023.

\bibitem{long20246g}
S.~Long, F.~Tang, Y.~Li, T.~Tan, Z.~Jin, M.~Zhao, and N.~Kato, ``6g
  comprehensive intelligence: network operations and optimization based on
  large language models,'' \emph{IEEE Network}, 2024.

\bibitem{liu2025model}
Z.~Liu and H.~Du, ``Model context protocol-based internet of experts for
  wireless environment-aware llm agents,'' \emph{arXiv preprint
  arXiv:2505.01834}, 2025.

\bibitem{chen2024enabling}
Z.~Chen, Q.~Sun, N.~Li, X.~Li, Y.~Wang, and C.-L. I, ``Enabling mobile ai
  agent in 6g era: Architecture and key technologies,'' \emph{IEEE Network},
  vol.~38, no.~5, pp. 66--75, 2024.

\bibitem{xiao2024llm}
Z.~Xiao, C.~Ye, Y.~Hu, H.~Yuan, Y.~Huang, Y.~Feng, L.~Cai, and J.~Chang, ``Llm
  agents as 6g orchestrator: A paradigm for task-oriented physical-layer
  automation,'' \emph{arXiv preprint arXiv:2410.03688}, 2024.

\bibitem{xu2024large}
M.~Xu, D.~Niyato, J.~Kang, Z.~Xiong, S.~Mao, Z.~Han, D.~I. Kim, and K.~B.
  Letaief, ``When large language model agents meet 6g networks: Perception,
  grounding, and alignment,'' \emph{IEEE Wireless Communications}, vol.~31,
  no.~6, pp. 63--71, 2024.

\bibitem{10815060}
W.~Yang, Z.~Xiong, Y.~Yuan, W.~Jiang, T.~Q.~S. Quek, and M.~Debbah,
  ``Agent-driven generative semantic communication with cross-modality and
  prediction,'' \emph{IEEE Transactions on Wireless Communications}, vol.~24,
  no.~3, pp. 2233--2248, 2025.

\bibitem{10972177}
W.~Yang, Z.~Xiong, S.~Mao, T.~Q.~S. Quek, P.~Zhang, M.~Debbah, and
  R.~Tafazolli, ``Rethinking generative semantic communication for multi-user
  systems with large language models,'' \emph{IEEE Wireless Communications},
  pp. 1--9, 2025.

\bibitem{10845514}
X.~Jia, X.~Wang, Y.~Zhang, M.~Sheng, and G.~Cheng, ``Resource allocation for
  multi-cell semantic communication systems based on drl,'' in \emph{2024 12th
  International Conference on Information Systems and Computing Technology
  (ISCTech)}, 2024, pp. 1--6.

\bibitem{zhang2023toward}
H.~Zhang, H.~Wang, Y.~Li, K.~Long, and V.~C. Leung, ``Toward intelligent
  resource allocation on task-oriented semantic communication,'' \emph{IEEE
  Wireless Communications}, vol.~30, no.~3, pp. 70--77, 2023.

\bibitem{jiang2024links}
S.~Jiang, B.~Lin, Y.~Wu, and Y.~Gao, ``Links: Large language model integrated
  management for 6g empowered digital twin networks,'' in \emph{2024 IEEE 100th
  Vehicular Technology Conference (VTC2024-Fall)}.\hskip 1em plus 0.5em minus
  0.4em\relax IEEE, 2024, pp. 1--6.

\bibitem{tong2025wirelessagent}
J.~Tong, W.~Guo, J.~Shao, Q.~Wu, Z.~Li, Z.~Lin, and J.~Zhang, ``Wirelessagent:
  Large language model agents for intelligent wireless networks,'' \emph{arXiv
  preprint arXiv:2505.01074}, 2025.

\bibitem{xiao2025towards}
Y.~Xiao, G.~Shi, and P.~Zhang, ``Towards agentic ai networking in 6g: A
  generative foundation model-as-agent approach,'' \emph{arXiv preprint
  arXiv:2503.15764}, 2025.

\bibitem{wu2025llm}
X.~Wu, Y.~Wang, J.~Farooq, and J.~Chen, ``Llm-driven agentic ai approach to
  enhanced o-ran resilience in next-generation networks,'' \emph{Authorea
  Preprints}, 2025.

\bibitem{nguyen2024large}
T.~Nguyen, H.~Nguyen, A.~Ijaz, S.~Sheikhi, A.~V. Vasilakos, and P.~Kostakos,
  ``Large language models in 6g security: challenges and opportunities,''
  \emph{arXiv preprint arXiv:2403.12239}, 2024.

\bibitem{onsu2024leveraging}
M.~A. Onsu, P.~Lohan, and B.~Kantarci, ``Leveraging edge intelligence and llms
  to advance 6g-enabled internet of automated defense vehicles,'' \emph{arXiv
  preprint arXiv:2501.06205}, 2024.

\bibitem{cao2025exploring}
X.~Cao, G.~Nan, H.~Guo, H.~Mu, L.~Wang, Y.~Lin, Q.~Zhou, J.~Li, B.~Qin, Q.~Cui
  \emph{et~al.}, ``Exploring llm-based multi-agent situation awareness for
  zero-trust space-air-ground integrated network,'' \emph{IEEE Journal on
  Selected Areas in Communications}, 2025.

\bibitem{lin2024airvista}
F.~Lin, Y.~Tian, Y.~Wang, T.~Zhang, X.~Zhang, and F.-Y. Wang, ``Airvista:
  Empowering uavs with 3d spatial reasoning abilities through a multimodal
  large language model agent,'' in \emph{2024 IEEE 27th International
  Conference on Intelligent Transportation Systems (ITSC)}.\hskip 1em plus
  0.5em minus 0.4em\relax IEEE, 2024, pp. 476--481.

\bibitem{sezgin2025scenario}
A.~Sezgin, ``Scenario-driven evaluation of autonomous agents: Integrating large
  language model for uav mission reliability,'' \emph{Drones}, vol.~9, no.~3,
  p. 213, 2025.

\bibitem{zhu2024task}
F.~Zhu, F.~Huang, Y.~Yu, G.~Liu, and T.~Huang, ``Task offloading with
  llm-enhanced multi-agent reinforcement learning in uav-assisted edge
  computing,'' \emph{Sensors}, vol.~25, no.~1, p. 175, 2024.

\bibitem{sautenkov2025uav}
O.~Sautenkov, Y.~Yaqoot, M.~A. Mustafa, F.~Batool, J.~Sam, A.~Lykov, C.-Y. Wen,
  and D.~Tsetserukou, ``Uav-codeagents: Scalable uav mission planning via
  multi-agent react and vision-language reasoning,'' \emph{arXiv preprint
  arXiv:2505.07236}, 2025.

\bibitem{liu2023ai}
B.~Liu, S.~Mazumder, E.~Robertson, and S.~Grigsby, ``Ai autonomy:
  Self-initiated open-world continual learning and adaptation,'' \emph{AI
  Magazine}, vol.~44, no.~2, pp. 185--199, 2023.

\bibitem{yu2024recent}
D.~Yu, X.~Zhang, Y.~Chen, A.~Liu, Y.~Zhang, P.~S. Yu, and I.~King, ``Recent
  advances of multimodal continual learning: A comprehensive survey,''
  \emph{arXiv preprint arXiv:2410.05352}, 2024.

\bibitem{ru2024maintaining}
X.~Ru, X.~Cao, Z.~Liu, J.~M. Moore, X.-Y. Zhang, X.~Zhu, W.~Wei, and G.~Yan,
  ``Maintaining adversarial robustness in continuous learning,'' \emph{arXiv
  preprint arXiv:2402.11196}, 2024.

\bibitem{chaudhry2018riemannian}
A.~Chaudhry, P.~K. Dokania, T.~Ajanthan, and P.~H. Torr, ``Riemannian walk for
  incremental learning: Understanding forgetting and intransigence,'' in
  \emph{Proceedings of the European conference on computer vision (ECCV)},
  2018, pp. 532--547.

\bibitem{friedman2023large}
R.~Friedman, ``Large language models and logical reasoning,''
  \emph{Encyclopedia}, vol.~3, no.~2, pp. 687--697, 2023.

\bibitem{lightman2023let}
H.~Lightman, V.~Kosaraju, Y.~Burda, H.~Edwards, B.~Baker, T.~Lee, J.~Leike,
  J.~Schulman, I.~Sutskever, and K.~Cobbe, ``Let's verify step by step,'' in
  \emph{The Twelfth International Conference on Learning Representations},
  2023.

\bibitem{xi2024training}
Z.~Xi, W.~Chen, B.~Hong, S.~Jin, R.~Zheng, W.~He, Y.~Ding, S.~Liu, X.~Guo,
  J.~Wang \emph{et~al.}, ``Training large language models for reasoning through
  reverse curriculum reinforcement learning,'' \emph{arXiv preprint
  arXiv:2402.05808}, 2024.

\bibitem{9709543}
L.~Cheng, R.~Guo, R.~Moraffah, P.~Sheth, K.~S. Candan, and H.~Liu, ``Evaluation
  methods and measures for causal learning algorithms,'' \emph{IEEE
  Transactions on Artificial Intelligence}, vol.~3, no.~6, pp. 924--943, 2022.

\bibitem{wagner2021neural}
B.~Wagner and A.~d. Garcez, ``Neural-symbolic integration for interactive
  learning and conceptual grounding,'' \emph{arXiv preprint arXiv:2112.11805},
  2021.

\bibitem{ribeiro2016should}
M.~T. Ribeiro, S.~Singh, and C.~Guestrin, ``" why should i trust you?"
  explaining the predictions of any classifier,'' in \emph{Proceedings of the
  22nd ACM SIGKDD international conference on knowledge discovery and data
  mining}, 2016, pp. 1135--1144.

\bibitem{bau2018gan}
D.~Bau, J.-Y. Zhu, H.~Strobelt, B.~Zhou, J.~B. Tenenbaum, W.~T. Freeman, and
  A.~Torralba, ``Gan dissection: Visualizing and understanding generative
  adversarial networks,'' \emph{arXiv preprint arXiv:1811.10597}, 2018.

\bibitem{zhu2024survey}
X.~Zhu, J.~Li, Y.~Liu, C.~Ma, and W.~Wang, ``A survey on model compression for
  large language models,'' \emph{Transactions of the Association for
  Computational Linguistics}, vol.~12, pp. 1556--1577, 2024.

\bibitem{xiao2023smoothquant}
G.~Xiao, J.~Lin, M.~Seznec, H.~Wu, J.~Demouth, and S.~Han, ``Smoothquant:
  Accurate and efficient post-training quantization for large language
  models,'' in \emph{International Conference on Machine Learning}.\hskip 1em
  plus 0.5em minus 0.4em\relax PMLR, 2023, pp. 38\,087--38\,099.

\bibitem{xu2024survey}
X.~Xu, M.~Li, C.~Tao, T.~Shen, R.~Cheng, J.~Li, C.~Xu, D.~Tao, and T.~Zhou, ``A
  survey on knowledge distillation of large language models,'' \emph{arXiv
  preprint arXiv:2402.13116}, 2024.

\bibitem{singh2025agentic}
A.~Singh, A.~Ehtesham, S.~Kumar, and T.~T. Khoei, ``Agentic retrieval-augmented
  generation: A survey on agentic rag,'' \emph{arXiv preprint
  arXiv:2501.09136}, 2025.

\bibitem{yang2025agentnet}
Y.~Yang, H.~Chai, S.~Shao, Y.~Song, S.~Qi, R.~Rui, and W.~Zhang, ``Agentnet:
  Decentralized evolutionary coordination for llm-based multi-agent systems,''
  \emph{arXiv preprint arXiv:2504.00587}, 2025.

\bibitem{saleh2025usercentrix}
A.~Saleh, S.~Tarkoma, P.~K. Donta, N.~H. Motlagh, S.~Dustdar, S.~Pirttikangas,
  and L.~Lov{\'e}n, ``Usercentrix: An agentic memory-augmented ai framework for
  smart spaces,'' \emph{arXiv preprint arXiv:2505.00472}, 2025.

\bibitem{tran2025multi}
K.-T. Tran, D.~Dao, M.-D. Nguyen, Q.-V. Pham, B.~O'Sullivan, and H.~D. Nguyen,
  ``Multi-agent collaboration mechanisms: A survey of llms,'' \emph{arXiv
  preprint arXiv:2501.06322}, 2025.

\bibitem{krishnan2025advancing}
N.~Krishnan, ``Advancing multi-agent systems through model context protocol:
  Architecture, implementation, and applications,'' \emph{arXiv preprint
  arXiv:2504.21030}, 2025.

\bibitem{habler2025building}
I.~Habler, K.~Huang, V.~S. Narajala, and P.~Kulkarni, ``Building a secure
  agentic ai application leveraging a2a protocol,'' \emph{arXiv preprint
  arXiv:2504.16902}, 2025.

\bibitem{zhuge2024agent}
M.~Zhuge, C.~Zhao, D.~Ashley, W.~Wang, D.~Khizbullin, Y.~Xiong, Z.~Liu,
  E.~Chang, R.~Krishnamoorthi, Y.~Tian \emph{et~al.}, ``Agent-as-a-judge:
  Evaluate agents with agents,'' \emph{arXiv preprint arXiv:2410.10934}, 2024.

\bibitem{yehudai2025survey}
A.~Yehudai, L.~Eden, A.~Li, G.~Uziel, Y.~Zhao, R.~Bar-Haim, A.~Cohan, and
  M.~Shmueli-Scheuer, ``Survey on evaluation of llm-based agents,'' \emph{arXiv
  preprint arXiv:2503.16416}, 2025.

\end{thebibliography}

	\newpage
\end{document}